\pdfoutput=1

\documentclass[opre,nonblindrev]{informs3}

\OneAndAHalfSpacedXII

\usepackage{natbib}
 \bibpunct[, ]{(}{)}{,}{a}{}{,}%
\usepackage{mathtools}
\usepackage{xcolor}
\usepackage{wrapfig}
\usepackage{caption}
\usepackage{subfig}
\usepackage{hyperref}

\definecolor{lightblue}{RGB}{0, 10, 185}

\usepackage{amsmath}
{
      \theoremstyle{plain}
      \newtheorem{assumption}{Assumption}
}

\newcommand{\bs}[1]{\boldsymbol{#1}}
\newcommand{\mathbbm}[1]{\text{\usefont{U}{bbm}{m}{n}#1}} 
\newcommand{\indep}{\mathrel{\perp\!\!\!\perp}}

\begin{document}

\MANUSCRIPTNO{}

\TITLE{Responsible Machine Learning via \\ Mixed-Integer Optimization}

\RUNAUTHOR{Justin, Sun, G\'omez, and Vayanos}

\RUNTITLE{Responsible Machine Learning via Mixed-Integer Optimization}


\ARTICLEAUTHORS{
\AUTHOR{Nathan Justin}
\AFF{Center for Artificial Intelligence in Society and Department of Computer Science, University of Southern California, Los Angeles, California 90089, USA,
\EMAIL{njustin@usc.edu}}
\AUTHOR{Qingshi Sun}
\AFF{Center for Artificial Intelligence in Society and Department of Industrial \& Systems Engineering, University of Southern California, Los Angeles, California 90089, USA,
\EMAIL{qingshis@usc.edu}}
\AUTHOR{Andr\'{e}s G\'{o}mez}
\AFF{Center for Artificial Intelligence in Society and Department of Industrial \& Systems Engineering, University of Southern California, Los Angeles, California 90089, USA,
\EMAIL{gomezand@usc.edu}}
\AUTHOR{Phebe Vayanos}
\AFF{Center for Artificial Intelligence in Society, Department of Industrial \& Systems Engineering, and Department of Computer Science, University of Southern California, Los Angeles, California 90089, USA,
\EMAIL{phebe.vayanos@usc.edu}}
}


\ABSTRACT{
In the last few decades, Machine Learning (ML) has achieved significant success across domains ranging from healthcare, sustainability, and the social sciences, to criminal justice and finance. But its deployment in increasingly sophisticated, critical, and sensitive areas affecting individuals, the groups they belong to, and society as a whole raises critical concerns around fairness, transparency and robustness, among others. As the complexity and scale of ML systems and of the settings in which they are deployed grow, so does the need for {responsible ML} methods that address these challenges while providing guaranteed performance in deployment. 

Mixed-integer optimization (MIO) offers a powerful framework for embedding responsible ML considerations directly into the learning process while maintaining performance. For example, it enables learning of inherently transparent models that can conveniently incorporate fairness or other domain specific constraints. This tutorial paper provides an accessible and comprehensive introduction to this topic discussing both theoretical and practical aspects. It outlines some of the core principles of responsible ML, their importance in applications, and the practical utility of MIO for building ML models that align with these principles. Through examples and mathematical formulations, it illustrates practical strategies and available tools for efficiently solving MIO problems for responsible ML. It concludes with a discussion on current limitations and open research questions, providing suggestions for future work.}


\maketitle

\section{Introduction}


In the last few decades, Machine Learning (ML) has achieved tremendous success across domains ranging from healthcare, sustainability, and the social sciences, to criminal justice and finance. For instance, ML has been employed to predict breast cancer risk, significantly enhancing the identification of high-risk individuals—from an accuracy of 0.45 using conventional risk calculators to as high as 0.71 \cite{yala2019deep}. It has also been utilized to forecast wildfire risk, achieving a remarkable prediction accuracy of $98.32\%$ \cite{SAYAD2019130}. Additionally, ML models have been applied to assess the risk of suicidal ideation and death by suicide, offering critical insights into the key factors associated with these adverse outcomes \cite{RiskofSuicide}. But its deployment in increasingly sophisticated, critical, and sensitive domains affecting individuals, the groups they belong to, and society as a whole raises critical concerns. 

Inaccurate predictions may have grave consequences. For example, they may exclude those most at risk of suicide from appropriate preventive interventions \cite{RiskofSuicide}. Unfair ML systems (e.g., those trained on biased historical data) can reinforce or amplify existing social biases, leading to unfair treatment of individuals based on race, gender, age, disability, or other protected attributes \cite{mehrabi2021survey, chen2023ethics, pmlr-v81-buolamwini18a}. For example, an ML system for illness risk prediction that systematically underestimates risk for socially disadvantaged groups will result in individuals from that group being denied treatments they need at greater rates, exacerbating social inequities \cite{GroupBaisinHealth}. 
ML systems that are sensitive to changes of the input data may be prone to harmful manipulation or end up performing very differently than originally anticipated~\cite{DataPoisoning, MORENOTORRES2012521, quinonero2022dataset, pmlr-v37-xiao15, poisonFrogs}. For example, a fleet of autonomous self driving cars that can detect stop signs with high accuracy but that is sensitive to perturbations of the input that are imperceptible to the human eye, may be easily attacked by malicious agents, causing major disruptions to road networks \cite{IEEEautonomousVehicle,9425267}. ML systems that leak personal or confidential data may violate laws about the collection, use, and storage of data and about the controls that consumers have on how their data is used \cite{KHALID2023106848, SecureML, PrivacyASurvey}. An ML algorithm trained on data, such as X-rays, pooled from many different hospitals without privacy considerations would violate privacy laws and create major security and compliance risks \cite{9153891, hipaa1996}. In fact, even the perception of bias, privacy violation, or model inadequacy, may erode trust and support in the ML model, hindering adoption and reducing effectiveness of otherwise useful tools \cite{TrustworthyAI, 9635182}. For these reasons, it is essential that models used for high-stakes decisions be transparent so that stakeholders can scrutinize them and identify potential performance or fairness issues and flag potential unintended outcomes before the system is deployed. Due to these growing concerns about the use of ML in such settings, several governments have proposed and adopted regulations aimed at curtailing the risks associated with the increasing use of artificial intelligence~\citep{goodman2017european,eu_ai_act_proposal_2021, ised_ai_code_2023}.

These considerations are the unifying theme of responsible ML. Models need to be \textbf{accurate} in their predictions, but also \textbf{fair} in ways that are sensitive to preexisting inequalities and that align with anti-discrimination laws, \textbf{robust} to potential distribution shifts and to malicious attacks, \textbf{privacy-preserving} ensuring no personal or sensitive data is leaked, and \textbf{interpretable} allowing downstream decisions to be easily explained, justified, and scrutinized. Naturally, the precise meanings of accuracy, fairness, robustness, privacy, and interpretability are highly context-dependent: they require close collaboration between developers of the system and key stakeholders to be defined in ways that align with human values, and flexible ML models that can embody a variety of instantiations of these desiderata (e.g., different notions of fairness or robustness).


In the last ten years, mixed-integer optimization (MIO), a class of models and associated solution algorithms tailored for constrained optimization problems involving both real-valued and integer-valued decision variables, has emerged as a powerful and flexible framework for learning responsible ML models. The combinatorial nature of interpretable models is naturally amenable to modeling with MIO whose flexibility allows for capturing a variety of objectives by changing the loss function that measures errors, and incorporating diverse constraints encoding fairness or other domain-specific requirements. Robust variants of these problems can be used to learn models that provide guaranteed performance under distribution shifts, adversarial attacks, and other unmodeled phenomena.



\subsection{Frameworks for Responsible ML}

ML models can be learned responsibly in the context of all of supervised learning, unsupervised learning, and off-policy learning tasks. Independently of the precise form of the model considered and of the specific task at hand, the problem of learning ML models using responsible considerations can typically be modeled as a constrained optimization problem involving at least some combinatorial choices. We present the general forms of these models in the context of unsupervised, supervised, and policy learning in turn. These rely on a training dataset of size~$n$, indexed in the set~$\mathcal{I}$. The aim of both tasks is to learn the parameters~$\bs{\theta}$ of a function~$f_{\bs{\theta}}$ using the training data to minimize some loss function~$\ell$, where $\ell$ is chosen in a way that~$f_{\bs{\theta}}$ performs well under some metric on new data in the testing set and/or in deployment.

For \textit{unsupervised learning} tasks, the training dataset is given by~$\{\mathbf{x}^i\}_{i \in \mathcal{I}}$ where, for each datapoint $i\in \mathcal I$,~$\mathbf{x}^i \in \mathbb{R}^d$ collects its $d$ covariates/features. We denote the $j$th feature of sample~$i$ by~$x^i_j$,~$j \in [d] = {1, \ldots, d}$. The goal is to learn a function~$f_{\bs{\theta}} : \mathbb{R}^d \to \mathcal{Z}$ that maps the input covariates to some target space~$\mathcal{Z}$, which is determined by the specific task—for example, assigning cluster labels or producing a lower-dimensional representation. The parameters~$\bs{\theta}$ are learned by minimizing a task-specific loss function~$\ell$, resulting in the optimization problem: 
\begin{equation}
\label{eq:unsupervised} 
\min_{\bs{\theta} \in \Theta} \;\; \sum_{i \in \mathcal{I}} \ell(f_{\bs{\theta}}(\mathbf{x}^i)), 
\end{equation}
where $\Theta$ is a (potentially data-dependent) set enforcing e.g., fairness or interpretability constraints, or other domain specific requirements.


For \textit{supervised learning} tasks, the training dataset is given by~$\{(\mathbf{x}^i, y^i)\}_{i \in \mathcal{I}}$, including both input features~$\mathbf{x}^i \in \mathbb R^d$ and corresponding labels~$y^i \in \mathcal Y$. In \textit{classification} tasks, the label space $\mathcal Y$ is finite. A common special case is \textit{binary classification}, where the labels are typically encoded as either~$\{0, 1\}$ or~$\{-1, 1\}$, depending on the modeling convention. In \textit{regression} tasks, the labels~$y^i$ are real-valued.
The objective in supervised learning is to learn parameters~$\bs{\theta}$ of a function~$f_{\bs{\theta}} : \mathbb{R}^d \to \mathcal{Y}$ that solve
\begin{equation}\label{eq:supervised}
    \min_{\bs{\theta} \in \Theta} \;\; \sum_{i \in \mathcal{I}} \ell(f_{\bs{\theta}}(\mathbf{x}^i), y^i),
\end{equation}
where the loss function~$\ell$ quantifies the discrepancy between the model predictions~$f_{\bs{\theta}}(\mathbf{x}^i)$ and the true labels~$y^i$ in the data. 

For \emph{policy learning} tasks, the dataset contains a \textit{treatment} assigned to each datapoint~$i$, which is an independent variable that corresponds to, e.g., a group assigned in a clinical trial or an intervention made by a governing body. An \textit{outcome} is then observed that measures the effectiveness of the treatment for each individual. The training dataset is given by~$\{(\mathbf{x}^i, a^i, y^i)\}_{i \in \mathcal{I}}$ where for each datapoint $i \in \mathcal I$, $\mathbf{x}^i$ collects its features/covariates,~$a^i \in \mathcal A$ is the treatment historically received by datapoint $i$ where~$\mathcal{A}$ is a finite set indexing possible treatments, and~$y^i \in \mathcal{Y}$ is the observed outcome for this datapoint, i.e., its outcome under treatment~$a^i$. Without loss of generality, we assume in this tutorial that a larger outcome value is preferred (e.g., in the context of substance abuse treatment policy selection, outcomes could correspond to days ``clean''). The goal of such tasks is to learn a policy~$f_{\bs{\theta}} : \mathbb{R}^d \mapsto \mathcal{A}$ that assigns treatments to covariates in a way that maximizes expected outcomes. 
More details on policy learning tasks are given in section~\ref{sec:observational}.

In responsible ML, due to the key consideration of interpretability, the model $f_{\bs{\theta}}$ and/or the set $\Theta$ are often combinatorial in nature to model, e.g., logical relationships or sparsity constraints. Robust variants of these problems that seek to learn models that perform well even under adversarial attacks or distribution shifts often need to hedge against discrete perturbations of the data (to cater for the discrete nature of data which is pervasive in high-stakes settings). For these reasons, problems~\eqref{eq:unsupervised}, \eqref{eq:supervised}, their policy learning counterpart, and their robust variants are often combinatorial in nature. MIO is a powerful tool for modeling and solving such constrained combinatorial problems, making them a natural tool of choice for responsible ML. This important observation has spearheaded, over the last two decades, an important line of research aimed at leveraging the power of MIO to address many important problems in responsible ML, which is the topic of this tutorial. We restrict the methods discussed to mixed-integer optimization techniques, relying on constructing convex relaxations and using branch-and-bound. Other exact methods for discrete optimization, including constraint programming and dynamic programming, are not covered in the tutorial.

\subsection{Intended Audience and Background Needed for this Tutorial}

This tutorial paper provides an accessible and comprehensive introduction to the topic of using MIO for responsible ML, accessible to all constituents of the INFORMS community, including current students, practitioners, faculty, and researchers. It assumes basic knowledge of optimization and machine learning (at the level of a first year PhD class) and no knowledge of responsible ML \cite{boyd2004convex, hastie2009elements}. The background needed on MIO is introduced in the next section. Knowledge needed on responsible ML will be picked up throughout the tutorial.

\subsection{Background on Mixed-Integer Optimization}

We begin with a general introduction to mixed-integer optimization. Let~$\mathbf{z} \in \mathbb{Z}^p$ and~$\mathbf{r} \in \mathbb{R}^q$ be vectors collecting integer and real-valued decision variables, respectively. Consider an optimization problem defined over both these variables, where the objective is to minimize an objective function~$g(\mathbf{z}, \mathbf{r}):\mathbb{Z}^p \times \mathbb{R}^q \rightarrow \mathbb{R}$,  subject to a set of constraints. Such a problem is an MIO, which can be expressed as
\begin{equation}\label{eq:mio}
    \begin{aligned}
        \min_{\mathbf{z} \in \mathbb{Z}^p, \mathbf{r} \in\mathbb{R}^q} \ & g(\mathbf{z}, \mathbf{r})\\
        \text{s.t.} \hspace{5.5mm} & h_j(\mathbf{z}, \mathbf{r}) \leq 0 &\forall j\in \mathcal{J},
    \end{aligned}
\end{equation}
\begin{wrapfigure}{r}{0.32\textwidth}
  \begin{center}
\includegraphics[width=0.32\textwidth, trim={0cm 0cm 0cm 2cm}, clip]{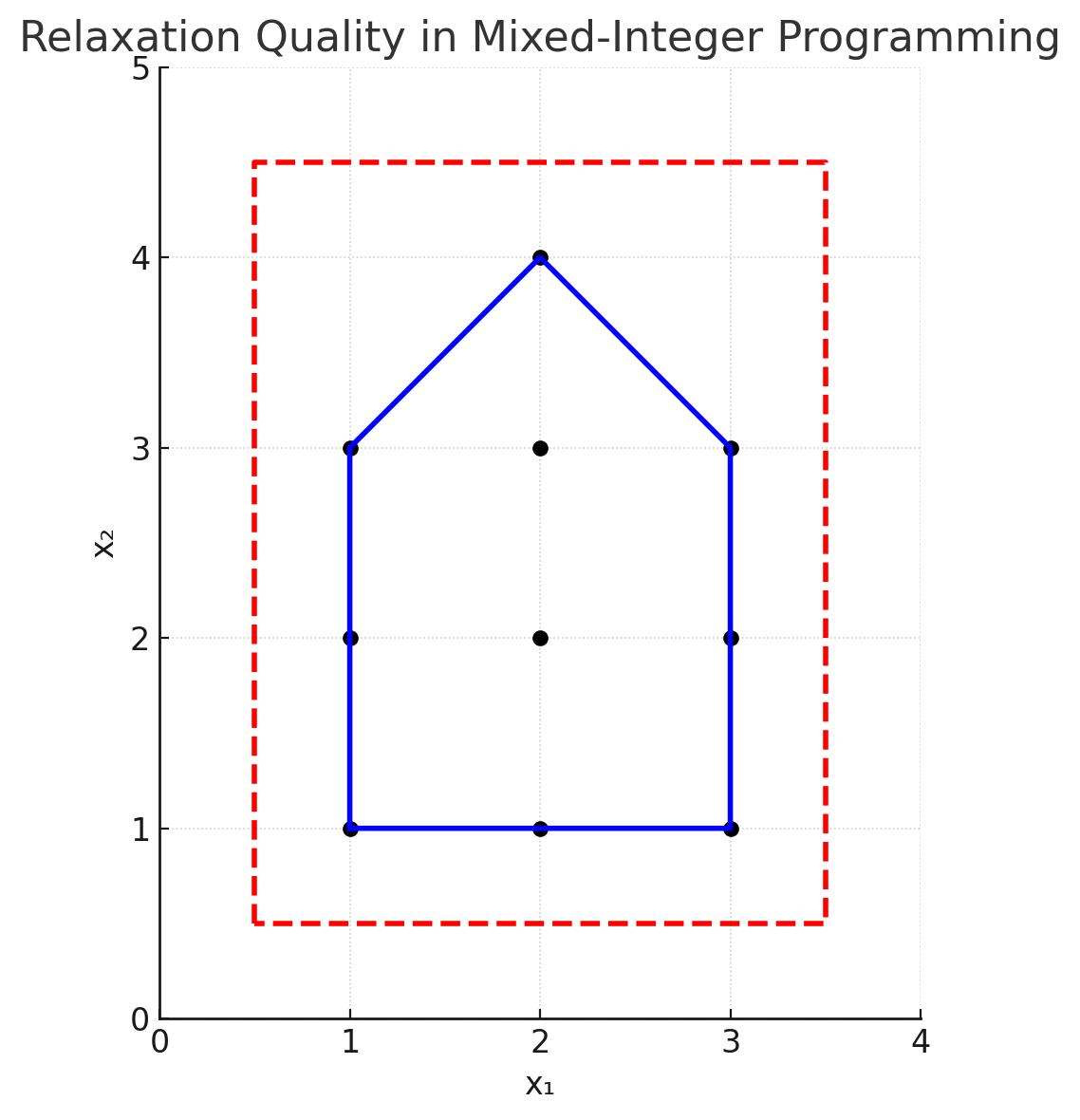}
  \end{center}
  \caption{MIO relaxations. Black dots represent the discrete feasible region of an MIO problem. Solid lines depict the convex hull, while dashed lines depict a weaker linear relaxation.}
  \label{fig:mio}
\end{wrapfigure}
where $h_j(\mathbf{z}, \mathbf{r}):\mathbb{Z}^p \times \mathbb{R}^q \rightarrow \mathbb{R}$ are functions encoding hard constraints and $\mathcal{J}$ is the set 
of indices of those constraints.
If all functions~$g$ and~$h_j$, $j \in \mathcal J$ are affine, the problem is known as mixed-integer linear optimization (MILO). If~$g$ is quadratic and~$h$ is linear, it is referred to as mixed-integer quadratic optimization (MIQO). If~$h$ includes conic constraints—such as second-order cone, semidefinite, or exponential cones—the problem falls under mixed-integer conic optimization (MICO). When all functions $g$ and $h$ are convex, then the problem is referred as a convex mixed-integer nonlinear optimization (convex MINLO). Finally, when~$g$ and~$h$ are general nonlinear functions, the problem is called simply mixed-integer nonlinear optimization (MINLO)~\citep{belotti2013mixed, floudas1995nonlinear, lazimy1982mixed, vielma2015mixed, wolsey1999integer}. These classes of problems are nested, e.g., 
MILOs are special cases of MIQOs which in turn are special cases of MICOs, and so forth. As a simple rule of thumb, MILOs are generally significantly easier to solve than the other classes with current technology, and general MINLO are substantially more difficult than convex MINLOs. 

Optimization problem \eqref{eq:mio} is non-convex and may be hard to solve due to presence of 
integrality constraints $\mathbf{z} \in \mathbb{Z}^p$, while its natural continuous relaxation obtained by dropping these constraints is easier to handle. In particular, for convex MINLOs, the continuous relaxation is typically easy to solve to optimality since any local minimum is a global minimum. MIO problems with strong continuous relaxations, that is, with continuous relaxations that are close approximations of the discrete problems, can often be solved to optimality at medium or large scales. The best possible convex relaxation of an MIO, which we refer to as the convex hull relaxation, ensures that it can be solved as a convex problem, see Figure~\ref{fig:mio} for a depiction. MIO problems with weak relaxations, on the other hand, cannot be solved to optimality without substantial enumeration, which may be prohibitive computationally at large scales. In fact, strength of the continuous relaxation is often a better predictor of the difficulty of an MIO problem than more natural metrics such as number of variables or constraints. 

MIOs have become a powerful tool for ML in recent years, thanks to their modeling capabilities and advancements in solving techniques. For instance, we can model ML tasks as MIOs by letting the decision variables~$\mathbf{z}$ and~$\mathbf{r}$ correspond to model parameters~$\bs{\theta}$, the constraints of~\eqref{eq:mio} define the space of~$f_{\bs{\theta}}$, and the objective function of~\eqref{eq:mio} calculate the loss function~$\ell$. Unlike traditional methods for learning ML models, the objective function and constraints can be more easily customized to address various model desiderata and domain-specific constraints. Furthermore, MIOs can be modeled in and solved with a number of solvers to optimality or with guarantees on solution quality. Moreover, state-of-the-art MIO solvers are equipped with powerful general purpose heuristics, and can typically find high-quality solutions to optimization problems in seconds.

\subsection{Structure of this Tutorial}

This tutorial will explore recent advances on using MIOs for responsible ML. Section~\ref{sec:interpretability} examines MIO-based methods for developing interpretable ML models and for explaining black-box models to enhance transparency. We then discuss robustness to the issues of data availability, quality, and bias in Section~\ref{sec:robustness}. 
Section~\ref{sec:observational} is devoted to policy learning problems and 
in section~\ref{sec:privacy-fair}, we examine how MIO can be used to enforce some important principles in ML, particularly fairness. We conclude the tutorial with section~\ref{sec:conclusion}, where we highlight unresolved and/or underexplored issues in responsible ML that MIO has the potential to address, paving the way for future research on this topic.

\section{Interpretability and Explainability} \label{sec:interpretability}
Model transparency in ML can be achieved through two strategies: building interpretable models or generating explanations for black-box models.

\textit{Interpretable machine learning} refers to learning white-box models
such that the learned ML model is ``comprehensible'' for humans. Comprehensibility is subjective and domain-specific, making interpretability difficult to measure. Depending on the application, white-box models may correspond to simple classes of \( f_{\bs{\theta}} \) (e.g., additive rules), functions that can be expressed and/or evaluated using pencil and paper, sequential logical statements, and/or visualizations that enhance understanding~\cite{rudin2022interpretable}.
One method of quantifying interpretability is proposed by~\citet{jo2023learning}, which they call \textit{decision complexity}. The decision complexity of a classifier is the minimum number of parameters needed to make a prediction on a new datapoint. This definition captures interpretability through the concepts of simpler, human-comprehensible model classes and sparsity constraints. A critical consideration when using an interpretable model is a potential interpretability-accuracy trade-off, where accuracy may decrease as the models considered become more interpretable~\citep{alcala2006hybrid, baryannis2019predicting, dziugaite2020enforcing, johansson2011trade, you2022interpretability}. The magnitude of such a trade-off and the benefits of gaining model interpretability under some loss in accuracy is domain-specific.

\textit{Explainable artificial intelligence} (XAI) refers to methods used to explain black-box models for human comprehension. Explanations can use texts, visuals, examples, relevant features, or a comparable interpretable model, and can be localized to a subspace of the data or globally applied. The major challenge is learning explanations that are simple, are accurate, and explain a large portion of the data~\citep{li2022optimal, ribeiro2018anchors}. Note that in high-stakes domains, interpretable models are preferred over explanations if the learning task, dataset, and domain allow for it and the tradeoff in accuracy is low~\citep{rudin2019stop}. XAI methods, on the other hand, can preserve the power of any ML model while providing some transparency in the form of explanations. 

MIOs have recently emerged as a natural modeling and solution tool for interpretable ML, as they are able to model logical constraints, integrality restrictions, and sparsity inherent in such models. By solving training problems to optimality (instead of resorting to heuristics), models obtained by solving MIOs achieve a better interpretability-accuracy tradeoff in sample. While some researchers have questioned whether exact models ``over-optimize" and overfit \cite{zharmagambetov2021non,dietterich1995overfitting,hastie2017extended}, others argue that with proper regularization exact models lead to superior performance out-of-sample \cite{mazumder2023subset,van2024optimal}. Settling this debate is beyond the scope of this paper; nonetheless, the authors believe that MIO methods can indeed achieve better performance out-of-sample, at the expense of a significant increase in computational cost.

The rest of this section is organized as follows.
Sections~\ref{sec:additive} and~\ref{sec:logic} explore how MIO has been used for interpretable supervised learning for additive and logical models, respectively. 
We also explore MIOs for unsupervised learning tasks such as principal component analysis (PCA), network learning and clustering in section~\ref{sec:int_unsupervised}. In section~\ref{sec:counterfact_ex}, we show how MIOs have been used to help explain the outputs of black-box models via counterfactual explanations. Finally, we discuss hyperparameter selection in section~\ref{sec:hyperparameter}.

\subsection{Additive Models for Supervised Learning}\label{sec:additive}
In this section, we explore how MIO can be used to learn (generalized) linear ML models. Specifically,~$f_{\bs{\theta}}$ is defined as a linear function, that is,
\begin{equation}\label{eq:add_ml}
    f_{\bs{\theta}}(\mathbf{x}) =  \boldsymbol{\theta}^\top\mathbf{x}.
\end{equation}
The function~$f_{\bs{\theta}}$ can be used for regression tasks where~$f_{\bs{\theta}}$ maps to a predicted label (linear regression) or a predicted probability (logistic regression and risk scoring), or used in binary classification by thresholding the predicted value of~$f_{\bs{\theta}}$ by 0. To enhance interpretability,~$\boldsymbol{\theta}$ can be restricted to be integer-valued and/or sparse. We explore four additive methods in this section: linear regression (section~\ref{sec:subsetselect_lr}), logistic regression and risk scores (section~\ref{sec:logreg_riskscore}), support vector machines (section~\ref{sec:svm}), and generalized additive models (section~\ref{sec:gam}).

\subsubsection{Subset Selection and Sparse Linear Regression} \label{sec:subsetselect_lr}
\begin{figure}
\begin{center}
\begin{minipage}{0.61\textwidth}
\subfloat[OLS (partial), $R^2=0.68$]{\includegraphics[width=\textwidth]{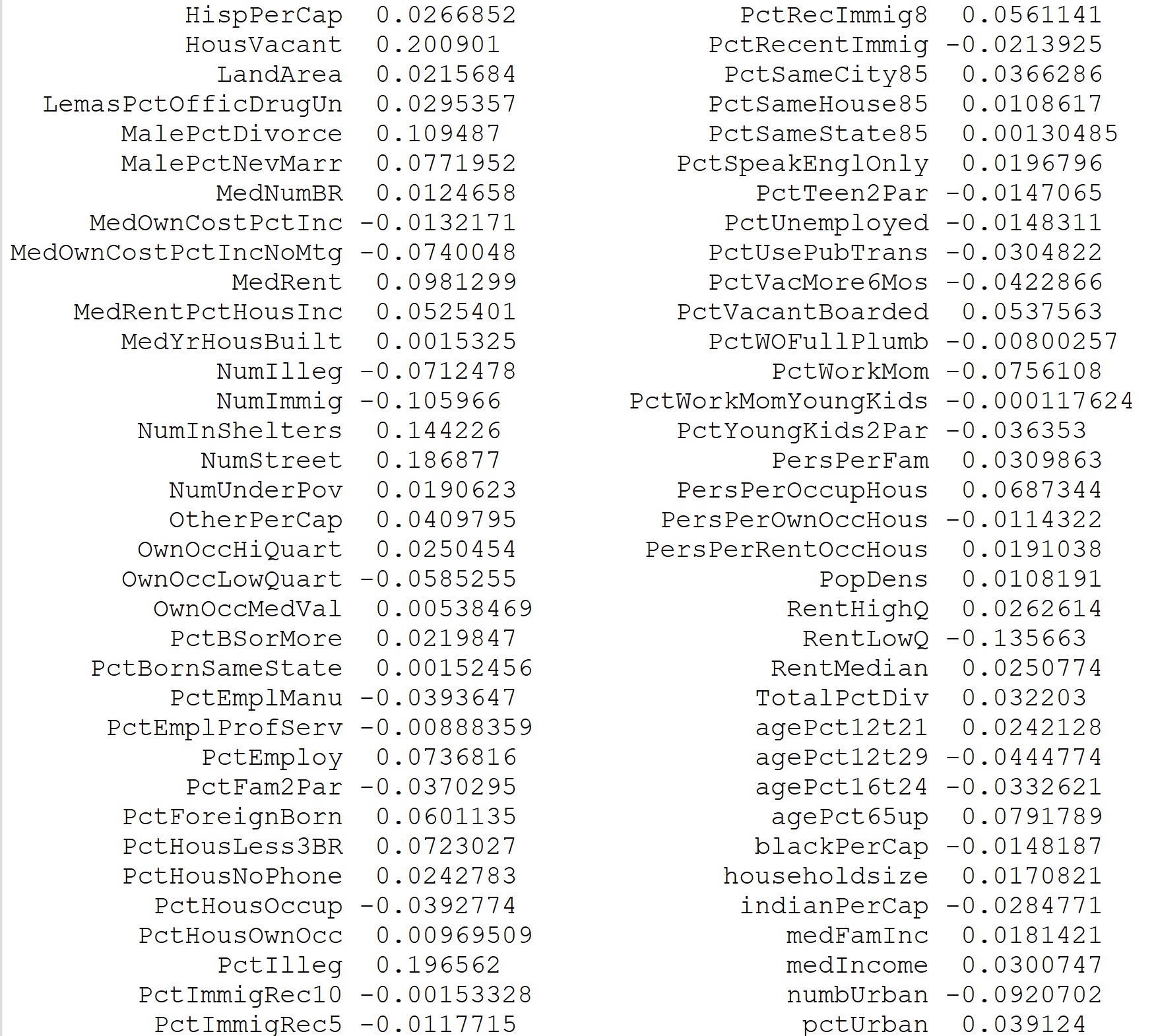}}
\end{minipage}
\hfill
\begin{minipage}{0.38\textwidth}
\vspace{0.1cm}
\begin{center}
\subfloat[``Dense" elastic net, $R^2=0.62$]{\includegraphics[width=0.92\textwidth, trim={0cm 10.3cm 16.5cm 0cm},clip]{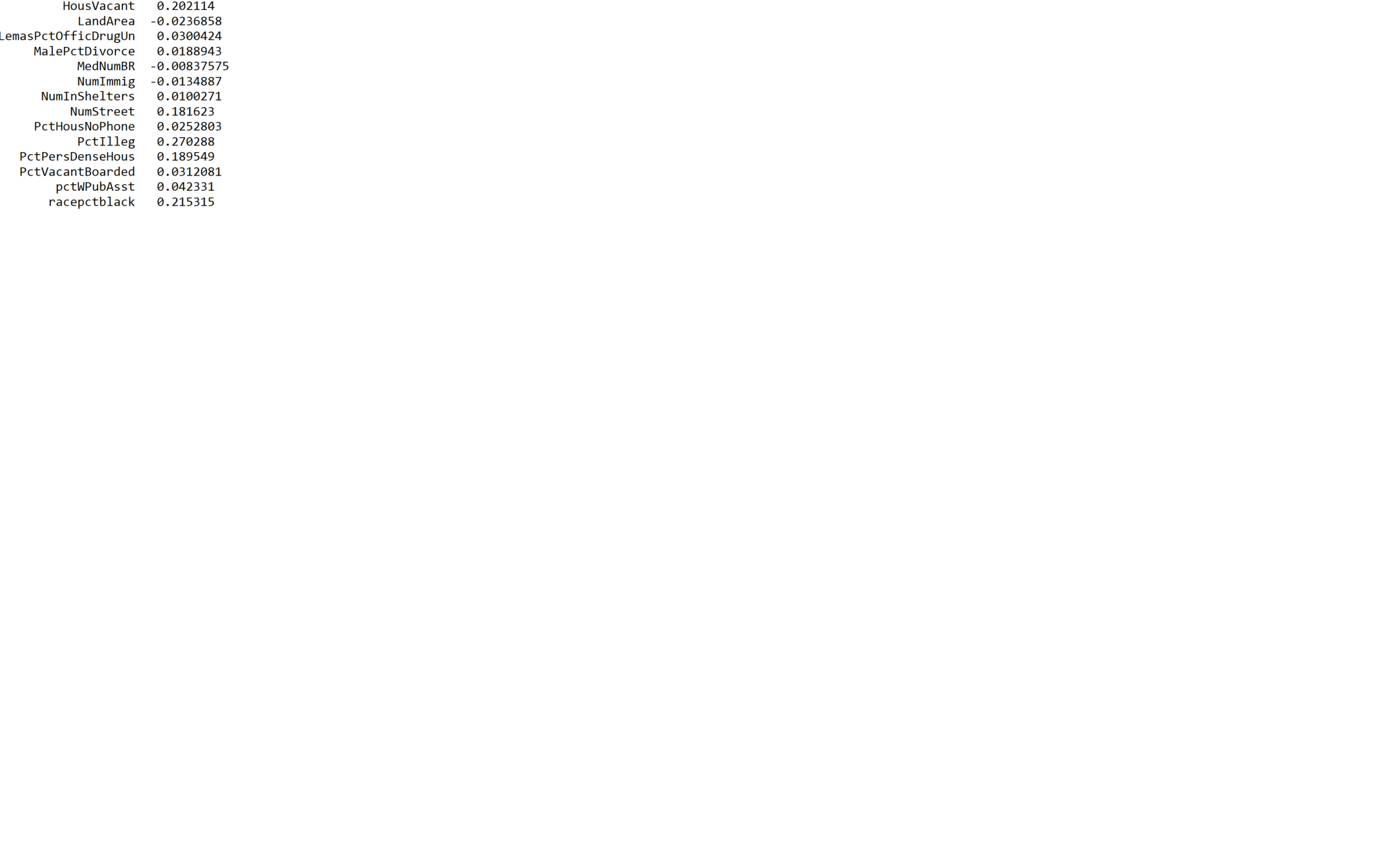}}

\vspace{0.1cm}

\subfloat[``Sparse" elastic net, $R^2=0.33$]{\includegraphics[width=0.92\textwidth, trim={0cm 12.7cm 17cm 0cm},clip]{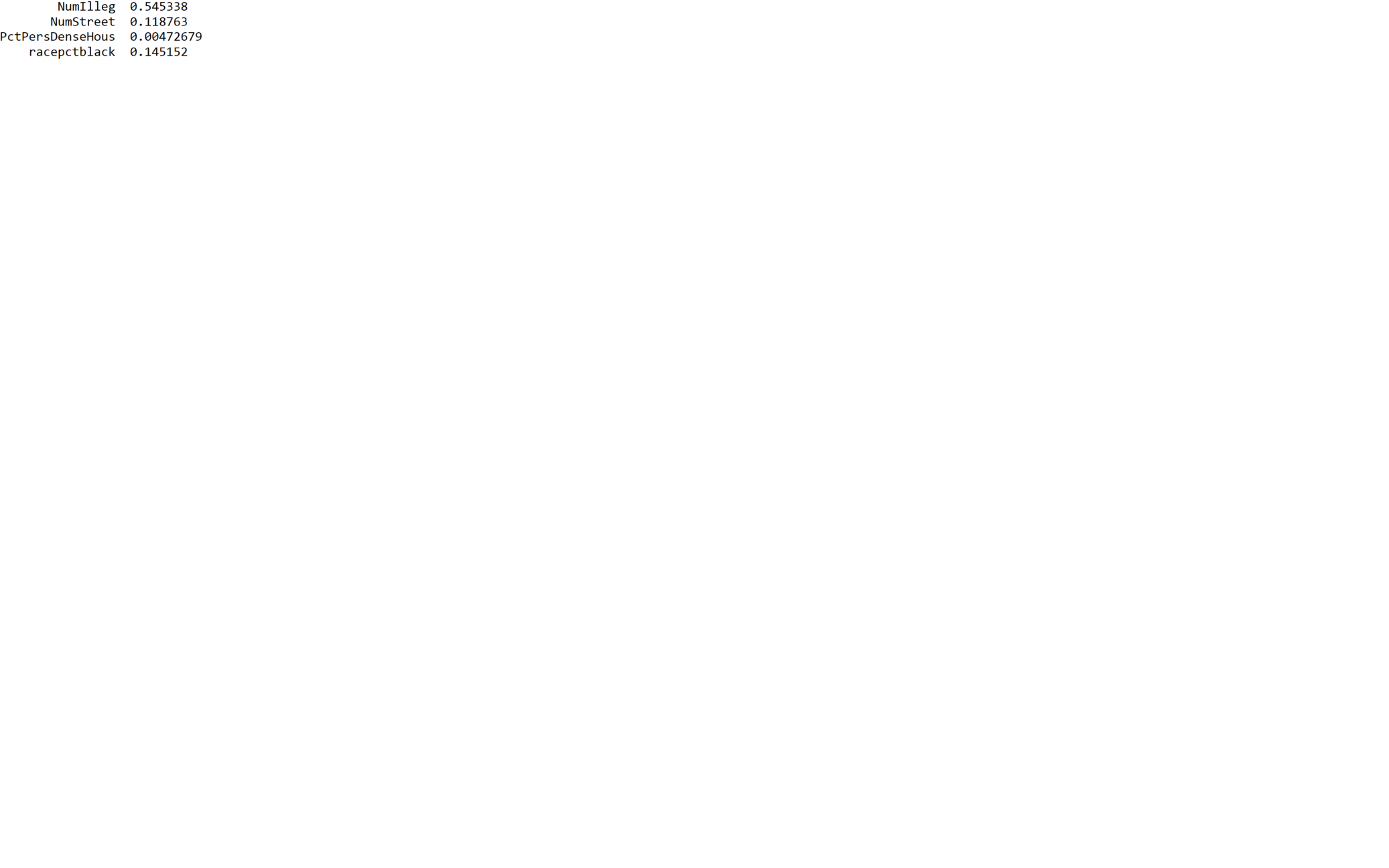}}

\vspace{0.1cm}

\subfloat[Problem \eqref{eq:l0l2_regress}, $R^2=0.63$]{\includegraphics[width=0.88\textwidth, trim={0cm 12.1cm 14cm 0cm},clip]{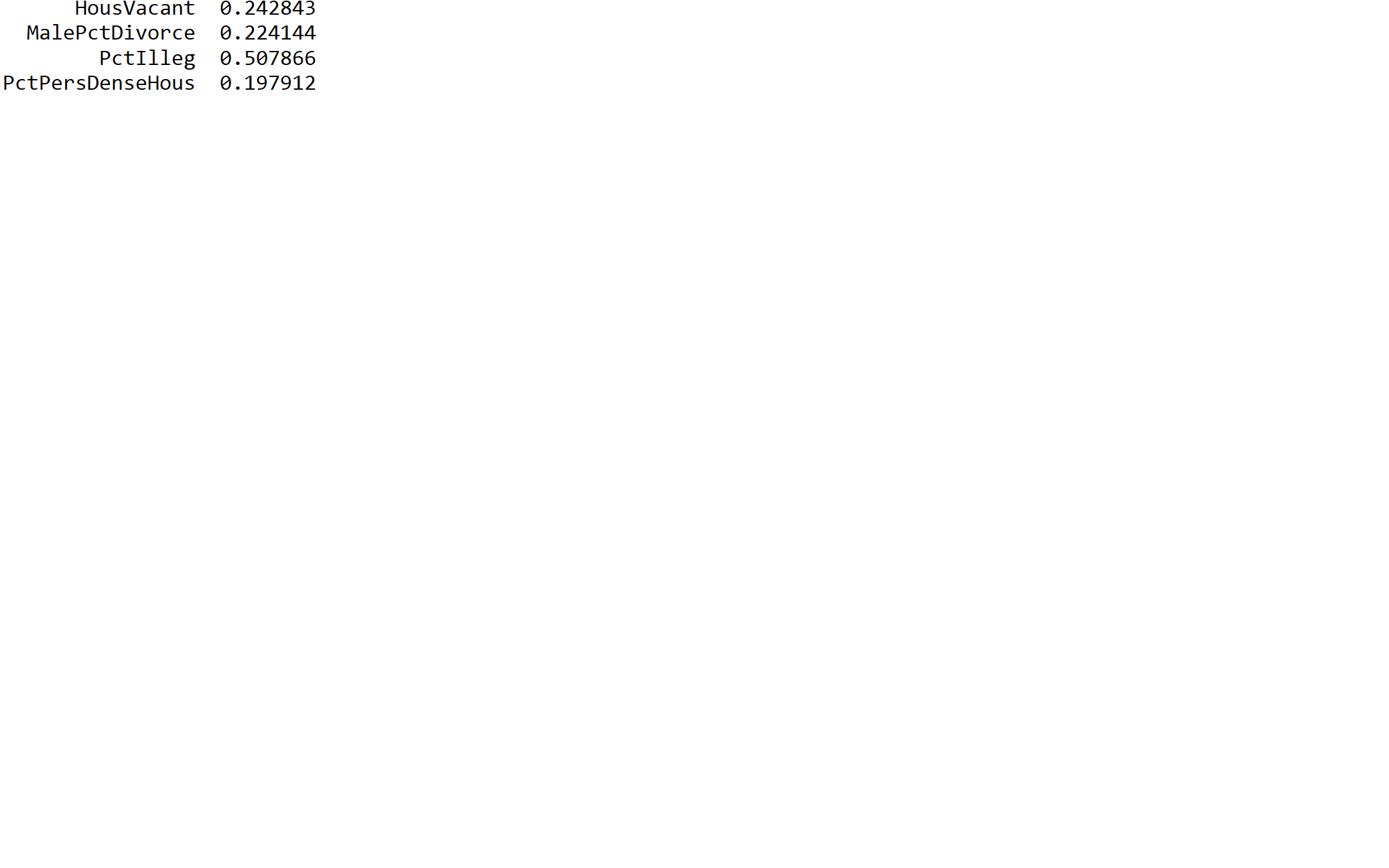}}
\end{center}
\end{minipage}
\caption{Solutions from three linear regression methods. The ordinary least squares (a) produces a solution that is not interpretable. The lasso estimator (b)-(c) can be used to induce interpretability, but may suffer from low accuracy. Using MIO (d), an accurate and interpretable solution is found.  PctIlleg: percentage of kids born to never married, NumStreet: number of homeless people counted in the street, PctPersDenseHous: percent of persons in dense housing (more than 1 person per room), racepctblack: percentage of population that is african american, HousVacant: number of vacant households, MalePctDivorce: percentage of males who are divorced.}
\label{fig:outputsLinReg}
\end{center}
\end{figure}

Given a covariate~$\mathbf{x}$, linear regression predicts a response via the learned relation~$f_{\bs{\theta}}(\mathbf{x}) = \boldsymbol{\theta}^\top\mathbf{x}$. The function~$f_{\bs{\theta}}(\mathbf{x})$ is found by solving problem~\eqref{eq:supervised} using the least squares loss~$ \ell(f_{\bs{\theta}}(\mathbf{x}^i), y^i) = (y^i - \boldsymbol{\theta}^\top\mathbf{x}^i)^2$, a method called ordinary least squares (OLS). If the number of features $d$ is small, OLS solutions are interpretable, as the sign and magnitude of a coefficient associated with each feature serves as a simple proxy of the effect of that feature. However, interpretability quickly fades if $d$ is large. Consider for example the \href{https://archive.ics.uci.edu/dataset/183/communities+and+crime}{``Communities and Crime"} dataset, containing socio-economic data, law enforcement data and crime data from $n=1,993$ cities and involving $d=100$ features. Figure~\ref{fig:outputsLinReg}a depicts the OLS solution: while effective at explaining the data, with a reasonably large $R^2$ value, the output is hard to interpret or use by policy makers.


To add interpretability to a linear regression model, penalties or constraints that limit the number of non-zero elements (sparsity) of~$\boldsymbol{\theta}$ can be added. Traditional approaches in the statistical literature add an L1 penalty \cite{tibshirani1996regression} $\lambda_1\|\bs{\theta}\|_1$ (a method called \textit{lasso}), or a combination of an L1 and L2 penalties  \cite{zou2005regularization} $\lambda_1\|\bs{\theta}\|_1+\lambda_2\|\bs{\theta}\|_2^2$ (referred to as \textit{elastic net}), where the regularization parameters $\lambda_1,\lambda_2\in \mathbb{R}_+$ need to be tuned with common methods like cross-validation and grid search (see~\citet{hastie2009elements} for an introduction on hyperparameter tuning). Indeed, if $\lambda_1$ is large enough, optimal solutions are sparse and interpretable. Figure~\ref{fig:outputsLinReg}b shows the elastic net solution where parameter $\lambda_1$ is tuned to yield an $R^2\approx0.62$, and Figure~\ref{fig:outputsLinReg}c shows the elastic net solution where $\lambda_1$ is tuned to produce four nonzero elements. While elastic net can produce accurate models with close to 15 features, it struggles to find accurate models with four features. The loss of accuracy occurs since the L1 penalty is an approximation of the sparsity of $\bs{\theta}$, but does not model it explicitly. A more accurate representation involves the so-called L0 norm, defined as~$\Vert\boldsymbol{\theta}\Vert_0=\sum_{j=1}^d\mathbbm{1}[\theta_j\neq 0]$ for $\mathbbm{1}$ the indicator function. Note that the L0 penalty is not actually a norm as it violates homogeneity, and (unlike norms) is non-convex. As a consequence, optimization problems involving such terms may be difficult to solve.
The least squares problem with L0 and L2 regularization is
\begin{equation} \label{eq:l0l2_regress}
        \min_{\boldsymbol{\theta} \in \mathbb{R}^d} \ \sum_{i\in \mathcal{I}} (y^i - \boldsymbol{\theta}^\top\mathbf{x}^i)^2  + \lambda_0 \Vert \boldsymbol{\theta}\Vert_0 + \lambda_2 \Vert\boldsymbol{\theta}\Vert_2^2,
\end{equation}
where $\lambda_0,\lambda_2\in \mathbb{R}_+$ are hyperparameters to be tuned. If $\lambda_2=0$ and the L0 norm is enforced as a constraint instead of a penalty, problem \eqref{eq:l0l2_regress} is the \textit{best subset selection} problem \cite{miller1984selection}. Figure~\ref{fig:outputsLinReg}d depicts the solution of problem \eqref{eq:l0l2_regress}, which is both interpretable and accurate.

There has been tremendous progress in solving \eqref{eq:l0l2_regress} using MIO technology over the past decade. Earlier works \cite{bertsimas2016best,gomez2021mixed,wilson2017alamo,miyashiro2015subset,miyashiro2015mixed} were based on reformulating the L0 constraint using big-M constraints. Letting $\mathbf{z}\in \{0,1\}^d$ be binary variables used to encode the support of $\bs{\theta}$, problem \eqref{eq:l0l2_regress} can be reformulated as 
\begin{equation}\label{eq:l0l2_mio}
    \begin{aligned}
        \min_{ \boldsymbol{\theta} \in \mathbb{R}^d, \mathbf{z} \in \{0,1\}^d} \ & \sum_{i\in \mathcal{I}} (y^i - \boldsymbol{\theta}^\top\mathbf{x}^i)^2  + \lambda_0 \sum_{j=1}^d z_j + \lambda_2 \sum_{j=1}^d\theta_j^2 \\
        \text{s.t.} \hspace{1mm}
        & -Mz_j \leq \theta_j \leq Mz_j \hspace{10mm} \forall j \in [d],
    \end{aligned}
\end{equation}
where the value of constant $M$ is typically chosen heuristically. As pointed out in \cite{dong2015regularization}, optimal solutions of the continuous relaxation of \eqref{eq:l0l2_mio} satisfy $z_j^*=|\theta_j|/M$, thus the continuous relaxation reduces to penalizing the L1 norm of the regression coefficients, i.e., the popular lasso estimator \cite{tibshirani1996regression}. Using the natural formulation \eqref{eq:l0l2_mio} led to the possibility of solving to optimality problems with $d$ in the dozens or low hundreds \cite{bertsimas2016best}-- a notable improvement over alternative exact methods available at the time, but insufficient in several practical applications.

To broaden the utility of MIO methods for sparse least squares on larger datasets, a major breakthrough in tackling \eqref{eq:l0l2_mio} was achieved by using the \textit{perspective reformulation}~\cite{xie2020scalable,dong2015regularization,bertsimas2020sparse,pmlr-v119-atamturk20a}, an MIO convexification technique originally derived using disjunctive programming~\cite{akturk2009strong,ceria1999convex,frangioni2006perspective,gunluk2010perspective}. At a high level, the reformulation first introduces additional epigraphical variables $r_j\geq \theta_j^2$, moving the nonlinearities induced by the ridge regularization to the constraints. Then it multiplies the left-hand side of the new constraints by $z_j$, resulting in formulation
\begin{subequations}\label{eq:l0l2_soc}
    \begin{align}
        \min_{\bs{\theta},\mathbf{z},\mathbf{r}} \ & \mathrlap{\sum_{i\in \mathcal{I}} (y^i - \boldsymbol{\theta}^\top\mathbf{x}^i)^2  + \lambda_0 \sum_{j=1}^d z_j + \lambda_2 \sum_{j=1}^d r_j} \\
        \text{s.t.} \hspace{1.5mm} & \mathrlap{\theta_j^2 \leq  z_jr_j} & \forall j \in [d] \label{eq:l0l2_soc_rotated}\\
        & \mathrlap{-Mz_j \leq \theta_j \leq Mz_j} & \forall j \in [d]\\
        &\mathrlap{\boldsymbol{\theta} \in \mathbb{R}^d, \mathbf{z} \in \{0,1\}^d, \mathbf{r} \in \mathbb{R}^d.}
    \end{align}
\end{subequations}
Since $0\leq z_j\leq 1$ in any feasible solution of the continuous relaxation, constraints \eqref{eq:l0l2_soc_rotated} produce a tighter relaxation than simple big-M formulations -- in fact, the big-M constraints can be safely removed. Moreover, since $\theta_j^2 \leq  z_jr_j \Leftrightarrow \sqrt{(2\theta_j)^2+(z_j-r_j)^2}\leq z_j+r_j$ (an identity that can be easily verified by squaring both sides of the inequality), and since function $h(\theta,z,r)=\sqrt{(2\theta)^2+(z-r)^2}- z+r$ is convex (since it is the sum of an L2 norm and a linear function), problem \eqref{eq:l0l2_soc} is an MICO. Constraints \eqref{eq:l0l2_soc_rotated}, referred to as rotated cone constraints, 
can in fact be modeled directly as is in most MICO solvers, with the software automatically performing the aforementioned transformation automatically. In fact, since 2022, the MIO solver Gurobi~\citep{gurobi} automatically performs this reformulation even when presented with the weaker formulation \eqref{eq:l0l2_mio}. The use of the perspective reformulation \eqref{eq:l0l2_soc} increased the dimension of problems that can be solved by an order-of-magnitude, with researchers reporting solutions to problems with~$d$ in the hundreds to low thousands.

A second major breakthrough was achieved by \citet{hazimeh2022sparse}: instead of relying on general purpose solvers, they implemented a branch-and-bound algorithm tailored for sparse regression based on formulation \eqref{eq:l0l2_soc}. Critically, they rely on first-order coordinate descent methods to solve the continuous relaxations, which scale to large-dimensional problems much better than the interior point methods or outer approximations used by most off-the-shelf solvers. The authors report optimal solutions to problems with $d$ in the hundreds of thousands to millions.

The progress in solving \eqref{eq:l0l2_regress} to optimality illustrates the potential of MIO technology for learning accurate and interpretable ML models on large datasets. In just a decade, the scale of problems that could be solved to optimality increased by several orders-of-magnitude, from dozens to hundreds of thousands, expanding the applicability of MIO technology for optimal and sparse linear regression. Some of the limitations of current methods is the dependence on the ridge term $\lambda_2\|\bs{\theta}\|_2^2$ to strengthen the formulation, as well as tailored branch-and-bound methods depending on the absence of additional constraints on the regression variables. Naturally, there is ongoing research addressing these limitations \cite{atamturk2025rank,tillmann2024cardinality}. In addition, there is also additional effort in developing primal methods to find high-quality solutions quickly as well as to better understand the statistical merits of solving \eqref{eq:l0l2_regress} to optimality \cite{hazimeh2020fast,mazumder2023subset}.


\subsubsection{Logistic Regression and Risk Scores} \label{sec:logreg_riskscore}
In logistic regression, the aim is to learn the probabilities that a given covariate~$\mathbf{x}$ has some binary class label~$y \in \{-1,1\}$. 
\begin{wrapfigure}{r}{0.4\textwidth}
  \begin{center}
    \includegraphics[width=0.4\textwidth]{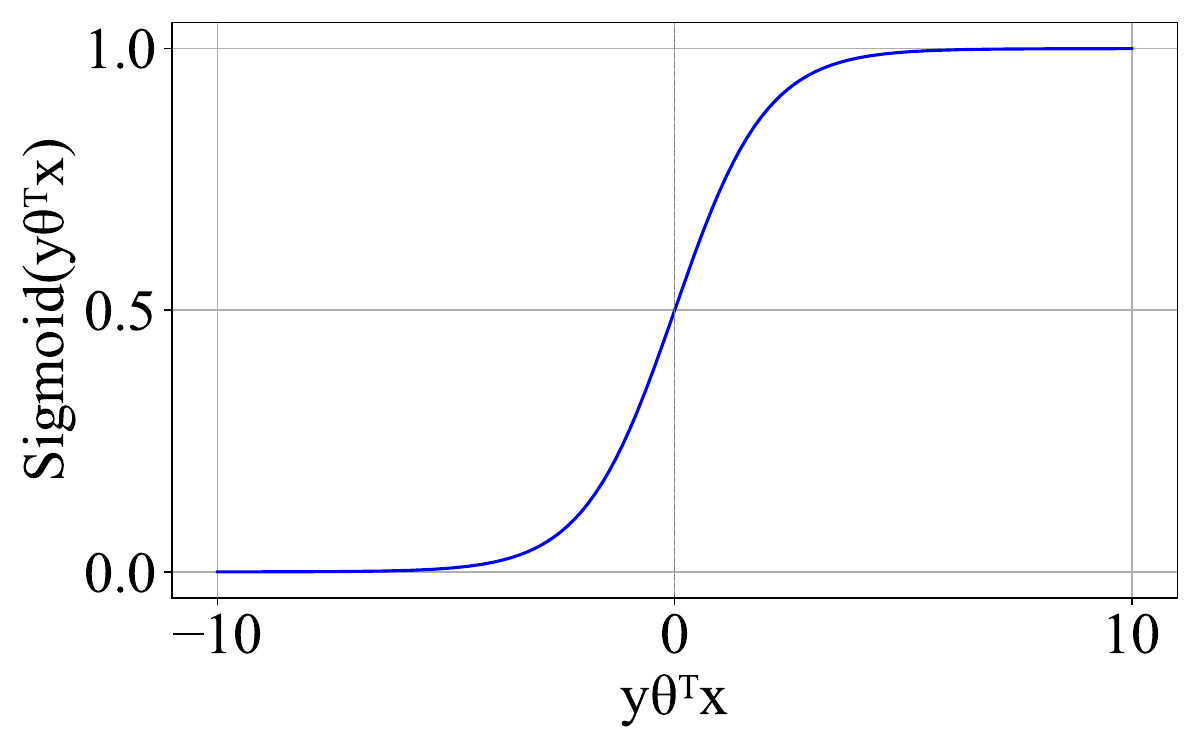}
  \end{center}
  \caption{The sigmoid function $\sigma(y\bs{\theta}^\top \mathbf{x}) = \frac{1}{1 + \exp (-y\bs{\theta}^\top \mathbf{x})}$, whose value represents the estimated probability that an individual with covariates $\mathbf{x}$ has label $y$. The logistic loss is $-\log(\sigma(y\bs{\theta}^\top \mathbf{x}))$.}
  \label{fig:sigmoid}
\end{wrapfigure}Namely, we learn a linear model $\bs{\theta}^\top\mathbf{x}$ where a larger value indicates a higher probability that the true label of~$\mathbf{x}$ is $y = 1$.
The loss function for logistic regression is the logistic loss, also called the cross-entropy loss, which involves the sigmoid function and takes the form~$\ell(f_{\bs{\theta}}(\mathbf{x}^i), y^i) = \log(1 + \exp (-y^i(\boldsymbol{\theta}^\top\mathbf{x}^i)).$ We refer to Figure~\ref{fig:sigmoid} for an explanation of the logistic loss. Logistic regression is considered relatively interpretable provided that the number of features $d$ is sufficiently small~\citep{logisticregression}.

Similarly to least squares regression, a penalty $\lambda_0\|\bs{\theta}\|_0$ can be added to logistic regression to encourage sparsity by selecting only a few non-zero coefficients. However, while the logistic loss is convex, it is also significantly harder to handle than quadratic losses in the context of branch-and-bound algorithms, limiting the effectiveness of MIO methods.~\citet{sato2016feature} present an MILO formulation which uses a piecewise linear convex approximation of the logistic loss in conjunction with big-M constraints to model the~L0 sparsity penalty (as described in section~\ref{sec:subsetselect_lr}). If a separable regularization term is present in the objective function of the logistic regression, then perspective methods similar to those reported in section~\ref{sec:subsetselect_lr} can be used, e.g., see~\cite{deza2022safe}. Alternatively, researchers have recently started to consider other relaxations for sparse logistic regression involving more 
sophisticated convexification techniques \cite{shafiee2024constrained}. Nonetheless, effort towards solving sparse logistic regression is relatively recent and current MIO technology is less mature than methods for solving least square regression problems \eqref{eq:l0l2_regress}. To date, researchers have reported the ability to solve problems with $d$ in the hundreds. Naturally, further improvements may be 
achieved in a near future, but for now the best approaches in high-dimensional settings may be primal methods that do not guarantee optimality \cite{dedieu2021learning}.

\begin{wrapfigure}{r}{0.4\textwidth}
  \begin{center}
    \includegraphics[width=0.4\textwidth]{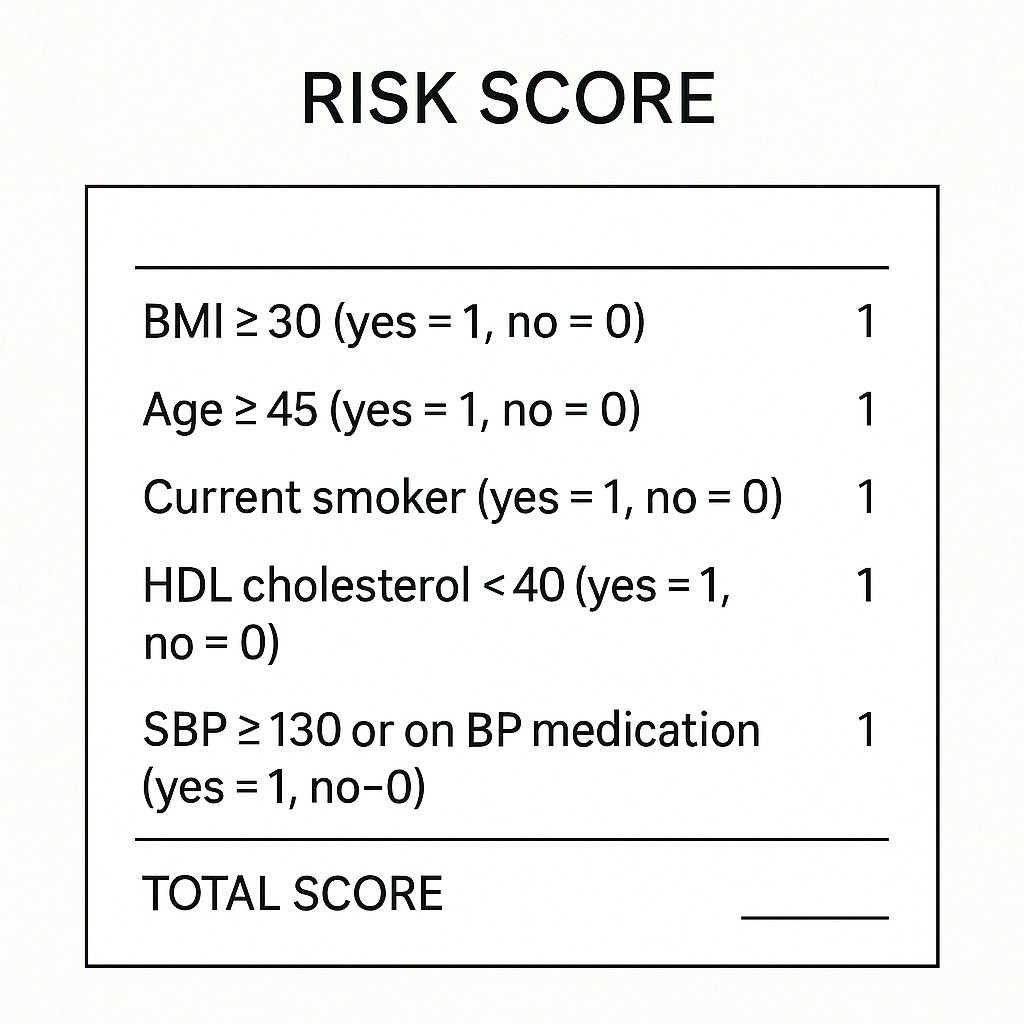}
  \end{center}
  \caption{A possible scoring card to assess medical risk. See \cite{ustun2019learning} for samples of optimal risk scoring cards in a variety of high-stakes domains.}
  \label{fig:risk}
\end{wrapfigure}

Sparse logistic regression can also be used as a basis for learning \textit{risk scores}, see Figure~\ref{fig:risk} for an example. A risk score is defined by a small set of questions with integral answers (in the example, answers are binary corresponding to yes/no questions); a decision-maker can then quickly add all the answers to obtain a score representing a prediction or risk of a patient. 
Traditionally, risk scores were designed by human experts, but there has been recent work in learning risk scores optimally from data. Indeed, 
a risk scoring scheme can be learned by solving the following problem using the logistic 
loss:
\begin{equation}\label{eq:risk_score}
    \begin{aligned}
        \min_{\boldsymbol{\theta} \in \mathbb{Z}^d,  \mathbf{z}\in \{0,1\}^d} \ & \sum_{i\in \mathcal{I}} \left[ \log(1 + \exp (-y^i(\boldsymbol{\theta}^\top\mathbf{x}^i))\right] + \lambda_0 \sum_{j=1}^d z_j \\
        \text{s.t.} \hspace{7.5mm}
        &lz_j\leq \theta_j\leq u z_j  \hspace{10mm} \forall j \in [d],
    \end{aligned}
\end{equation}
where $l$ and $u$ are lower and upper bounds on the potential answers. The two main
differences between sparse logistic regression and risk scores learning are: \textit{(a)} variables $\bs{\theta}$ are restricted to be integer, as the addition of integer numbers by hand is substantially easier than adding rational or real numbers; \textit{(b)} variables $\bs{\theta}$ are bounded, as addition of small integer numbers is easier than large ones. 
\citet{ustun2019learning} 
use MILO-based approaches to solve~\eqref{eq:risk_score}, constructing outer approximations of the logistic loss -- note that the approach differs from \citet{sato2016feature} in the sense that the linear outer approximation is dynamically constructed within the branch-and-bound tree, instead of being defined a priori. The authors report solutions to problem with $d$ in the dozens -- sufficient to tackle several problems in high-stakes domains, but this approach would not scale to large-dimensional problems. We point out that since the release of that paper, several general-purpose off-the-shelf solvers support
branch-and-bound algorithms with exponential cones, and can directly tackle \eqref{eq:risk_score} without requiring tailored implementations of outer approximations. \citet{molero2024optimal} generalize the approach to continuous data $\{\mathbf{x}^i\}_{i\in \mathcal{I}}$, although this generalization results in much more difficult problems: exact solutions were obtained by the authors in instances with $d\leq 5$, and heuristics are proposed for larger dimensions.

\subsubsection{Support Vector Machines}\label{sec:svm}
\begin{wrapfigure}{r}{0.45\textwidth}
  \begin{center}
\includegraphics[width=0.45\textwidth]{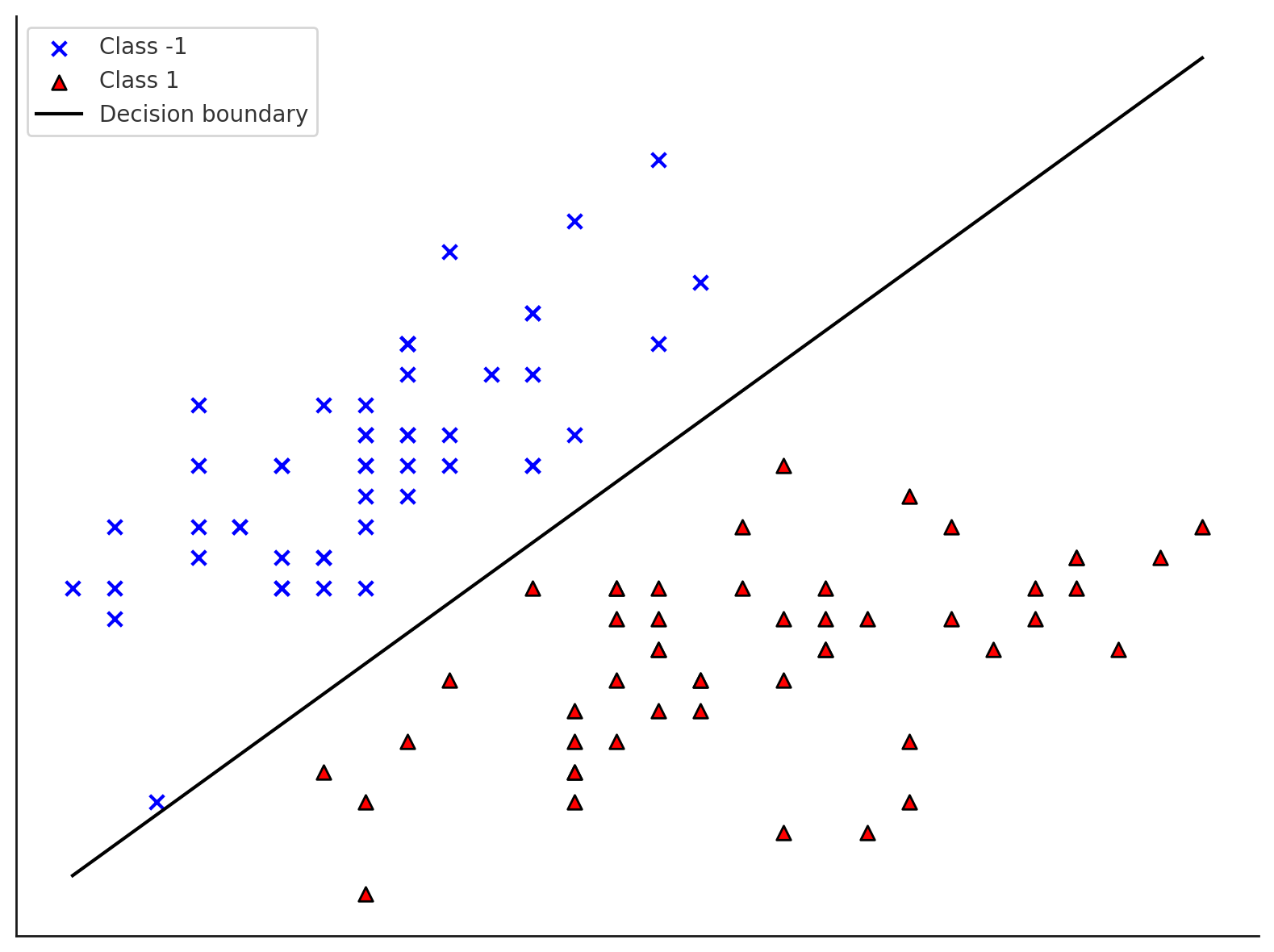}
  \end{center}
  \caption{A SVM classifier on perfectly separable data.}
  \label{fig:SVM}
\end{wrapfigure}

Given data with possible labels $\mathcal{Y}=\{-1,1\}$, support vector machines 
(SVMs) are 
classification models that use a separating hyperplane~$f_{\bs{\theta}} (\mathbf{x})$ of the form~\eqref{eq:add_ml} to divide the space of covariates~$\mathbf{x}$ into two, with each section corresponding to a possible label, see Figure~\ref{fig:SVM}. Requiring that points with different labels are on different sides of the hyperplane can be encoded via constraints $ \boldsymbol{\theta}^\top \mathbf{x}^i>0$ for all points such that $y^i=1$, and $ \boldsymbol{\theta}^\top \mathbf{x}^i<0$ for all points such that $y_i=-1$ -- or, concisely, as $y_i\cdot \boldsymbol{\theta}^\top \mathbf{x}^i\geq \epsilon$ for all $i\in\mathcal{I}$, where $\epsilon$ can be interpreted as a small number used to represent the strict inequality. However, multiple hyperplanes may separate the data equally well, and SVMs call for the one that is farthest from the data. Mathematically, this can be accomplished by choosing $\epsilon$ to be as large as possible while keeping the norm of $\bs{\theta}$ constant (as indeed, any vector of the form $c\bs{\theta}$ with $c\in \mathbb{R}$ represents the same hyperplane), or ensuring that $\bs{\theta}$ is of minimum norm while keeping $\epsilon$ constaint, e.g., $\epsilon=1$. Since data is rarely separable in practice, a \emph{hinge} penalty of the form $\max\{0,\epsilon-y_i\cdot \boldsymbol{\theta}^\top \mathbf{x}^i\}$ is added, which has value $0$ for correctly classified points but penalizes misclassifications proportionally to their distance to the hyperplane. Finally, similarly to previous subsections, a sparsity 
penalty can be added to enforce interpretability, resulting in the
optimization problem \cite{dedieu2021learning}
\begin{subequations}\label{eq:svm}
    \begin{align}
        \min_{\bs{\theta},\mathbf{z},\mathbf{r}} \ & \mathrlap{\frac{1}{2}\Vert\boldsymbol{\theta}\Vert_2^2 + \lambda_1\sum_{i\in \mathcal{I}} r_i + \lambda_0\sum_{j=1}^d z_j}\\
        \text{s.t.} \hspace{1.5mm} & \mathrlap{y^i\left( \boldsymbol{\theta}^\top \mathbf{x}^i\right) \geq 1 - r_i} & \forall i \in \mathcal{I}\label{eq:svm_hinge}\\
         & \mathrlap{-Mz_j \leq \theta_j \leq Mz_j} & \forall j \in [d]\\
        & \mathrlap{\boldsymbol{\theta} \in \mathbb{R}^d,\bs{z}\in \{0,1\}^d, \mathbf{r} \in \mathbb{R}_+^n,}
    \end{align}
\end{subequations}
where $\lambda_1$ and $\lambda_0$ are hyperparameters to be tuned and $r_i$ represents the value of the hinge penalty for datapoint $i\in \mathcal{I}$. 

Interestingly, while SVMs are among to most widely used models in statistics and ML, there are surprisingly limited results in the literature concerning solving their interpretable version \eqref{eq:svm} to optimality. The 
methods proposed by \citet{dedieu2021learning} focus mostly on first order methods that deliver high-quality solutions quickly, but no proof of optimality. The authors also showcase the performance of exact methods in some synthetic instances, but it is unclear how the results would extend to real data. \citet{guan2009mixed} propose to use perspective relaxations (similar to those discussed in section~\ref{sec:subsetselect_lr}) but report numerical issues when using off-the-shelf solvers -- note however that solvers have improved substantially in the~15 years since the publication of that paper. \citet{benitez2019cost} propose a method of learning sparse SVMs with pre-determined bounds on the true positive and true negative rates, solved with an MILO combined with a convex quadratic problem and evaluated on benchmark datasets. \citet{lee2022mixed} propose branch-and-bound algorithms for a related 
SVM problem with group sparsity, and they report solutions to problems with $d\approx 60$.


\subsubsection{Generalized Additive Models}\label{sec:gam}
Generalized additive models (GAMs) broaden the definition of~$f_{\bs{\theta}}(\mathbf{x})$ given in~\eqref{eq:add_ml}, allowing the elements of~$\mathbf{x}$ to be transformed by potentially nonlinear functions. For example, given a predefined class of one-dimensional functions~$f_k:\mathbb{R}\to\mathbb{R}$ indexed by set $\mathcal{K}$, we can write the GAM
\begin{equation}\label{eq:gam_pairwise}
    f_{\bs{\theta}}(\mathbf{x}) = \sum_{j=1}^d\sum_{k\in \mathcal{K}}\theta_{jk} f_k(x_j),
\end{equation}
where $\bs{\theta}$ are the parameters to be estimated. Unlike the previous additive models, GAMs are useful for capturing more complex and nonlinear relationships while still preserving a simpler structure for interpretability. \citet{navarro2023feature} consider sparse least squares problems where functions $f$ are given by splines, and \citet{cozad2014learning} use elementary functions compatible with the solver Baron for a downstream decision-making task. Another usual form of GAM is obtained by adding pairwise interactions of the form 
\begin{equation}\label{eq:gam}
    f_{\bs{\theta}}(\mathbf{x}) =  \sum_{j=1}^d\sum_{l=1}^d\theta_{ij}x_ix_j.
\end{equation}
see \cite{ibrahim2024grand,wei2022ideal} for methods in the context of sparse least squares problems. 
While exisiting approaches have been mostly proposed for least squares regression problems, the approaches in sections \ref{sec:logreg_riskscore} and \ref{sec:svm} can also be naturally extended to account for GAMs. 

Note that a direct implementation of GAMs may result in more complex, less interpretable models. For example, a product of two features may not have a clear meaning to a human user. Thus, imposing a sparsity constraint on vector $\bs{\theta}$ in the associated learning problem may not be sufficient to guarantee simple statistical models. Thankfully, the modeling power of MIO allows for a seamless inclusion of alternative logical considerations that can even lead to greater interpretability than simple additive models. 

\begin{figure}[!h]
\begin{center}
\subfloat[Hierarchical representation of categorical variable]{\includegraphics[width=0.49\textwidth]{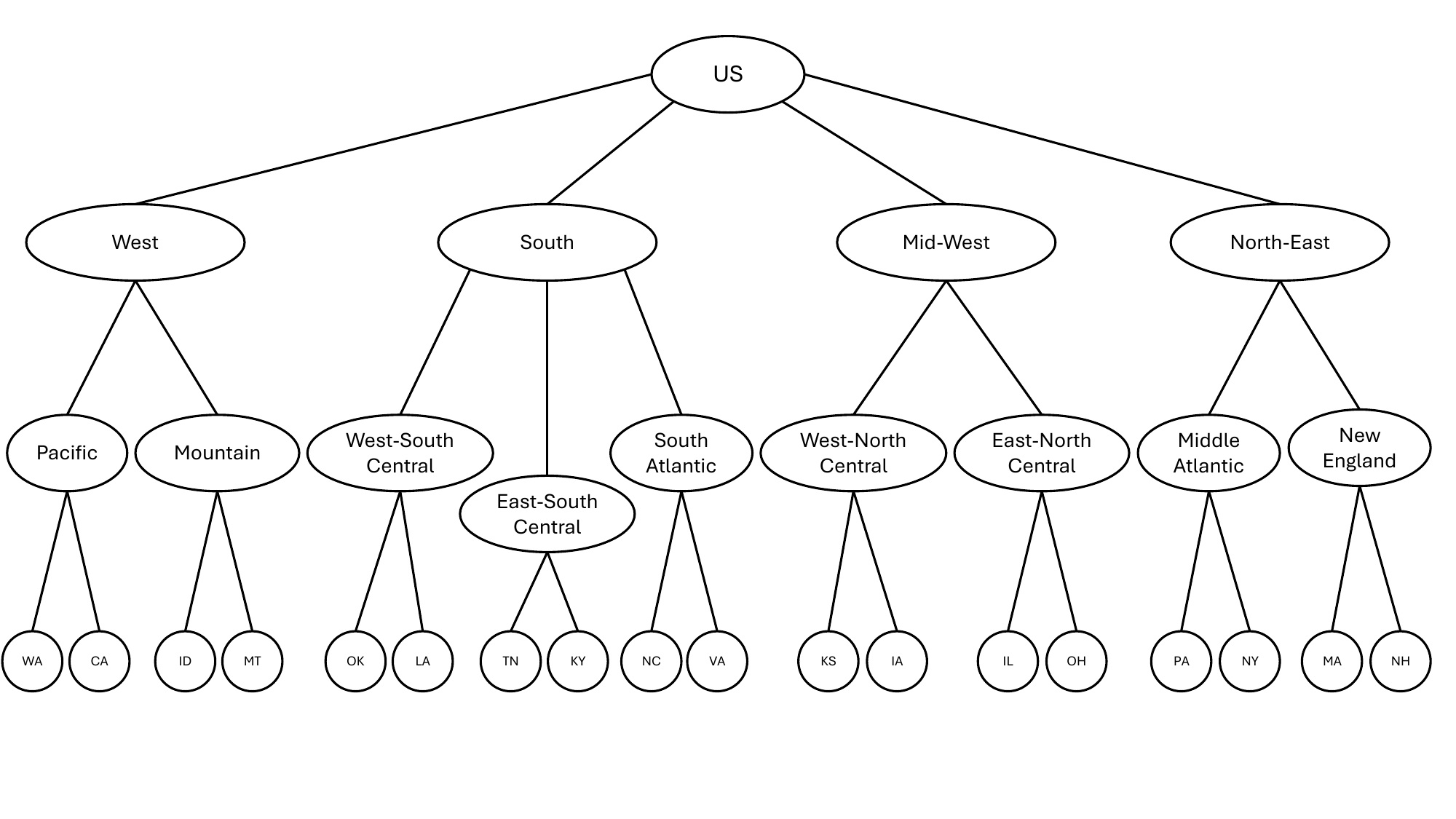}}\hfill \subfloat[Potential solution with nine regression coefficients]{\includegraphics[width=0.49\textwidth]{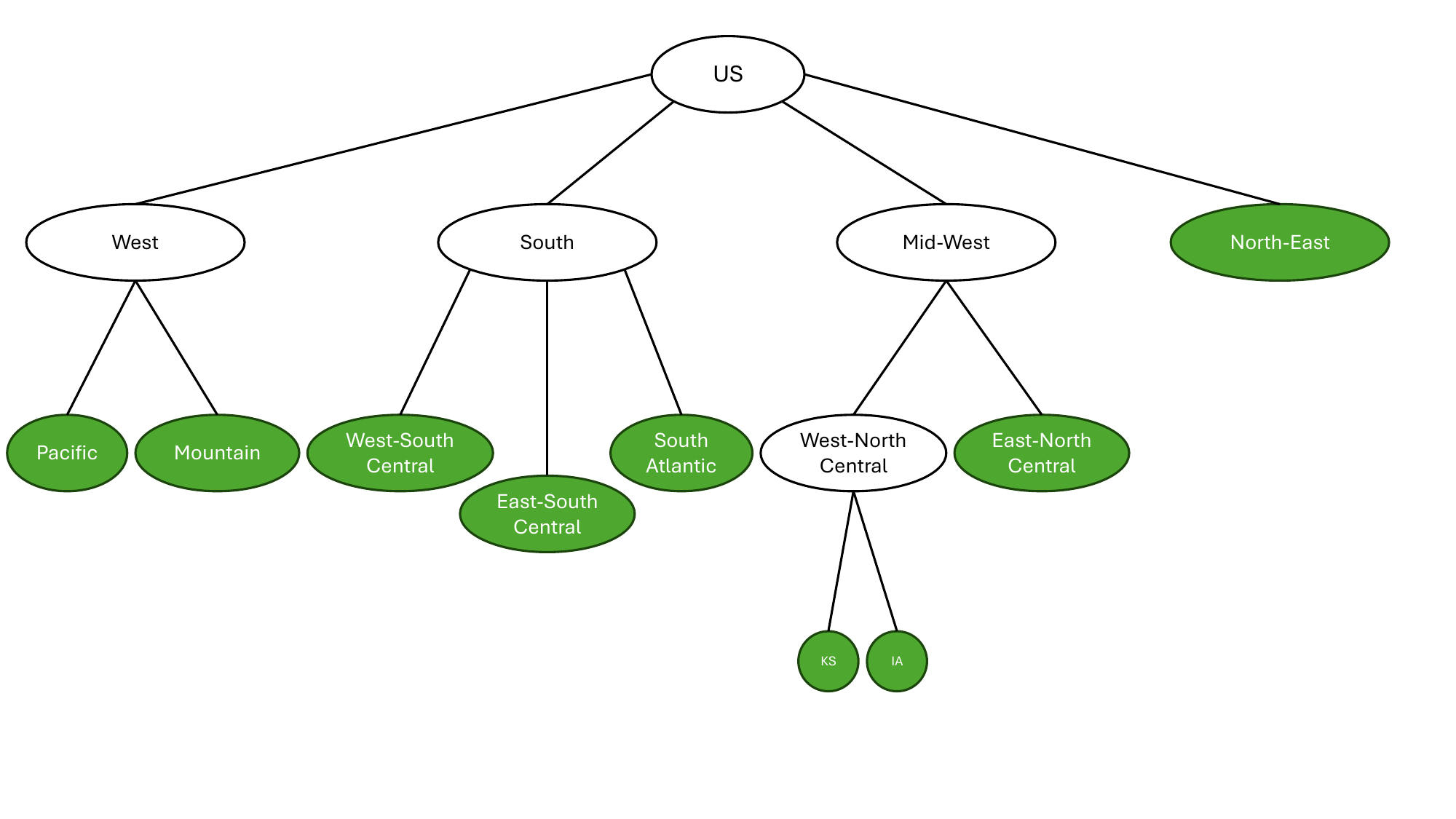}}

\caption{Visualization of \emph{geography} variable in the \emph{cancer trials} dataset, recreated from \cite{carrizosa2022tree}. At its most granular state, it consists of a categorical variables with 51 possible values, corresponding to all states in the US. }
\label{fig:geography}
\end{center}
\end{figure}

One such method is proposed in \cite{carrizosa2022tree} for learning problems involving categorical features that can be represented by a hierarchical structure. We illustrate their approach with the example from Figure~\ref{fig:geography}. Figure~\ref{fig:geography}a illustrates a single categorical variable, \emph{geography}, with~51 possible values. Standard techniques call for one-hot encoding of this feature, that is, adding~51 binary features $\{x_i\}_{i=1}^{51}$, each indicating whether a given datapoint belongs to  a particular state or not: additive models based on this one-hot encoding may not be interpretable, requiring~51 regression coefficients to encode \emph{geography}. However, in this context, products of certain features may have clear meanings, as for example the product of binary variables associated with states WA and CA indicates whether the datapoint is located in the Pacific region or not. A simpler encoding of \emph{geography} is shown in Figure~\ref{fig:geography}b, requiring only nine regression coefficients.

To produce an interpretable model that encodes such categorical variables, artificial features associated with all possible values and subcategories are added to the model.
Let $\mathcal{T}$ be a tree encoding the hierarchical relationships between subcategories and values -- note that the original values of the variable correspond to leaves of this tree. Let $\mathcal{V}$ denote the vertices of $\mathcal{T}$, let $\mathcal{L}\subseteq \mathcal{V}$ denote its leaves, and for every $j\in \mathcal{L}$ let $\mathcal{P}_j\subseteq \mathcal{V}$ be the vertices in the unique path from the root to $j$. 
Similarly to sparse regression models, constraints~$-Mz_j\leq \theta_j\leq Mz_j$ for $j\in \mathcal{V}$ are used. In addition, constraints
$$\sum_{j\in \mathcal{P}_j}z_j\leq 1\qquad \forall j\in \mathcal{L}$$
are added, ensuring that no two categories representing the same value are selected. As shown in \cite{carrizosa2022tree}, the output of this GAM is more interpretable than the one obtained by simply one-hot encoding the categorical feature, with little loss in accuracy.

\subsection{Logical Models for Supervised Learning}\label{sec:logic}
In this section, we explore how MIO is useful for creating optimal and interpretable logical models. Logical models use ``if-then,'' ``or,'' and ``and'' relationships between statements to define~$f_{\bs{\theta}}(\mathbf{x})$. Thus, they can be easily be described via text and/or visuals, making them inherently interpretable.  Unlike linear models, logical models can capture non-linear relationships and naturally work well with categorical and multiclass datasets~\citep{rudin2022interpretable}. Typically, MIO-based logical models minimize the 0-1 loss~$\ell(f_{\bs{\theta}}(\mathbf{x}^i), y^i) = \mathbbm{1}[f_{\bs{\theta}}(\mathbf{x}^i) \neq y^i]$, which directly maximizes the training accuracy. These models often include some sparsity penalty that limits the number of statements that define~$f_{\bs{\theta}}$. In this section, we discuss three types of logical models: decision trees in section~\ref{sec:decision_tree}, and rule lists and decision sets in section~\ref{sec:decision_listset}.

\subsubsection{Decision Trees} \label{sec:decision_tree}
Decision trees are a popular interpretable logical model due to
its ease of visualization, and are used widely in applications. A classification tree is a model which takes the form of a binary tree. At each branching node, a test is performed, which directs a sample to one of its descendants. Thus, each data sample, based on its features, follows a path from the root of the tree to a leaf node, where a label is predicted. Because learning decision trees that optimize training accuracy is~$NP$-hard, there exist many heuristic approaches that locally optimize other metrics like Gini impurity or information gain~\citep{Breiman2017ClassificationTrees}.

\citet{bertsimas2017optimal} provide one of the first MILO formulations for optimal classification trees, where the tree is constrained by a pre-specified depth~$D$ and big-M constraints are used to specify the routing of datapoints in the tree. Following this work, there has since been a growing body of work to improve on such a formulation to improve computational tractability. For example,~\citet{verwer2017learning, verwer2019learning} propose to reduce the number of variables and constraints of~\citet{bertsimas2017optimal} in order to explore the branch-and-bound tree faster, albeit at the cost of weakening the continuous relaxation. Alternatively, \citet{aghaei2024strong} and \citet{gunluk2021optimal} propose stronger formulations for settings with binary or categorical data, without increasing the size of the MIO formulations.

\begin{wrapfigure}{r!}{0.32\textwidth}
  \begin{center}
\includegraphics[width=0.32\textwidth]{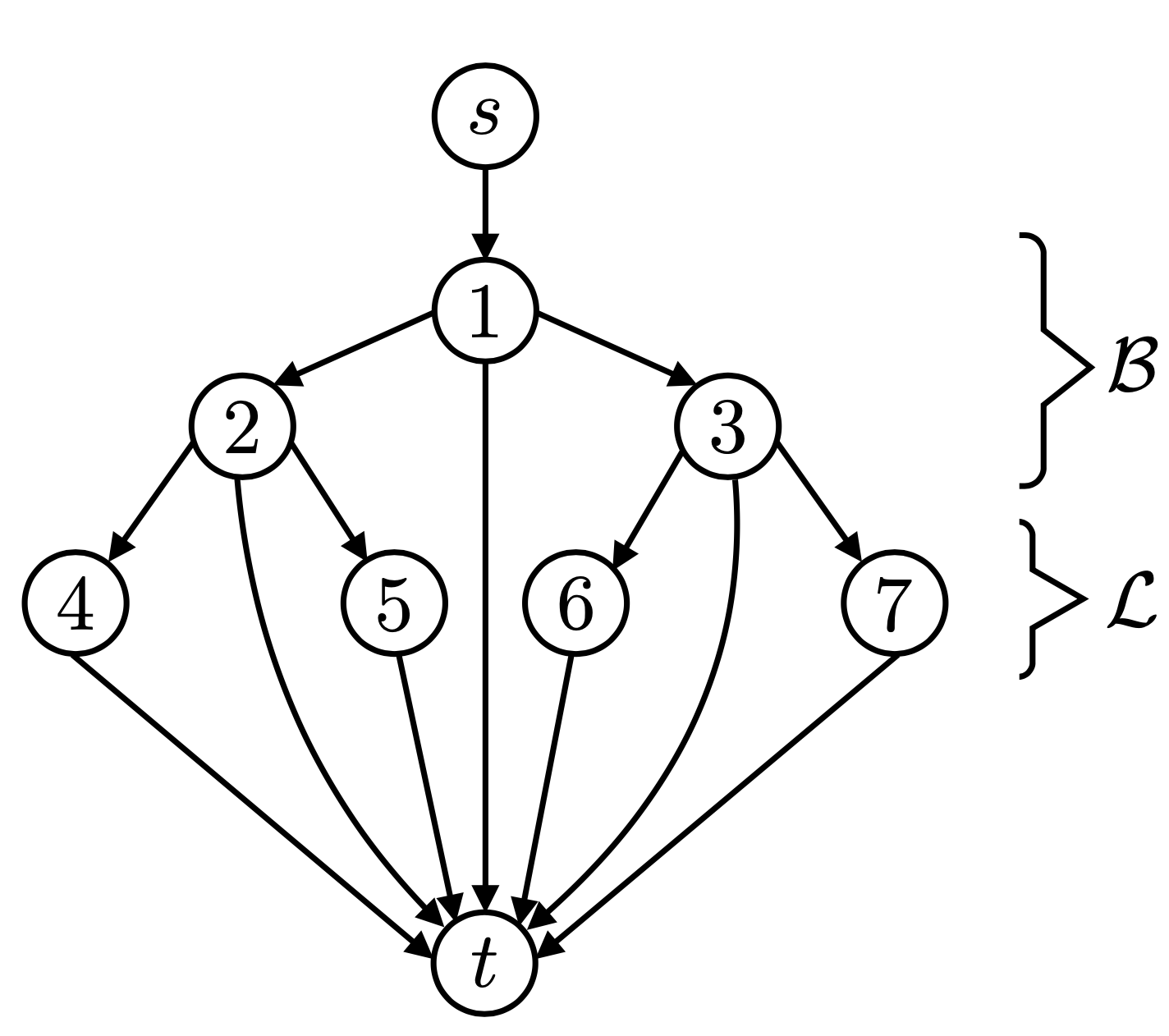}
  \end{center}
  \caption{A visualization of the flow network used to train a decision tree of depth $D=2$ in~\citet{aghaei2024strong}.}
  \label{fig:tree}
\end{wrapfigure}

We now discuss the approach of~\citet{aghaei2024strong}, which uses a flow-based formulation. 
We visualize the notation and flow network used in Figure~\ref{fig:tree}, and provide details of the formulation in the following. Consider a binary tree of maximum depth~$D$ where each node is indexed using breadth-first search from 1 to $2^D -1$. We let~$\mathcal{B} := \{1,\ldots,2^{D-1} - 1\}$ contain the indexes of all internal nodes and~$\mathcal{L} := \{2^{D-1}, \ldots,2^D - 1\}$ be the indexes of all leaf nodes. For node $k \in \mathcal{B}$, we denote the left child as $l(k) := 2k$, the right child as $r(k):=2k+1$, and the
parent node as $a(k)$. Also denote $\mathcal{A}(k) \subset \mathcal{B}$ as the set of all ancestors of $k$, i.e., the set of all nodes traversed between~$s$ and $k$. Assuming that the covariates are binarized in the training data (i.e., $\mathbf{x}^i \in \{0,1\}^d$ for all $i \in \mathcal{I}$), let $\theta_{k,j} \in \{0,1\}$ be the decision variable that equals~$1$ if and only if at node~$k \in \mathcal{B}$, feature $j \in [d]$ is tested on. Then if $x^i_j = 0$ where~$b_{k,j} = 1$, datapoint $i$ flows to $l(k)$, and otherwise is routed to $r(k)$. We let decision variable~$\phi_{k,y} \in \{0,1\}$ equal~1 if and only if node $k \in \mathcal{B \cup L}$ predicts label $y$. Note that we allow for internal nodes in $\mathcal{B}$ to make predictions, so any nodes below them can be removed.

We now create a flow network that will be used to formulate the MILO for learning optimal classification trees. We introduce source node $s$ and sink node $t$, where we add an edge from $s$ to $1$ as an entry point into the network and an edge from every node $k \in \mathcal{B \cup L}$ to the sink $t$. We introduce binary decision variables $z^i_{a(k), k} \in \{0,1\}$ for every datapoint $i \in \mathcal{I}$ and $k \in \mathcal{B \cup L}$ that equals 1 if and only if the $i$th datapoint is correctly classified in the tree and traverses the edge from $a(k)$ to $k$. We also define flow variables to the sink~$z^i_{k,t} \in \{0,1\}$ for all~$k \in \mathcal{B \cup L}$ and $i \in \mathcal{I}$ that equals 1 if and only if $\phi_{k,y^i} = 1$ and~$i$ visits node $k$ (i.e.,~$i$ is correctly classified at node $k$). We can minimize the 0-1 misclassification loss by maximizing the number of correct classifications, where $i$ is correctly classified if and only if~$\sum_{k \in \mathcal{B \cup L}} z^i_{k,t} = 1$. We then have the following formulation to learn an optimal classification tree:
\begin{subequations} \label{eq:flowtree}
    \begin{align}
        \max_{\bs{\theta},\mathbf{z},\bs{\phi}} \ & \sum_{i \in \mathcal{I}}\sum_{k \in \mathcal{B \cup L}} z^i_{k,t} \\
        \text{s.t.} \hspace{2.5mm} & \sum_{j =1}^d \theta_{k,j} + \sum_{y \in \mathcal{Y}} \phi_{k,y}+ \sum_{k' \in \mathcal{A}(k)}\sum_{y \in \mathcal{Y}} \phi_{k',y} = 1 & \forall k \in \mathcal{B} \label{eq:flowtree_univariate}\\ 
        & \sum_{y \in \mathcal{Y}} \phi_{k,y}+ \sum_{k' \in \mathcal{A}(k)}\sum_{y \in \mathcal{Y}} \phi_{k',y} = 1 &\forall k \in \mathcal{L} \label{eq:flowtree_leafpredict}\\
        & z^i_{a(k), k} = z^i_{k, l(k)} + z^i_{k, r(k)} + z^i_{k,t} & \forall k \in \mathcal{B} \label{eq:flowtree_conserve_branch} \\ 
        & z^i_{a(k), k} = z^i_{k, t} & \forall k \in \mathcal{L} \label{eq:flowtree_conserve_leaf}\\
        & z^i_{k, l(k)} \leq \sum_{j\in[d]:x^i_j = 0} \theta_{k,j} & \forall k \in \mathcal{B} \label{eq:flowtree_left} \\
        & z^i_{k, r(k)} \leq \sum_{j\in[d]:x^i_j = 1} \theta_{k,j} & \forall k \in \mathcal{B} \label{eq:flowtree_right}\\
        & z^i_{k,t} \leq \phi_{k,y^i} & \forall k \in \mathcal{B \cup L} \label{eq:flowtree_correct}\\
        & \theta_{k,j} \in \{0,1\} & \forall k \in \mathcal{B}, j \in [d] \\
        & \phi_{k,y} \in \{0,1\} & \forall k \in \mathcal{B \cup L}, y \in \mathcal{Y} \\
        & z^i_{a(k),k} \in \{0,1\} & \forall i \in \mathcal{I}, k \in \mathcal{B \cup L} \\
        & z^i_{k,t} \in \{0,1\} & \forall i \in \mathcal{I}, k \in \mathcal{L}
    \end{align}
\end{subequations}
Constraints~\eqref{eq:flowtree_univariate} ensure that for any node in $\mathcal{B}$, either only one feature is tested on, one label is predicted, or the node was removed (due to an ancestor predicting a label). Likewise, constraints~\eqref{eq:flowtree_leafpredict} ensure that for any node in $\mathcal{L}$, either one label is predicted or the node was removed.
We specify flow conservation constraints in~\eqref{eq:flowtree_conserve_branch} and~\eqref{eq:flowtree_conserve_leaf}. Constraints~\eqref{eq:flowtree_left} and~\eqref{eq:flowtree_right} ensure that datapoint $\mathbf{x}^i$ gets routed through the tree correctly based on the set of tests $\theta_{k,j}$. We also restrict $z^i_{k,t}$ to be 0 if leaf $k$ does not predict the correct label for datapoint~$i$ in constraint~\eqref{eq:flowtree_correct}. Since then, there have been several extensions and improvements over the aformentioned works \cite{ales2024new,hua2022scalable,liu2024optimal,d2024margin,zhu2020scalable,elmachtoub2020decision}. We refer to the review paper of~\citet{carrizosa2021mathematical} for further discussion on MILO for decision trees.


\subsubsection{Rule Lists and Decision Sets}\label{sec:decision_listset}
Rule lists (also called decision lists) and decision sets are logical models that are often expressed with text. 

\begin{figure}[!h]
\begin{center}
\subfloat[Rule List]{\includegraphics[scale=0.3]{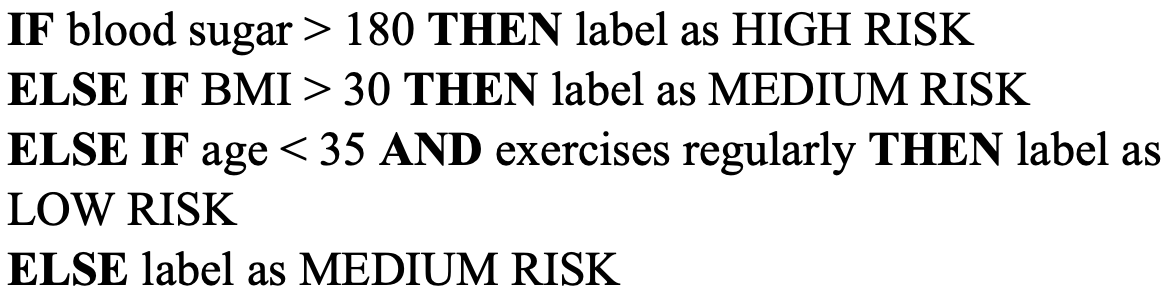} \label{fig:rule-list}} 
\subfloat[Decision tree representation of a rule list.]{\includegraphics[scale=0.4]{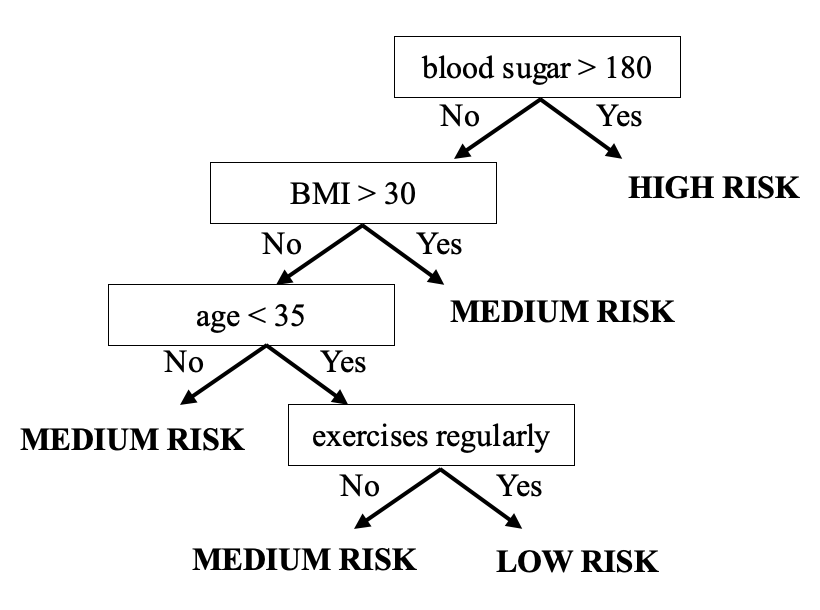}\label{fig:rule-list-tree}} \hfill
\subfloat[Decision Set]{\includegraphics[scale=0.3]{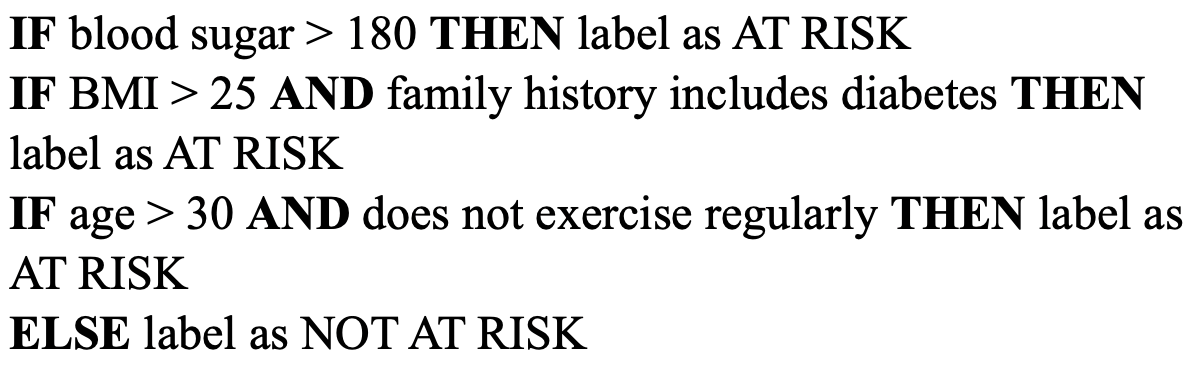}\label{fig:decision-set}}
\caption{A rule list and decision set example on predicting diabetes risk.}
\label{fig:ifelse}
\end{center}
\end{figure}

\paragraph{Rule Lists} Rule lists use ``if-then-else'' statements in succession. At each statement, if~$\mathbf{x}$ satisfies a conjunction of statements, then a prediction is made, else the next conjunction of statements is tested. At the last conjunction, a prediction is made. We provide an example of a rule list in Figure~\ref{fig:rule-list}. Rule lists can be expressed as a special case of binary decision trees, where each node that a test is performed has at least one child that predicts a class. We provide an example decision tree in Figure~\ref{fig:rule-list-tree} using the rule list in Figure~\ref{fig:rule-list}. Thus, one method of learning rule lists is to modify the MIO formulation for learning classification trees to require the tree to be expressible as a rule list. For example, building on  formulation~\eqref{eq:flowtree}, to learn a rule list, we can add a constraint that requires any internal node $k \in \mathcal{B}$ where a test is perform to have at least one of its children predict a class:
\begin{equation}\label{eq:rule_list_constraint}
    \sum_{j=1}^d \theta_{k,j} \leq \sum_{y \in \mathcal{Y}} \phi_{l(k),y} + \phi_{r(k),y} \hspace{5mm} \forall k \in \mathcal{B}.
\end{equation}

\citet{rudin2018learning} formulate an MILO for creating \textit{falling} rule lists, which is a rule list with monotonically decreasing probabilities of the positive class in each successive ``then''. 
They encode rule lists through a set of binary decision variables: variables to denote which ``if-then'' statements are selected, variables to denote the ordering of selected statements, and variables to track rule list predictions on the training datapoints.
Using some of the bounds introduced in~\citet{rudin2018learning},~\citet{angelino2018learning} improve on solve times with a branch-and-bound algorithm for learning optimal rules lists for categorical data.

\paragraph{Decision Sets} Decision sets are binary classification models in the form of statements in disjunctive normal form, i.e., if one from a set of ``and'' statements are satisfied, then one label is assigned; otherwise, another label is assigned.
Decision sets differ from decision trees and rule lists in that they are unordered and not hierarchical, expressible as a collection of ``if-else'' statements rather than nested statements. We provide an example of a decision set in Figure~\ref{fig:decision-set}. Decision sets can be expressed as a set of binary decision variables that indicate which conjunctions are used and what prediction is made for each training datapoint. Specifically, for a dataset with labels $\{-1,1\}$, we let $\mathcal{R}$ contain all possible conjunctions to consider, and let $\mathcal{K}^i \subset \mathcal{K}$ be the subset of conjunctions that datapoint $i$ satisfies. We then let $\theta_k \in \{0,1\}$ be a binary variable that indicates whether $k \in \mathcal{K}$ is a conjunction in the decision set. To keep track of the complexity of each conjunction, we let $c_k$ be a parameter that indicates the number of literals in conjunction $k \in \mathcal{K}$, and restrict the number of total literals used in the decision set by $\delta$ to enhance interpretability.
Letting $z^i \in \{0,1\}$ indicate whether~$i$ is correctly classified by the learned decision set, we minimize the 0-1 misclassification loss.~\citet{lawless2023interpretable} present the following MILO formulation for learning decision sets:
\begin{subequations} \label{eq:decision_set}
    \begin{align}
    	\max_{\boldsymbol{\theta} \in \{0,1\}^{|\mathcal{K}|}, \mathbf{z} \in \{0,1\}^n} \ & \sum_{i \in \mathcal{I}} z^i \label{eq:ds_obj} \\
    	\text{s.t.} \hspace{11mm} & \theta_k \leq 1-z^i & \forall i \in \mathcal{I}:y^i = -1, k \in \mathcal{K}^i \label{eq:ds_neg} \\
    	& \sum_{k \in \mathcal{K}^i} \theta_k \geq z^i & \forall i \in \mathcal{I}: y^i = 1 \label{eq:ds_pos} \\
    	& \sum_{k \in \mathcal{K}} c_k\theta_k \leq \delta. \label{eq:ds_complexity}
    \end{align}
\end{subequations}
The objective~\eqref{eq:ds_obj} maximizes the number of correct classifications. Constraint~\eqref{eq:ds_neg} states that for a datapoint $i$ with label $y^i = -1$, a misclassification occurs if the learned model has any conjunctions in $\mathcal{K}^i$. Constraint~\eqref{eq:ds_pos} states that for a datapoint $i$ with label $y^i = 1$, a misclassification occurs if the learned model has no conjunctions in $\mathcal{K}^i$. Constraint~\eqref{eq:ds_complexity} restricts the complexity of the decision set. Although the formulation has a linear objective and linear constraints, the number of possible conjunctions (and thus variables and constraints) grows significantly with the number of features in the dataset. Recent work has aimed to enhance the scalability of MILO formulations for decision sets using approximation schemes~\citep{wang2015learning} and column-generation-based methods~\citep{dash2018boolean, lawless2023interpretable, balvert2024iterative}.

\subsection{Interpretable and Explainable Unsupervised Learning}\label{sec:int_unsupervised}
We will now explore some transparent models for unsupervised learning tasks of the form~\eqref{eq:unsupervised}. In particular, we explore interpretable dimensionality reduction, graph learning and clustering techniques. \textit{Dimensionality reduction} methods learn a mapping~$f_{\bs{\theta}}(\mathbf{x}) : \mathbb{R}^d \mapsto \mathbb{R}^{d'}$ for~$d' < d$, which can be used to map~$\mathbf{x}$ to a lower dimensional space while maintaining as much information as possible. \textit{Graph learning} seeks to represent the joint distribution of a set of variables as a graph encoding conditional independence assumptions. \textit{Clustering} refers to methods that group values of~$\mathbf{x}^i$ in the data together based on some metric, learning a mapping~$f_{\bs{\theta}}(\mathbf{x})$ from the data to an assigned cluster. We discuss MILO-based methods for dimensionality reduction (specifically, principal component analysis) in section~\ref{sec:pca}, inference of Bayesian networks in section~\ref{sec:causal_discovery}, and clustering in section~\ref{sec:clustering}.

\subsubsection{Sparse Principal Component Analysis}\label{sec:pca}

Principal component analysis (PCA)
\begin{wrapfigure}{r}{0.4\textwidth}
  \begin{center}
\includegraphics[width=0.4\textwidth, trim={1.3cm 1.3cm 1.3cm 1.3cm},clip]{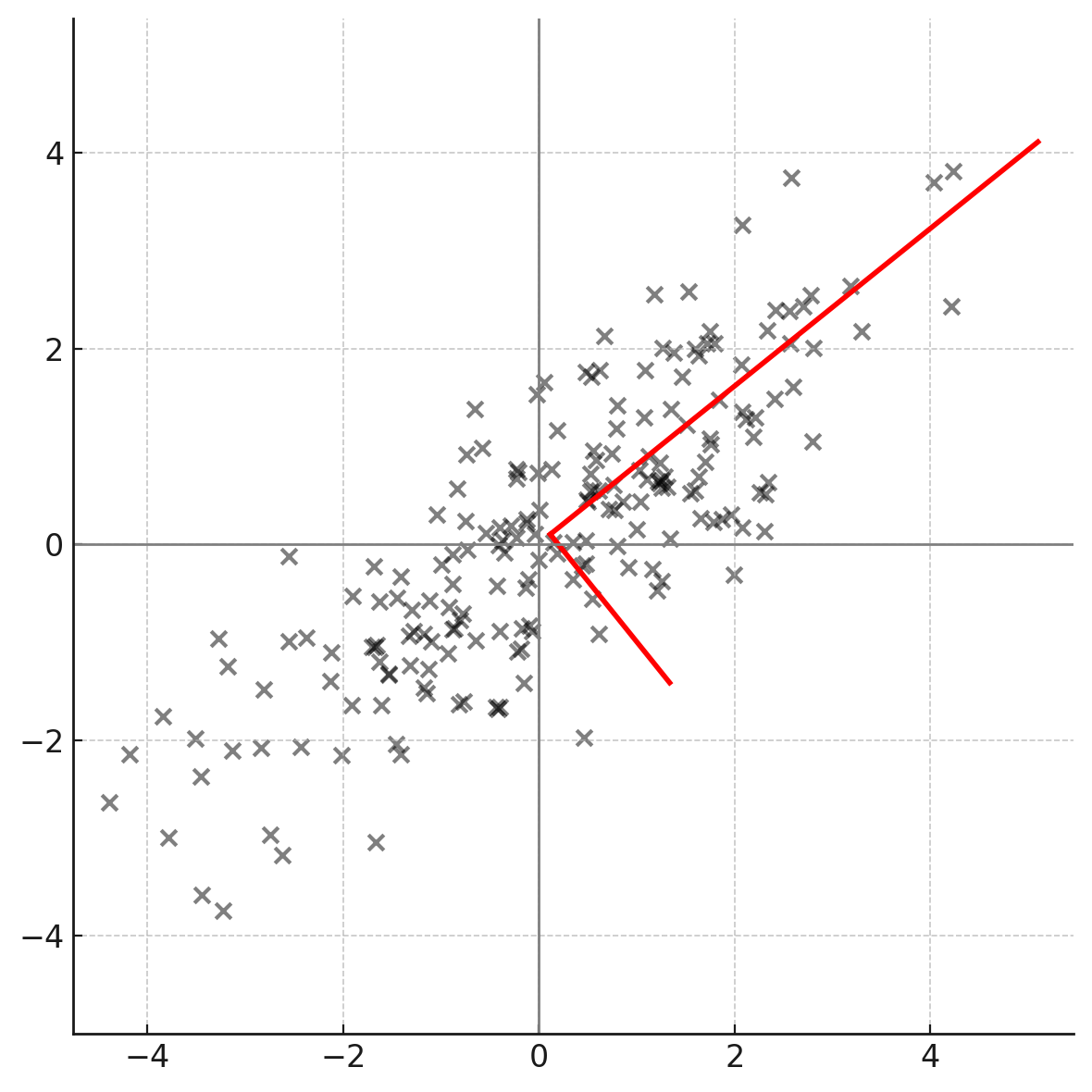}
  \end{center}
  \caption{2D dataset and principal components, where lengths represent the amount of variance captured.}
  \label{fig:pca}
\end{wrapfigure}
 is a classical technique for dimensionality reduction in unsupervised learning. At a high level, PCA seeks the most important axes or directions, called \emph{principal components}, that explain the variation in the data. In particular, 
the first principal component $\bs{\theta}\in \mathbb{R}^d$ is the vector such that $f_{\bs{\theta}}(\bs{x})=\bs{\theta^\top x}$ has the largest variance, which can be used for example to create a linear rule that serves to best differentiate the population. Consider the example shown in Figure~\ref{fig:pca}: if the data is projected into its first principal component (oriented up and right), an accurate summary of the data can be obtained requiring only one value instead of two. 

Mathematically, the first $k$ principal components correspond to the eigenvectors of $\mathbf{X^\top X}$ (sample covariance matrix of the training data) associated with the $k$ largest eigenvalues, where the matrix of training samples $\mathbf{X} \in \mathbb{R}^{n \times d}$ is assumed to be centered. While they can be computed easily, the principal components are typically dense and often not interpretable.  Similarly to the methods discussed in section~\ref{sec:additive}, interpretability can be enforced by requiring the components to be sparse via an L0 constraint. Formally, the problem of finding the first sparse principal component can be formulated as the MINLO
\begin{subequations} \label{eq:spca_misdo}
    \begin{align}
        \max_{\mathbf{z} \in \{0,1\}^d, \bs{\theta} \in \mathbb{R}^d} \ & \frac{1}{n}\mathbf{\bs{\theta}^\top \mathbf{X^\top X} \bs{\theta}} \\
        \text{s.t.} \hspace{8mm} & \|\bs{\theta}\|_2^2 \leq 1 \label{eq:spca_misdo_norm}\\
        & -z_j \leq \theta_j \leq z_j \hspace{10mm} \forall j \in [d]\label{eq:spca_misdo_bigM}\\
        & \sum_{j=1}^d z_j \leq \delta.
    \end{align}
\end{subequations}
Observe that the norm constraint \eqref{eq:spca_misdo_norm} implies that $|\theta_j|\leq 1$ for all $j\in [d]$. Therefore, \eqref{eq:spca_misdo_bigM} is in fact a big-M constraint using the implied bound of one. If this big-M constraint is removed, essentially making the discrete variables redundant since $\bs{z}=\bs{0}$ is optimal, then~\eqref{eq:spca_misdo} reduces to a well-known characterization of the largest eigenvalue of matrix $\mathbf{X^\top X}$. In other words,~\eqref{eq:spca_misdo} seeks a sparse approximation of the maximum eigenvalue. Observe that \eqref{eq:spca_misdo} computes only a single sparse component, and the problem needs to be resolved iteratively to obtain several components.

The continuous relaxation of problem~\eqref{eq:spca_misdo} is non-convex, as the objective corresponds
to the \emph{maximization} of a convex function. While several general purpose MINLO solvers now accept problems with non-convex quadratic relaxations, such a direct approach is currently not particularly effective. Instead, note that for any given $\mathbf{z}\in \{0,1\}^d$, problem \eqref{eq:spca_misdo} reduces to a thrust region problem (that is, optimization of a quadratic function subject to a ball constraint), a class problems where the standard Shor semidefinite optimization (SDO) relaxation is exact. The Shor relaxation linearizes the objective by introducing a matrix variable $\bs{T}$ such that $T_{ij}=\theta_i\theta_j$; in matrix form, we can write $\bs{T}=\bs{\theta\theta^\top}$. With the introduction of this matrix variable, the objective $\sum_{i=1}^d\sum_{j=1}^d (\bs{X^\top X})_{ij}T_{ij}$ is linear in $\bs{T}$, and all non-convexities are subsumed into the equality constraint $\bs{T}-\bs{\theta\theta^\top}=\bs{0}$. The equality constraint is then relaxed as $\bs{T} -\bs{\theta\theta^\top}\in \mathcal{S}_+^d\Leftrightarrow \begin{pmatrix}1 & \bs{\theta^\top}\\\bs{\theta}&\bs{T}\end{pmatrix}\in \mathcal{S}_+^{d+1}$, where $\mathcal{S}_+^{d+1}$ denotes the cone of positive semidefinite matrices.

In a celebrated paper,~\citet{d2004direct} present a SDO approximation of \eqref{eq:spca_misdo}, where the original variables $\bs{\theta}$ are projected out and additional constraints on the matrix variable $\bs{T}$ are imposed. \citet{li2024exact} and \citet{kim2022convexification} further improve on this relaxation using MIO techniques, resulting in continuous relaxations with optimality gaps inferior to 1\% in practice. In a similar vein, ~\citet{moghaddam2005spectral} and~\citet{berk2019certifiably} propose algorithms for solving~\eqref{eq:spca_misdo} to optimality using branch-and-bound. Nonetheless, solving SDO problems is itself a challenging task, let alone mixed-integer SDOs: exact methods scale to problems with $d$ in the low hundreds. Therefore, there have also been approaches that approximate sparse PCA problems using MIO methods \cite{dey2022using,wang2020upper}.

\subsubsection{Bayesian Networks}  \label{sec:causal_discovery}
\textit{Bayesian networks} are critical tools in the field of causal
\begin{wrapfigure}{r}{0.35\textwidth}
  \begin{center}
\includegraphics[width=0.35\textwidth, trim={1.3cm 1.3cm 1.3cm 1.3cm},clip]{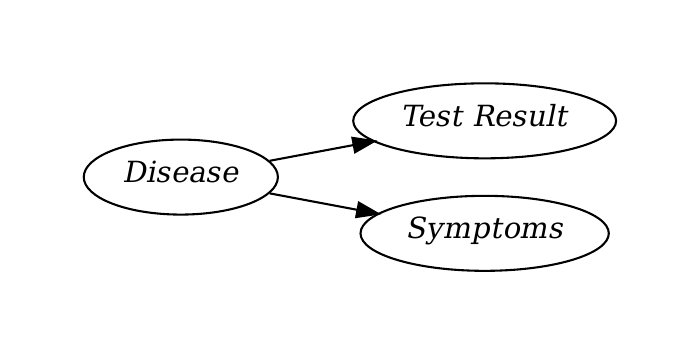}
  \end{center}
  \caption{Sample Bayesian network showing that the variables ``Test Result'' and ``Symptoms'' are conditionally independent given ``Disease''.}
  \label{fig:bayesian}
\end{wrapfigure}
inference and knowledge discovery. They aim to describe the joint probability distribution generating the data by means of a directed acyclic graph (DAG) $\mathcal{G}=(\mathcal{V},\mathcal{A})$. The vertex set $\mathcal{V}=[d]$ corresponds to the features in the data, while each arc $(j,k)\in \mathcal{A}$ encodes conditional independence relations in the data: specifically, each node is conditionally independent of its non-descendants given its parents. Figure~\ref{fig:bayesian} depicts a simple Bayesian network: if the status of a disease is known, the symptoms experienced by the patient and test results are independent, while being clearly correlated if the patient is not aware of whether they have a disease or not. 

Learning a Bayesian network can be formulated as an MIO. For all ordered pairs of features $(j,k)$, define a decision variables $z_{jk}$ to indicate whether $(j,k)\in \mathcal{A}$ ($z_{jk}=1$) or not ($z_{jk}=0$), and consider the optimization problem
\begin{subequations} \label{eq:bn}
    \begin{align}
        \min_{\bs{\theta}\in \Theta,\; \mathbf{z}\in \{0,1\}^{d\times d}}\ & \sum_{k\in \mathcal{V}} \sum_{i\in \mathcal{I}}\ell(f_{\bs{\theta},k}(\mathbf{x}^i\circ \mathbf{z}_{\cdot k}), x_k^i)+g(\bs{\theta},\mathbf{z})\\    \text{s.t.}\hspace{9mm}&z_{kk}=0\quad&\forall k\in \mathcal{V} \label{eq:bn_0}\\
        &\text{The graph $\mathcal{G}$ induced by $\mathbf{z}$ is acyclic}\label{eq:bn_cycle},
    \end{align}
\end{subequations}
where $\mathbf{z}_{\cdot k}\in \{0,1\}^d$ is the vector collecting all elements with second coordinate $k$, i.e.,~$(\mathbf{z}_{\cdot k})_j=z_{jk}$, and `$\circ$' refers to the Hadamard or entrywise product of vectors, so that~$\mathbf{x}^i\circ \mathbf{z}_{\cdot k}$ can be 
interpreted as the subvector of $\mathbf{x}^i$ induced by the nonzero elements of $\mathbf{z}_{\cdot k}$. The loss function~$\ell(f_{\bs{\theta},k}(\mathbf{x}^i\circ \mathbf{z}_{\cdot k}), x_k^i)$ indicates the prediction error of estimating $x_k^i$ using the other entries indicated by $\mathbf{z}$ with model $f_{\bs{\theta},k}$. Function $g$ is a function promoting regularization, for example~$g(\bs{\theta},\mathbf{z})=\sum_{j\in \mathcal{V}}\sum_{k\in \mathcal{V}}z_{jk}$ to ensure the graph is sparse and interpretable. Constraints~\eqref{eq:bn_0} ensures that no feature is used to estimate its own value, and constraints~\eqref{eq:bn_cycle}, which require an explicit mathematical formulation, ensure no cycles in the graph. 
Essentially, problem \eqref{eq:bn} attempts to design an acyclic graph such that the estimation of each feature based on the value of its parents is as accurate as possible. Note that if constraints~\eqref{eq:bn_cycle} are removed, then \eqref{eq:bn} reduces to $d$ independent versions of \eqref{eq:supervised} (potentially with additional sparsity constraints), but the acyclic constraints result in non-trivial couplings of these problems.

Two challenges need to be overcome to solve \eqref{eq:bn}. The first challenge is to find a tractable representation of the nonlinear and potential non-convex objective. \citet{jaakkola2010learning} considers the simplest case where functions $f_{\bs{\theta}}$ are separable, while several authors have considered more sophisticated cases where $f_{\bs{\theta}}$ are linear functions \cite{manzour2021integer,kucukyavuz2023consistent,park2017bayesian}. The second challenge is handling the combinatorial constraints \eqref{eq:bn_cycle}. A first approach is to use similar cycle breaking constraint as those arising in linear optimization formulations for spanning trees \cite{magnanti1995optimal} or MIO formulations for traveling salesman and vehicle routing problems \cite{laporte1987exact}:
\begin{equation}\label{eq:cycleBreak}\sum_{j \in S} \sum_{k\in S} z_{jk} \leq |S|-1 \qquad \forall S \subseteq \mathcal{V}.\end{equation}
Indeed, observe that cycles defined by nodes $S\subseteq \mathcal{V}$ involve exactly $|S|$ arcs: inequalities \eqref{eq:cycleBreak} thus prevent such cycles from happening. Constraints \eqref{eq:cycleBreak} typically result in strong relaxations, but the exponential number of constraints poses an algorithmic hurdle. In MILO settings, formulations with exponentially many constraints can typically be handled effectively \cite{bartlett2017integer}. Unfortunately, if modeling the loss functions requires nonlinear terms, then the technology is less mature and algorithms based on formulation \eqref{eq:cycleBreak} are not effective. Currently, a better approach is to introduce additional integer variables $\zeta_j\in \{1,\dots,d\}$ that can be interpreted as the position they appear in a topological ordering of the DAG. Then \eqref{eq:bn_cycle} can be reformulated as
$$1-d(1-z_{jk})\leq \zeta_k-\zeta_j\quad \forall j,k\in V,$$
which can be interpreted as follows: if $z_{jk}=0$ then the constraint is redundant, but if $z_{jk}=1$ then node $k$ is forced to have a later position than $j$ in the order. This alternative reformulation results in a substantially weaker relaxation, but requires only a quadratic number of constraints, resulting in relaxations that are easier to handle. Using these constraints, \citet{manzour2021integer} and \citet{kucukyavuz2023consistent} propose to tackle \eqref{eq:bn} for the case of least squares loss as 
\begin{subequations} \label{eq:bnLS}
    \begin{align}
        \min_{\bs{\theta},\mathbf{z},\bs{\zeta}}\ & \sum_{k\in \mathcal{V}} \sum_{i\in \mathcal{I}}\left(x_k^i-\sum_{j\in \mathcal{V}}\theta_{jk}x_j^i\right)^2+\lambda_0\sum_{j\in V}\sum_{k\in V}z_{jk}\\
\text{s.t.} \hspace{1.5mm} &\theta_{jj}=0\quad&\forall j\in V \\
        & -Mz_{jk}\leq \theta_{jk}\leq Mz_{kj} \quad&\forall i,j\in V\label{eq:bnLS_bigM}\\
        &1-d(1-z_{jk})\leq \zeta_k-\zeta_j\quad &\forall j,k\in V\\
        &\bs{\theta}\in \mathbb{R}^{d\times d},\; \mathbf{z}\in \{0,1\}^{d\times d},\;\bs{\zeta}\in \{0,1\}^d,
    \end{align}
\end{subequations}
where they also use perspective reformulations as described in section~\ref{sec:subsetselect_lr}. Observe that if $z_{jk}=0$, then constraints \eqref{eq:bnLS_bigM} ensure that $\theta_{jk}=0$ and therefore feature $j$ is not used in model that predicts feature $k$. 
Naturally, problem \eqref{eq:bnLS} is harder to solve than \eqref{eq:l0l2_soc} as it requires to estimate substantially more parameters, introduces many more variables and difficult constraints: the authors report solutions to problems with up to $d=100$ features, allowing for uses in some high-stakes domains such as healthcare. Other approaches that have used MIO to learn Bayesian Networks are \cite{chen2021integer,eberhardt2024discovering}.

\subsubsection{Interpretable clustering}\label{sec:clustering}
A clustering of a dataset $\{\mathbf{x}^i\}_{i\in \mathcal{I}}$ is a partition $\mathcal{C}_1\cup \mathcal{C}_2\cup\dots\cup \mathcal{C}_m=\mathcal{I}$, such that individuals assigned to the same group $\mathcal{C}_k$ are presumed to be similar, and dissimilar to individuals assigned to different clusters. Popular clustering methods in the ML literature such as K-means are not easily interpretable, as they do not provide a clear argument of why a given datapoint belongs to a given cluster. Recent works have proposed interpretable clustering methods by noting that logical methods for supervised learning, as described in section~\ref{sec:logic}, produce a partition of the data, e.g., the leaves of a decision tree. 

\citet{carrizosa2023clustering} propose to modify formulations such as \eqref{eq:flowtree} to design interpretable clustering methods as follows. Given a dissimilarity metric $d_{ij}$ between datapoints $i,j\in \mathcal{I}$, e.g., $d_{ij}=\|\mathbf{x}^i-\mathbf{x}^j\|$ for some given norm, and letting $z_{i}^k\in \{0,1\}$ be a binary variable that indicates datapoint $i$ is assigned to partition $k$, they use the objective function 
\begin{equation}\label{eq:clustering objective}
\min_{\mathbf{z}} \; \; \sum_{i\in \mathcal{I}}\sum_{j\in \mathcal{I}}\sum_{k\in \mathcal{B}\cup \mathcal{L}}d_{ij }z_{k}^iz_{k}^j.
\end{equation}
For example, in the context of decision trees, partitions correspond to leaves and variables $z_k^i$ in \eqref{eq:clustering objective} correspond to variables $z_{a(k),k}^i$ with $k\in \mathcal{L}$ in \eqref{eq:flowtree}. Note that \citet{carrizosa2023clustering} use logic rules instead of decision trees, but we adapt their approach to decision trees for notational simplicity. If two points are assigned to the same partition, then there is a penalty of $d_{ij}$ in the objective \eqref{eq:clustering objective}, while no penalty is incurred if assigned to different partitions. 

The objective function \eqref{eq:clustering objective} is nonlinear and non-convex as it involves products of binary variables. Nonetheless, there is a vast amount of literature associated with handling this class of functions -- we describe two of the most popular methods next. The first approach uses McCormick linearizations \cite{mccormick1976computability}: introduce additional variables $\zeta_{k}^{ij}$ representing the products $z_{k}^iz_{k}^j$, and add the linear constraints 
\begin{equation}\label{eq:mccormick}0\leq \zeta_{k}^{ij},\; -1+z_{k}^i+z_{k}^k\leq \zeta_{k}^{ij},\; \zeta_{k}^{ij}\leq z_{k}^i,\; \zeta_{k}^{ij}\leq z_{k}^j.
\end{equation}
Observe that for any $z_{k}^i,z_{k}^j\in \{0,1\}$, the only feasible solution to the McCormick system is indeed setting  $\zeta_{k}^{ij}=z_{k}^iz_{k}^j$. Linearization \eqref{eq:mccormick} is indeed the proposed approach in \cite{carrizosa2023clustering}. The second approach \cite{poljak1995convex} convexifies the nonlinear objective function as 
\begin{equation}\label{eq:clusteringNonlinear}
\min_{\mathbf{z}} \;\;  \sum_{i\in \mathcal{I}}\sum_{j\in \mathcal{I}}\sum_{k\in \mathcal{B}\cup \mathcal{L}}d_{ij }z_{k}^iz_{k}^j+\lambda\sum_{i\in \mathcal{I}}\sum_{k\in \mathcal{B}\cup \mathcal{L}}\left(\left(z_{k}^i\right)^2-z_{k}^i\right)
\end{equation}
with $\lambda>0$. Since for any $z_{k}^i\in \{0,1\}$ the identity $\left(z_{k}^i\right)^2=z_{k}^i$ holds and \eqref{eq:clusteringNonlinear} coincides with \eqref{eq:clustering objective} for integer solutions. However, the strength of the continuous relaxation degrades as $\lambda$ increases, since for any $0<z_{k,t}^i<1$ the inequality $\left(z_{k}^i\right)^2<z_{k}^i$ holds. Nonetheless, if $\lambda$ is sufficiently large, then \eqref{eq:clusteringNonlinear} is a convex quadratic function that can be handled effectively by MIQO solvers. Neither of these approaches is in general better than the other, since both degrade the relaxation quality (but in different ways), and while \eqref{eq:mccormick} requires introducing additional variables and constraints, \eqref{eq:clusteringNonlinear} introduces additional nonlinearities. Fortunately, commercial MIO solvers accept \eqref{eq:clustering objective} directly as an input and decide automatically which reformulation to use. 

Other approaches to design interpretable clusters based on more sophisticated metrics have been considered in the literature \cite{bertsimas2021interpretable,lawless2022interpretable}.
They consider using the Silhouette metric \cite{rousseeuw1987silhouettes}, defined as ratios involving the intracluster distances and distances to the second nearest cluster. However, this metric is highly non-convex, and the authors have thus resorted to heuristic approaches instead of exact methods.

\subsection{Counterfactual Explanations}\label{sec:counterfact_ex}
Counterfactual explanations (CEs) are a method of explaining black-box ML models through a statement like ``if this datapoint changed to this value, then this outcome would have been predicted instead''~\citep{lewis2013counterfactuals}. CEs are useful for explaining decisions that come from black box models, and provide actionables for individuals in the data on how to change their data to get a different outcome. For instance, a CE for some point~$\bar{\mathbf{x}}$ that received a class label $\bar{y}$ may be to find an~$\mathbf{x}$ close to~$\bar{\mathbf{x}}$ that received a desired class label~$y$. Mathematically, let some cost metric $c(\cdot, \cdot) : \mathbb{R}^d \times \mathbb{R}^d \mapsto \mathbb{R}$ encode information on how difficult it is to change~$\bar{\mathbf{x}}$ into~$\mathbf{x}$, where a larger value indicates an easier change.
For black-box model $f_{\bs{\theta}}$ that assigns class labels, we can find a CE~$\mathbf{x}$ for~$\bar{\mathbf{x}}$ by solving
\begin{equation} \label{eq:cf_explanation}
    \begin{aligned}
        \min_{\mathbf{x}} \ & c(\mathbf{x}, \bar{\mathbf{x}}) \\
        \text{s.t.} \hspace{1.25mm} & f_{\bs{\theta}}(\mathbf{x}) = y.
    \end{aligned}
\end{equation}
By solving problem~\eqref{eq:cf_explanation}, we obtain the value $\mathbf{x}$ that $\bar{\mathbf{x}}$ can most easily change to in order to get the desired outcome $y$. Finding the optimal $\mathbf{x}$ thus provides an explanation for the prediction of $\bar{\mathbf{x}}$ and the most feasible change to get to another outcome. The complexity of problem~\eqref{eq:cf_explanation} depends on the metric $c$ used, the encoding of $f_{\bs{\theta}}$ in the constraints, and additional domain requirements. For example, there may be constraints on which and how many features can change in practice, manifesting as sparsity constraints or penalities in~\eqref{eq:cf_explanation}. Such constraint can be modeled by introducing additional variables $\mathbf{z}\in \{0,1\}^d$, constraints $-Mz_j\leq x_j-\bar x_j\leq Mz_j$, and an additional penalty term $\lambda_0\sum_{j=1}^dz_j.$

\citet{cui2015optimal} and~\citet{parmentier2021optimal} use an MILO formulation to learn a counterfactual explanation for ensemble tree methods.~\citet{kanamori2020dace, kanamori2021ordered} propose MILO-based methods for~\eqref{eq:cf_explanation} that get CEs close to the empirical distribution of the data and account for interactions among features.~\citet{mohammadi2021scaling} provide CEs for neural network models using MILO, providing optimality and coverage guarantees in using their method.
Counterfactual explanations are not just useful for black-box models, but can also be useful for interpretable models.~\citet{ustun2019actionable} and~\citet{russell2019efficient} develop an MILO-based framework for counterfactual explanations in interpretable classification tasks like logistic regression, SVMs, and linearized rule lists and decision sets.~\citet{carreira2021counterfactual} learn CEs for multivariate decision trees, formulating the problem as an MIO and solving the integer variables by enumerating over tree leaves.

There are several recent works that 
extend beyond problem~\eqref{eq:cf_explanation}, and are able to generate multiple CEs at once.~\citet{carrizosa2024generating, carrizosa2024new} present MIQO formulations for learning CEs in score-based classifications (e.g., logistic regression and SVMs) in two settings: learning collective CEs across multiple individuals to help detect patterns and insights in CEs and learning CEs with functional data.~\citet{maragno2024finding} can also create robust CEs that are robust to small perturbations of the data, yielding a region of CEs instead of just one CE through robust optimization methods (see section~\ref{sec:robustness} on robustness) and MIO. We refer to~\citet{guidotti2024counterfactual} for more information on recent advancements in learning counterfactual explanations, and to~\citet{carrizosa2024mathematical} for an introduction to finding group CEs using MIOs.

\subsection{Hyperparameter Tuning}\label{sec:hyperparameter}

Most methods discussed in this section impose sparsity/interpretability via either a penalty term in the objective or a constraint. In both cases, hyperparameters controlling the importance of such interpretability terms need to be tuned carefully. Moreover, classical methods in the statistics/ML literature, involving a grid search over the space of hyperparameters and using cross-validation to select the best combination, may be prohibitive in the context of MIO methods due to the high cost associated with solving each training problem.
In some cases, the values of hyperparameters are determined due to operational constraints or by downstream decision-makers. For example, risk scores as depicted in Figure~\ref{fig:risk} are often meant to be printed in a physical scorecard, and space limitations in such scorecards impose a natural limit on the number of features used. In such cases, imposing sparsity via a hard constraint is preferable, as no parameter tuning is needed. 

More generally, it is possible to use metrics such as Aikake's \cite{akaike1974new} or the Bayesian information criteria \cite{schwarz1978estimating}, which provide guarantees on the out-of-sample performance of a model under suitable statistical assumptions. In some cases, the best model with respect to an information criterion reduces to a single MIO where the hyperparameters are determined a priori \cite{miyashiro2015subset}. In others, training the best model introduces additional nonlinearities; thankfully, specialized methods have been developed for such problems that are significantly faster than using grid search for cross-validation \cite{gomez2021mixed,miyashiro2015mixed}. Finally, recent works have studied how to efficiently perform cross-validation for MIO-based ML models without compromising quality, either by carefully exploring the hyperparameter space or \cite{cory2023stability} by designing parametric methods that allow for efficient reoptimization \cite{gomez2024real,bhathena2025parametric}. We point that the aforementioned approaches for efficient hyperparameter tuning of MIO ML models focus on sparse linear regression or similar variants, and more work for other classes of interpretable ML models is needed.

\section{Robustness} \label{sec:robustness}
Real-world datasets are often imperfect and deployment conditions can differ from training conditions. A robust ML model maintains reliable performance despite such discrepancies. 
In such settings, MIOs can be used to train a robust model, effectively handling challenges with discrete datasets and simultaneously preserving model interpretability and sparsity through the methods described in section~\ref{sec:interpretability}.
We explore how MIO can help train or verify models that are robust to three common data issues: outliers, adversarial attacks, and distribution shifts.
Outliers are anomalous observations not generated by the same process as the bulk of data, and can distort models if not handled. Adversarial attacks are deliberate, small perturbations to inputs designed to mislead the model. Distribution shifts are changes in the distribution of the data between training and deployment (e.g., due to evolving data collection or population drift).
This section discusses each of these challenges in turn and how to model and solve them with MIO.

\subsection{Outlier detection} \label{sec:lts}

\begin{figure}[h]
    \begin{center}
    \subfloat[Original signal]{\includegraphics[width=0.33\textwidth]{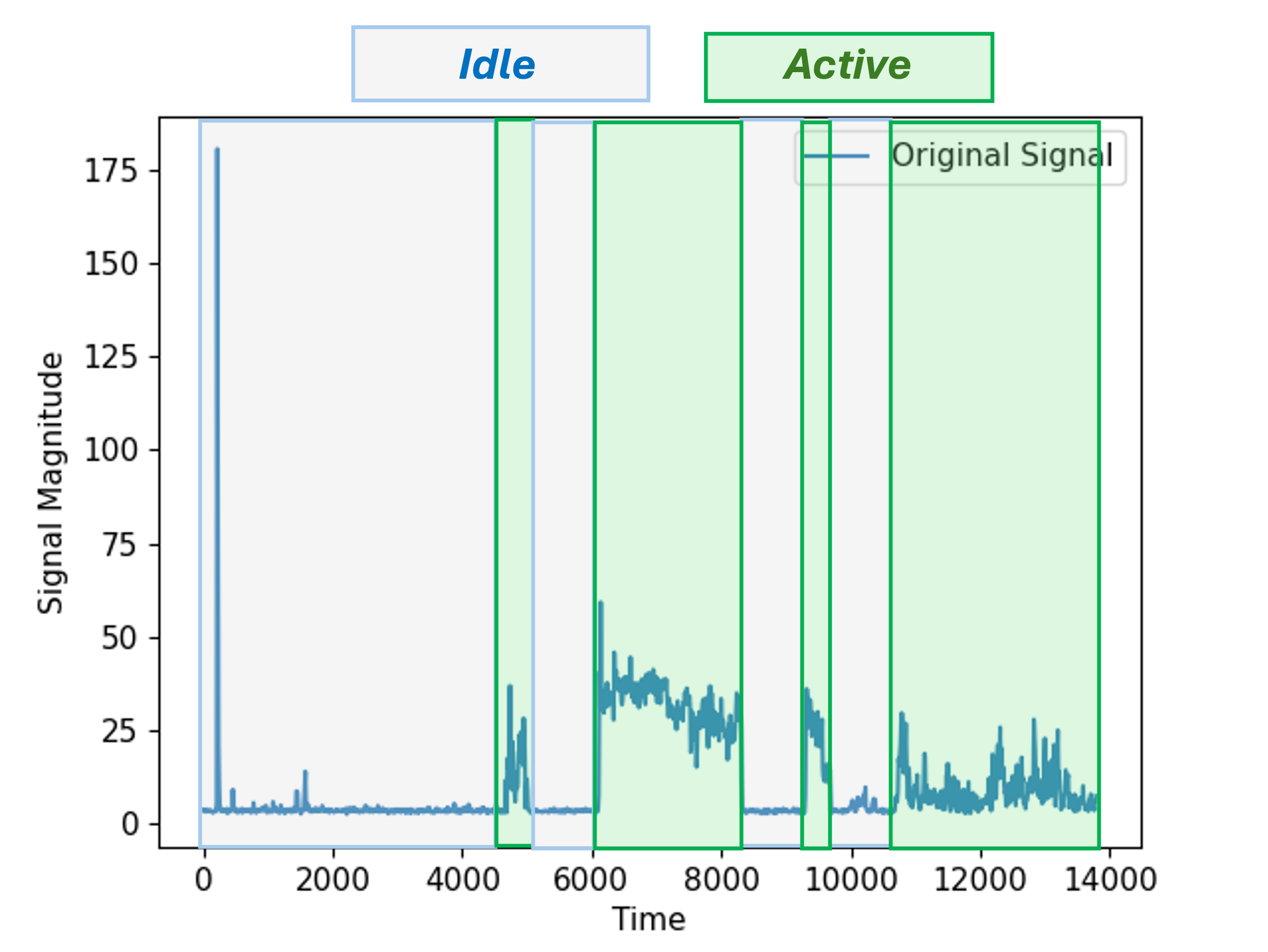}\label{fig::accelerometer_raw}}
    \subfloat[Exact outlier detection]{\includegraphics[width=0.33\textwidth]{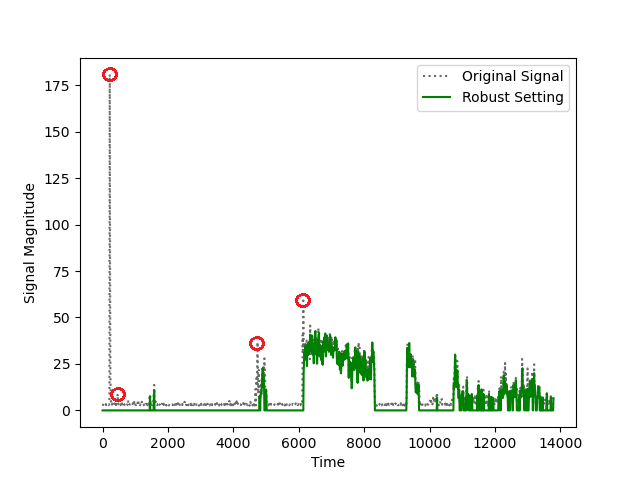}\label{fig::robust_outlier}}\newline
    \subfloat[$L_1$ relaxation]{\includegraphics[width=0.33\textwidth]{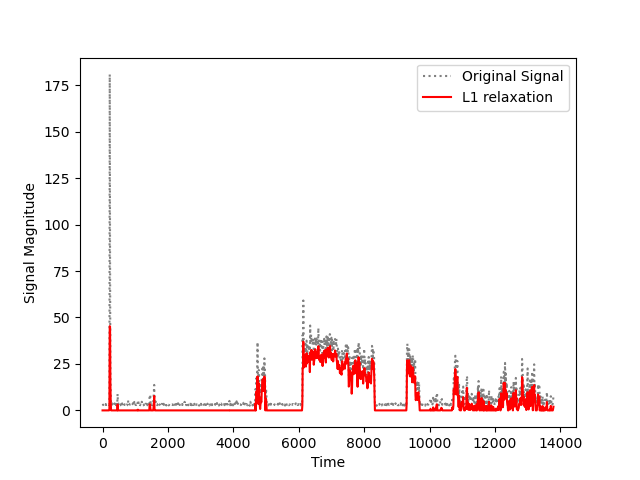}\label{fig::l1}}
    \hfill
    \subfloat[Wavelet denoising]{\includegraphics[width=0.33\textwidth]{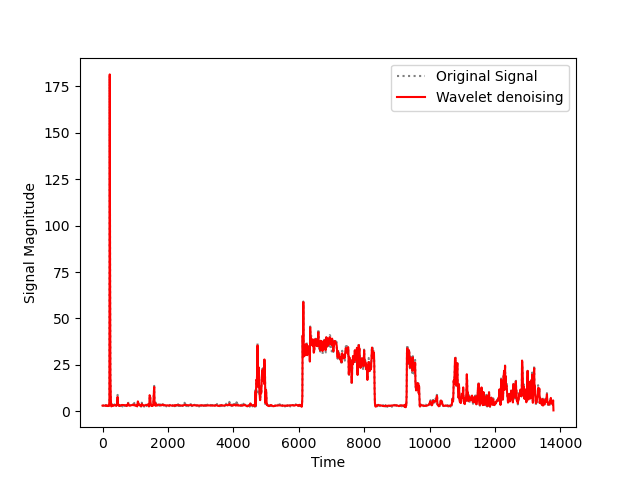}\label{fig::wavelet}}\hfill
    \subfloat[Low-pass filter]{\includegraphics[width=0.33\textwidth]{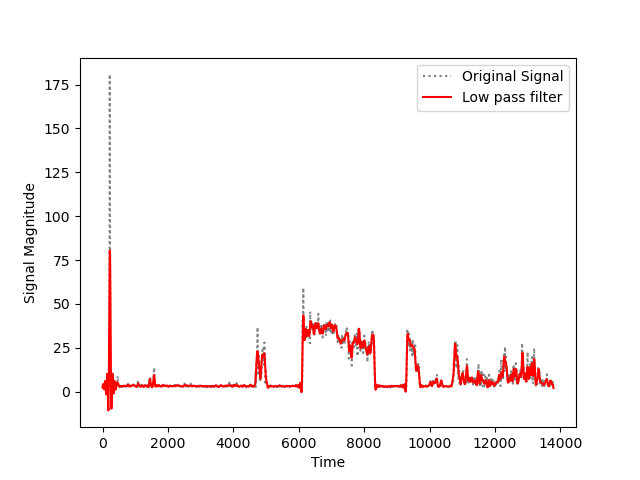}\label{fig::lowpass}}
    \caption{Data obtained from an accelerometer strapped to the chest, used in \cite{bhathena2025parametric}. The data is labeled, with ``true" activity periods shown in (a). The largest measurement recorded at the beginning of the time period is an outlier. Exact methods based on discrete optimization (b) are able to detect that outlier (and clearly mark anomalous points). In contrast, common denoising alternatives based on continuous optimization (c)-(e) fail to discard the outlier, and do not clearly identify any point as anomalous for further inspection.}
    \label{fig: outlier}
    \end{center}
    \end{figure}

Outliers can adversely affect model accuracy on the majority of data because many learning algorithms (like least squares) give every point equal influence in the objective. For example, Figure~\ref{fig: outlier}a shows real accelerometer data contaminated with an outlier observation. Failure to detect such outliers can immediately lead to incorrect statistical inferences and poor decisions.  To make linear models more robust to outliers, one approach is to limit or down-weight the influence of outlier points during training. A classical robust regression method is least trimmed squares (LTS), which fits an ordinary least squares regression while ignoring a certain fraction of the most extreme residuals \cite{rousseeuw1984least, rousseeuw2003robust}. In other words, LTS attempts to find model parameters~$\bs{\theta}$ that minimize the error on all but the worst $k$ datapoints (trimming off $k$ outliers). 

An optimization problem that simultaneously flags datapoints as outliers and fits a model with the rest of the training data can be formulated as an MIO by introducing binary variables to select which datapoints to exclude from the fit. Letting $z_i \in \{0,1\}$ indicate whether datapoint $i$ is discarded as an outlier ($z_i=1$) or included in the fit ($z_i=0$), an MINLO formulation is
\begin{equation}\label{eq:supervised_outliers}
\begin{aligned}
    \min_{\bs{\theta}\in \mathbb{R}^d,\mathbf{z}\in \{0,1\}^{n}} \;& \sum_{i \in \mathcal{I}} \ell(f_{\bs{\theta}}(\mathbf{x}^i), y^i)(1-z_i)\\
    \text{s.t.}\hspace{7.5mm} &\sum_{i\in \mathcal{I}}z_i\leq k,
\end{aligned}
\end{equation}
where $k\in \mathbb{Z}_+$ is the maximum number of outliers allowed. Figure~\ref{fig: outlier}b shows an optimal solution obtained from solving a problem of the form \eqref{eq:supervised_outliers} to optimality, which indeed is able to identify the outlier observation while correctly fitting the remainder of the data.

The continuous relaxation of \eqref{eq:supervised_outliers} is  highly nonlinear and non-convex due to the products of binary variables with the nonlinear loss function: in the case of the least trimmed squares where $\ell$ is the quadratic loss, the optimization problem is cubic and difficult to solve exactly \cite{giloni2002least}. The statistical and ML communities have typically relied on simpler proxies that are easier to compute. Unfortunately, such methods can yield lower-quality solutions: Figure~\ref{fig: outlier}c-\ref{fig: outlier}e shows how three commonly used alternatives fail to detect the outlier and may also result in worse accuracy in the remaining of the data.

Recent approaches to tackle \eqref{eq:supervised_outliers} have developed reformulations with simpler continuous relaxations: for example, consider
\begin{subequations}\label{eq:lts}
\begin{align}
    \min_{\bs{\theta}, \mathbf{z}, \bs{r}}\;& \ell(f_{\bs{\theta}}(\mathbf{x}^i), y^i-r_i)\\
    \text{s.t.}\hspace{1.25mm} & -Mz_i \leq r_i \leq Mz_i \label{eq:lts_bigM}\\
    & \sum_{i \in \mathcal{I}} z_i \leq k\\
    &\bs{\theta}\in \mathbb{R}^d, \mathbf{z}\in \{0,1\}^{n}, \bs{r}\in \mathbb{R}^d.
\end{align}
\end{subequations}
Intuitively, $\mathbf{r}$ can be interpreted as a vector of ``corrections" to the label $\mathbf{y}$. If $z_i=0$, then constraints \eqref{eq:lts_bigM} ensure $r_i=0$ and there is no correction to the data; if $z_i=1$, then $r_i$ is arbitrary and in optimal solutions $r_i=y_i-f_{\bs{\theta}}(\mathbf{x}^i)$, i.e., the correction is such that the label matches exactly the prediction from model $f_{\bs{\theta}}$, incurring a zero loss (thus effectively discarding the point). Formulation \eqref{eq:lts} was proposed by \citet{insolia2022simultaneous} in the context of the LTS problem, see also \cite{zioutas2005deleting,zioutas2009quadratic} for alternative formulations. The key advantage of \eqref{eq:lts} is that its continuous relaxation is of the same class as the nominal problem, and is convex if \eqref{eq:supervised} is. Unfortunately, the continuous relaxation is also extremely weak: if constraints $\mathbf{z}\in \{0,1\}^{\mathcal{I}}$ are relaxed, then setting $z_i=\epsilon$ for all $i\in \mathcal{I}$, $\bs{\theta}$ to any arbitrary value (e.g., such that $f_{\bs{\theta}}(\mathbf{x})=0$ is the constant function) and $r_i=y_i-f_{\bs{\theta}}(\mathbf{x})$ is optimal with a trivial objective value of $0$. As a consequence, formulation \eqref{eq:lts} struggles with problems with $n$ in the hundreds, even when the loss is quadratic.

Nonetheless, new innovations are being proposed in the literature. For example, \citet{gomez2023outlier} propose improved relaxations similar to those presented in \S\ref{sec:subsetselect_lr}, although the implementation is more complex. Moreover, several works consider special cases arising in inference of graphical models, where the additional structure can be leveraged to construct simpler relaxations or even design polynomial-time algorithms for special cases \cite{bhathena2025parametric,han2024compact,gomez2021outlier}. Moreover, while most of the work cited focuses on LTS problems or similar variants, recent approaches consider logistic regression \cite{stats4030040}, least median of squares problems \cite{bertsimas2014least,puerto2024fresh} or support vector machine problems with outliers \cite{cepeda2024robust}.  While the specific implementations may differ for other ML models, the methods share similar difficulties and potential.

\subsection{Adversarial Attacks and Verification} \label{sec:adv_ex}

Adversarial attacks refer to small, intentionally crafted perturbations of input data that cause an ML model $f_{\bs{\theta}}$ to misclassify the input. Formally, given an instance $\mathbf{x}$ with true (or initially predicted) label $y=f_{\bs{\theta}}(x)$, an adversarial attack finds a perturbation $\mathbf{u}$ that is small in terms of a prespecified norm and radius, ($\|\mathbf{u}\|\leq \epsilon$), but that results in a different label assignment ($y\neq f_{\bs{\theta}}(\mathbf{x}+\mathbf{u})$). Robustness to adversarial attacks means that no such perturbation exists. Verifying a model’s robustness can be posed as a constrained optimization: 
\begin{equation}\label{eq:evasion}
    \begin{aligned}
        \min_{\mathbf{u}\in \mathbb{R}^d} \; & \|\mathbf{u}\| \\
        \text{s.t.}\hspace{1.25mm}& f_{\bs{\theta}}(\mathbf{x}+\mathbf{u}) = k,
    \end{aligned}
\end{equation}
where $k\neq y$ is given a priori; if the objective value of \eqref{eq:evasion} is greater than $\epsilon$, then the ML model can be certified as robust \emph{for the specific input} $\mathbf{x}$ and label $k$. In practice, determining that a given ML model is truly robust requires solving \eqref{eq:evasion} a plethora of times for different possible inputs, which can be done in parallel if enough computational resources are available. Note that in this case the ML model is trained a priori, thus the parameters $\bs{\theta}$ are given and not decision variables. 

If function $f_{\bs{\theta}}$ is simple, e.g., affine in the input, then problem \eqref{eq:evasion} is easy and may be even solvable by inspection. However, if the ML model is complex, then the nonlinear equality constraint in problem \eqref{eq:evasion} may pose substantial difficulties. In such cases,
MIO provides a framework to exactly search for adversarial examples and certify robustness, especially for models with discrete or piecewise-linear structure. For instance, for a ReLU-based neural network, the non-convexities required to model $f_{\bs{\theta}}$ correspond to the activation functions \cite{tjeng2017evaluating}. Indeed, modeling the output of a trained artificial neural network involves handling non-convex equality constraints of the form 
\begin{equation} \label{eq:relu}\max\{0, \bs{\theta}^\top \mathbf{r}\}= s,
\end{equation}
where $\bs{\theta}$ are given weights, $\mathbf{r}$ are decision variables representing the \emph{pre-activation} variables at a given neuron (i.e., outputs from the previous layer), and $s$ is a decision variable representing the post-activation variable or output of the neuron.  If lower bounds and upper bounds such that $l\leq \bs{\theta}^\top \mathbf{r}\leq u$ are known, non-convex constraints such as \eqref{eq:relu} can be linearized using the system
\begin{equation*} 0\leq s,\; \bs{\theta}^\top \mathbf{r}\leq s,\; s\leq uz,\; s\leq \bs{\theta}^\top \mathbf{r}-l(1-z),
\end{equation*}
where $z\in \{0,1\}$ is an auxiliary binary variable representing the logical condition~$z=\mathbbm{1}[\bs{\theta}^\top \mathbf{r}\geq0]$.

\citet{tjeng2017evaluating} pioneered this MIO verification approach for piecewise-linear networks. The continuous relaxations can be weak, and thus direct MIO approaches do not scale to practical sizes. However, in a series of works by \citet{anderson2020strong,anderson2020tightened} and \citet{tjandraatmadja2020convex}, it is shown how to further improve the relaxation using MIO methods. The resulting formulation can either be used directly to verify small networks, or used as a basis for a propagation heuristic that scales to large datasets while improving the performance versus competing alternatives. Note that while solving these MIOs can be computationally intensive, they offer strong guarantees: any solution found is a genuine adversarial example, and if none is found, the model is provably robust (under the defined attack model). This is in contrast to heuristic attacks or incomplete verification methods which might miss worst-case attacks. See also \cite{aftabi2025feed,patil2022mixed,thorbjarnarson2021training,thorbjarnarson2023optimal,khalil2018combinatorial,venzke2020verification,hojny2024verifying} for related work on MIO methods for handling robustness with artificial neural networks. 
 MIO based methods have also been proposed to verify tree ensembles \cite{kim2022convexification,kantchelian2016evasion,zhang2020decision}.

We note that problem~\eqref{eq:evasion} is mathematically similar to the problem for finding counterfactual explanations~\eqref{eq:cf_explanation} as described in section~\ref{sec:counterfact_ex}. Indeed, both formulations aim to find a minimal perturbation of an instance $\mathbf{x}$ that causes the instance to be classified according to a given fixed class, and methods in one area may be applied to the other. However, the motivations for each are very different, leading to potentially different functions to minimize and different constraints to consider. For instance, sparsity may be required to make CEs explainable and actionable. In contrast, adversarial attacks may instead require \textit{imperceptibility}, i.e., an attack should be difficult to perceive in order to be undetectable by security measures. We refer to~\citet{freiesleben2022intriguing} for a further exploration on the relationship between these two fields.
 
\subsection{Robustness to Adversarial Attacks and to Distribution Shifts} \label{sec:robust_atk_shift}

Rather than only verifying robustness \emph{post hoc}, we can use MIO at \emph{training time} to build models that account for distribution shifts or adversarial attacks. A natural approach to enhance robustness is to integrate adversarial perturbation scenarios during training, e.g., by adding attacks detected with methods described in section~\ref{sec:adv_ex} as new instances in the training set. Such methods treat both the algorithms describing the learning model and the attack pattern as black boxes and can be used in a variety of contexts, but may be inefficient in practice as they require retraining the model and detecting attacks several times.

An alternative approach is to use \emph{robust optimization} (which prepares the model to handle malicious or extreme perturbations to features) or \emph{distributionally robust optimization} (which ensures good performance for all test distributions in some sense ``close'' to the training distribution) paradigms. Robust optimization methods require describing an uncertainty set $\mathcal{U}\subseteq \mathbb{R}^{n\times d}$, corresponding to all possible attacks, and tackles the robust counterpart of the nominal training problem \eqref{eq:supervised} defined as 
\begin{equation} \label{eq:robust_obj}
        \min_{\bs{\theta}\in \Theta} \;\; \max_{\mathbf{U} \in \mathcal{U}} \hspace{2mm} \sum_{i \in \mathcal{I}} \ell(f_{\bs{\theta}}(\mathbf{x}^i + \mathbf{u}^i), y^i),
\end{equation}
where $\mathbf{u}^i$ denotes the $i$-th row of $\mathbf{U}$ and corresponds to the attack/shift on datapoint $i\in \mathcal{I}$. Distributionally robust optimization formulations yield optimization problems similar in structure to \eqref{eq:robust_obj}. Robust and distributionally robust approaches have been widely used in machine learning, see for example \cite{kuhn2019wasserstein}. A vast majority of approaches operate under convexity or continuity assumptions, in which case \eqref{eq:robust_obj} can be conveniently handled by reformulating it as an equivalent deterministic optimization problem using duality arguments. 
However, duality methods cannot be directly applied in the presence of non-convexities. For example the loss function may not be non-convex, corresponding to a discrete $0-1$ loss or a piecewise linear classifier such as a decision tree or artificial neural network \cite{kurtz2021efficient,vos2022robust}. Alternatively, the uncertainty set $U$ may be discrete, e.g., when modeling perturbations to integer or categorical features \cite{sun2025mixed,belbasi2023s}, or a combination of both sources of non-convexities \cite{justin2021optimal, justin2023learning}. MIO technology can be effective in settings when duality fails.

A common approach to tackle robust optimization problems under non-convexities is to use delayed constraint generation methods. Problem \eqref{eq:robust_obj} can be reformulated with the introduction of an epigraphical variable as 
\begin{subequations} \label{eq:rowGen}
\begin{align}
        \min_{\bs{\theta},r}\;& r\\
        \text{s.t.}\;&\sum_{i \in \mathcal{I}} \ell(f_{\bs{\theta}}(\mathbf{x}^i + \mathbf{u}^i), y^i)\leq r\qquad \forall \mathbf{U}\in \mathcal{U},
        \end{align}
\end{subequations}
transforming the $\min$-$\max$ optimization problem into a deterministic problem with a large (potentially infinite) number of constraints. Delayed constraint generation methods alternate between: \textit{(a)} ``processing'' relaxations of problem \eqref{eq:rowGen} where set $\mathcal{U}$ is replaced with a finite subset containing candidate attacks/shifts, obtaining promising solutions $(\bar{\bs{\theta}}, \bar r)$; \textit{(b)} solving the \emph{separation} problem 
\begin{equation}\label{eq:separation}\min_{\mathbf{U}\in \mathcal{U}}\;\bar r-\sum_{i \in \mathcal{I}} \ell(f_{\bar{\bs{\theta}}}(\mathbf{x}^i + \mathbf{u}^i), y^i)\end{equation}
to detect the most damaging attack/shift to the solution $(\bar{\bs{\theta}}, \bar r)$, and adding it to the set of candidate attacks. Note that a naive implementation of constraint generation methods, where ``processing" corresponds to ``solving to optimality," reduces to the iterative procedure described in the first paragraph of the subsection. However, better implementations based on MIO algorithms, e.g., by adding inequalities as a subroutine while constructing a single branch-and-bound tree, can achieve orders-of-magnitude improvements in computational times. 
Even better methods are possible when the non-convex separation problem \eqref{eq:separation} can be reformulated (or accurately approximated) as a convex optimization problem. For example, in the setting considered by \citet{sun2025mixed}, problem \eqref{eq:separation} can be solved via dynamic programming, a method that solves combinatorial problems by breaking them down into simpler subproblems and solves each subproblem only once while avoiding redundant computations. They then exploit the fact that a dynamic programming scheme can be encoded as a shortest path problem on a DAG, and use linear optimization duality to recover an exact deterministic formulation of polynomial size. This deterministic reformulation is shown to perform an order-of-magnitude better than constraint generation approaches.

\section{Observational Settings}\label{sec:observational}
In many real-world settings such as healthcare, education, and public policy, a central goal is to understand how different interventions (called \textit{treatments}) affect \emph{outcomes} and to use this information to guide future decisions (e.g., to approve a vaccine for use on the population at large, to make individualized treatment decisions for a medication, or to decide how to allocate scarce resources to people who need them). While randomized controlled trials (RCTs) are the gold standard for estimating such \emph{treatment effects} and for making such intervention decisions, they are often impractical or unethical to conduct at scale, or may take too long to conduct. As a result, practitioners and policymakers frequently rely on \textit{observational data} -- data collected in deployment rather than during RCTs -- wherein historical treatments are assigned based on a (potentially unknown) policy (e.g., social workers or doctors triaging clients/patients based on their personal characteristics or symptoms). However, drawing causal conclusions from such data is fraught with challenges, including selection bias and distribution shifts. These issues must be carefully addressed to avoid misleading inferences and suboptimal or unfair decisions.

\subsection{Notation and Assumptions}
\label{sec:causal_assumptions}
We follow standard notation from the causal inference literature, see e.g.,~\cite{hernan2010causal}. Each individual is characterized by their \emph{covariates}, represented by random variable $X \in \mathbb R^d$ and their \emph{potential outcomes} $Y(a)$, $a \in \mathcal A \subseteq \mathbb R$ under treatment $a$, where set $\mathcal A$ collects all treatment options. A major challenge in observational settings is that we do not know the joint distribution of $(X, \{Y(a)\}_{a\in \mathcal A})$, neither do we have access to observations from this joint distribution. Instead, we have access to samples from the joint distribution of $(X,A, Y)$, where $A$ represents the treatment historically assigned and $Y$ is the \emph{observed outcome} under the treatment received. Specifically, this historical \emph{observational} dataset is of the form ~$\{\mathbf{x}^i, a^i, y^i\}_{i \in \mathcal{I}}$ for $\mathbf{x}^i \in \mathbb{R}^d$ the covariates of individual $i$, $a^i \in \mathcal A$ is the treatment given to individual $i$, and $y^i \in \mathbb{R}$ is the outcome we observed for them in the data.


Most of the literature on causal inference makes the following assumptions, which ensure that treatment effects are \emph{identifiable}, i.e., possible to estimate from the observed data:

\begin{assumption}[Conditional Exchangeability] \label{as:conditional_exchangeability}
    The treatment assignment $A$ is independent of the potential outcomes $Y(a)$ for all $a \in \mathcal{A}$ conditional on the covariates $X$. That is, for all~$a \in \mathcal{A}$,~$Y(a) \indep A \mid X$.
\end{assumption}
\begin{assumption}[Positivity] \label{as:positivity}
    The probability of receiving any treatment given X must be positive. That is, $\mathbb{P}(A = a \mid X = \mathbf{x}) > 0$ for all $a \in \mathcal{A}$ and $\mathbf{x} \in \mathbb{R}^d$.
\end{assumption}

\begin{assumption}[Consistency]\label{as:consistency}
    The potential outcome predicted is equal to the observed outcome in the data. That is, for $Y$ the observed outcome, $Y = Y(A)$.
\end{assumption}

Assumption~\ref{as:conditional_exchangeability} states that once we account for (or ``condition on'') the right set of variables, the different treatment groups become comparable -- as if they were randomly assigned.  If we can measure and control for all the relevant factors that influence both who gets the treatment and the outcome, then the remaining difference in outcomes between treatment groups can be interpreted as a (causal) treatment effect. Assumption~\ref{as:conditional_exchangeability} will hold if we have access to all information that was used historically to decide on treatments. While it cannot be verified from the data, it frequently holds in real settings and can often be checked by talking with domain experts or with people who did historically make the treatment assignment decisions. 

Assumption~\ref{as:positivity} means that every type of individual in the data -- based on their observed characteristics -- has a positive chance of receiving each treatment. This assumption ensures that for all types of individuals we have at least some information on how they will fare if they get any one of the treatments. This assumption is relatively easy to check from the data and can also often be validated from conversations with domain experts and practitioners. 

Assumption~\ref{as:consistency} connects the treatment we observe someone receive with the potential outcome we theoretically define for that treatment. Intuitively, it means: if a person actually receives a particular treatment, then their observed outcome is exactly the outcome they would have had under that treatment in our causal model. While this assumption seem obvious, it may fail in subtle ways. For example, if ``taking the medication'' is not clearly defined—some people skip doses, or take different versions of it—then the ``treatment'' becomes ambiguous. Similarly, if the way a treatment is delivered affects the outcome (e.g., surgery by different doctors), then consistency can break down unless those differences are part of the treatment definition. As with the other two assumptions, this assumption can be checked by understanding how treatment is recorded and delivered.

Assumptions~\ref{as:conditional_exchangeability},~\ref{as:positivity}, and~\ref{as:consistency} together allow us to treat observational data as if it came from an RCT by ensuring that, conditional on the observed covariates
$X$, treatment assignment is independent of potential outcomes and the observed outcome corresponds to the potential outcome under the treatment actually received. Therefore, we will make these assumptions throughout this section and reemphasize the need to check that these assumptions hold when using the methods in this section in practice.

With these standard causal inference assumptions in mind, we explore two streams that use causal inference and MIO together in observational settings. Section~\ref{sec:matching} details matching methods to estimate treatment effects, and section~\ref{sec:prescriptive_ml} explores learning prescriptive policies in observational settings.

\subsection{Matching to Estimate Treatment Effects} \label{sec:matching}
\begin{wrapfigure}{r}{0.35\textwidth}
  \begin{center}
\includegraphics[width=0.35\textwidth, trim={3cm 5cm 3cm 9cm},clip]{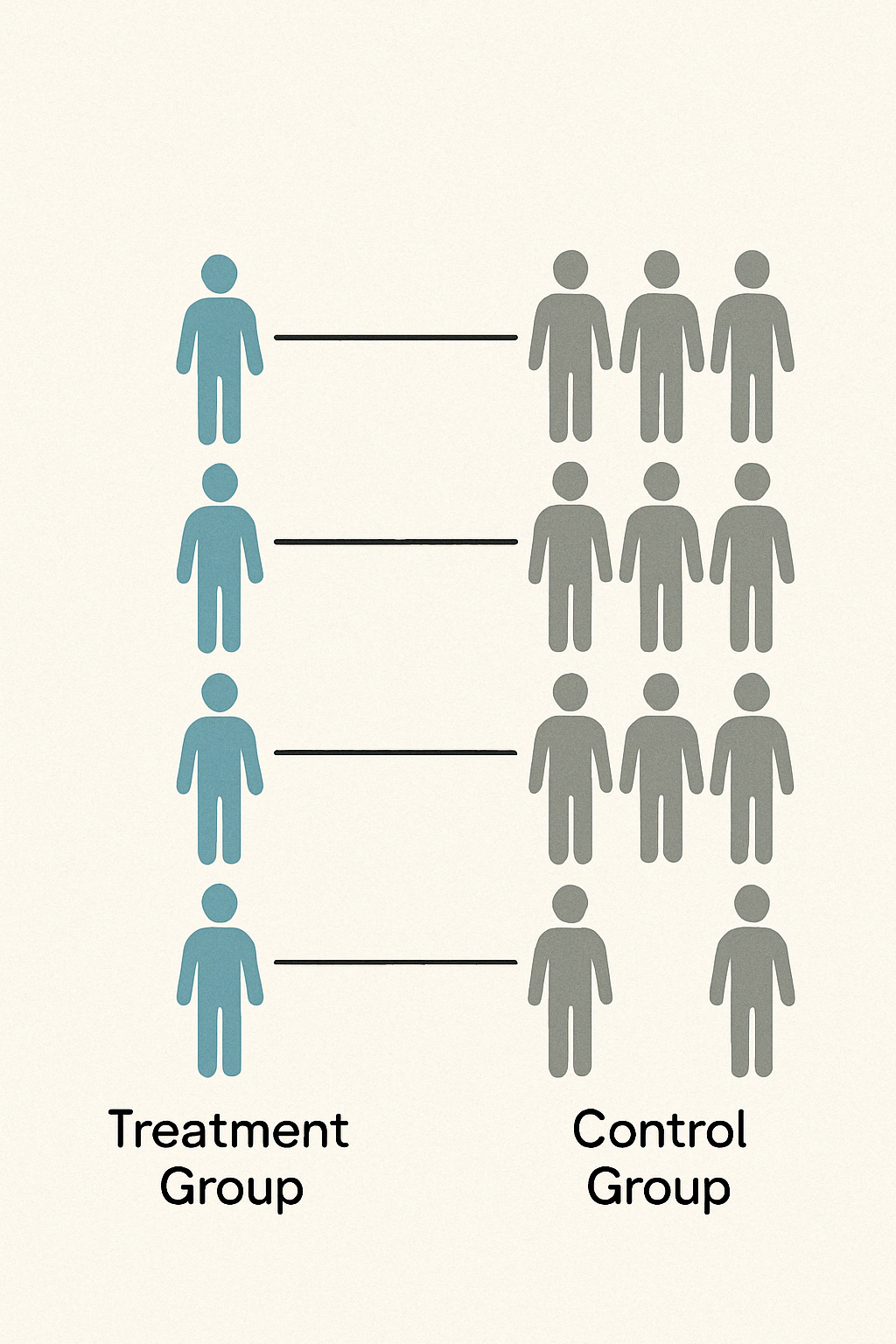}
  \end{center}
  \caption{Matching in observational studies: every individual in the treatment group is matched to (i.e., represented by) several similar individuals in the control group. The sizes of the subgroups matched to each individual are roughly the same across individuals, and no one in the control group is matched with two individuals.}
  \label{fig:matching}
\end{wrapfigure}

One method to estimate treatment effects is via matching. For example, given a treatment group and a control group, where the control group is presumed to be larger, the idea is to match each individual from the treatment group with one or several individuals from the control group with similar characteristics, as illustrated in Figure~\ref{fig:matching}. Then, counterfactual estimates for individuals in the treatment group, corresponding to outcomes if no intervention is performed, can be obtained by averaging the observed outcomes for matched individuals in the treatment group. 

Consider the following optimization problem to match between a treatment and a control group ($|\mathcal{A}|=2$), where $\mathcal{T}\subseteq \mathcal{I}$ are the individuals that received a treatment and $\mathcal{C}=\mathcal{I}\setminus \mathcal{T}$ denotes the control group, where we assume that $m|\mathcal{T}| \leq |\mathcal{C}|$ for some $m\in \mathbb{Z}_+$. Letting $d_{tc}$ be a metric of the dissimilarity between $t\in \mathcal{T}$ and $c\in \mathcal{C}$, e.g., $d_{tc}=\|\mathbf{x}^t-\mathbf{x}^c\|$, consider the optimization problem 
\begin{subequations} \label{eq:matching}
    \begin{align}
        \min_{\mathbf{z}\in \{0,1\}^{\mathcal{T}\times \mathcal{C}}}\; & \sum_{t \in \mathcal{T}}\sum_{c \in \mathcal{C}} d_{tc}z_{tc}\label{eq:matching_obj}\\
        \text{s.t.}\; & \sum_{c \in \mathcal{C}} z_{tc} = m & \forall t \in \mathcal{T} \label{eq:matching_m}\\
        & \sum_{t \in \mathcal{T}} z_{tc} \leq 1 & \forall c \in \mathcal{C}\label{eq:matching_1}
    \end{align}
\end{subequations}
where decision variables $z_{tc}=1$ if and only if individuals $t$ and $c$ are matched together. 
Constraint \eqref{eq:matching_m} ensures that each individual in the treatment group is matched with $m$ 
individuals of the control group, constraint \eqref{eq:matching_1} ensures that no individual in the treatment group is paired with two or more treated individuals, and the objective minimizes dissimilarities. 

Optimization problem \eqref{eq:matching} is easy to solve and does not require sophisticated MIO technology. In fact, the feasible region of the continuous relaxation has a special property: every extreme point of the polytope induced by constraints \eqref{eq:matching_m}-\eqref{eq:matching_1} and constraints $\mathbf{0}\leq \mathbf{z}\leq \mathbf{1}$ is integral. Thus, since the objective function is linear, there exists an optimal solution of the linear optimization relaxation that is an extreme point and thus satisfies the integrality restrictions. Recognizing polytopes with integral extreme points is in general hard, but some special cases can be recognized immediately. In the case of \eqref{eq:matching}, imagine that both sides of constraints \eqref{eq:matching_m} are multiplied by $-1$. Then observe that all variables appear in at most two constraints (not including bound constraints) with opposite signs: any constraint matrix satisfying this condition is \emph{totally unimodular}, and the resulting polytope is guaranteed to have integral extreme points. 

While formulation \eqref{eq:matching} is attractive due to its computational simplicity, the resulting matching could be low-quality and lead to poor counterfactual estimations. Indeed, while the overall dissimilarity is minimized, it is possible that the subset of the control population used differs substantially from the treated individuals in the values of some critical features. Mathematically, given a covariate $j$, $$\frac{1}{|\mathcal{T}|}\sum_{t\in \mathcal{T}}x_{j}^t\not\approx \frac{1}{m|\mathcal{T}|}\sum_{t\in \mathcal{T}}\sum_{c\in \mathcal{C}}x_j^cz_{tc},$$
where the left hand size is the average value of covariate $j$ in the treatment group, and the right hand size represents that average value for the selected individuals in the control group. Naturally, if feature $j$ is critical to the outcomes under no treatment, then any counterfactual estimation based on formulation \eqref{eq:matching} would be poor. 

Traditionally, this covariate imbalance has been tackled through heuristic iterative procedures. However, \citet{zubizarreta2012using} proposed to instead modify the objective of \eqref{eq:matching} to 
 $$\min_{\mathbf{z}\in \{0,1\}^{\mathcal{T}\times \mathcal{C}}}\;  \sum_{t \in \mathcal{T}}\sum_{c \in \mathcal{C}} d_{tc}z_{tc}+\sum_{j\in [d]}\lambda_j \left| \frac{1}{m|\mathcal{T}|}\sum_{t\in \mathcal{T}}\sum_{c\in \mathcal{C}}x_j^cz_{tc}-\frac{1}{|\mathcal{T}|}\sum_{t\in \mathcal{T}}x_{j}^t\right|,$$
 where $\lambda_j$ are weights representing the importance of each covariate, to ensure balanced covariates. This modification destroys the integrality property of the MIO formulation: while the feasible region has not changed, the objective is now nonlinear and optimal solutions of the continuous relaxation may no longer be extreme points. The MILO reformulation of the ensuing optimization problem given by 
    \begin{align*}
        \min_{\mathbf{z}\in \{0,1\}^{\mathcal{T}\times \mathcal{C}},\mathbf{r}\in \mathbb{R}^d}\; & \sum_{t \in \mathcal{T}}\sum_{c \in \mathcal{C}} d_{tc}z_{tc}+\sum_{j\in [d]}\lambda_j r_j\\
        \text{s.t.}\; & \sum_{c \in \mathcal{C}} z_{tc} = m & \forall t \in \mathcal{T} \\
        & \sum_{t \in \mathcal{T}} z_{tc} \leq 1 & \forall c \in \mathcal{C}\\
        &\frac{1}{m|\mathcal{T}|}\sum_{t\in \mathcal{T}}\sum_{c\in \mathcal{C}}x_j^cz_{tc}-\frac{1}{|\mathcal{T}|}\sum_{t\in \mathcal{T}}x_{j}^t\leq r_j&\forall j\in [d]\\
        &\frac{1}{|\mathcal{T}|}\sum_{t\in \mathcal{T}}x_{j}^t-\frac{1}{m|\mathcal{T}|}\sum_{t\in \mathcal{T}}\sum_{c\in \mathcal{C}}x_j^cz_{tc}\leq r_j&\forall j\in [d]
    \end{align*}
 destroys total unimodularity as every variable appears in multiple constraints. Nonetheless, despite the increased complexity, \citet{zubizarreta2012using} shows that the optimization problems can still be solved in practice using MIO, resulting in improved solutions. Further improvements or alternative formulations have been proposed in the literature, including problems with multiple treatment groups \cite{bennett2020building}, more sophisticated objective functions \cite{sun2016mutual}, including robustness \cite{morucci2022robust} or unifying formulations \cite{kallus2017framework,cousineau2023estimating}. 


\subsection{Prescriptive Models} \label{sec:prescriptive_ml}

In this section, we explore how MIOs are used to learn prescriptive ML models that assign treatments to indviduals based on their covariates. Specifically, the goal is to learn a policy $f_{\bs{\theta}}(\mathbf{x}) : \mathbb{R}^d \mapsto \mathcal{A}$ using data $\{\mathbf{x}^i, a^i, y^i\}_{i \in \mathcal{I}}$ that maximizes the overall well-being of the population, where $\bs{\theta}$ in this case are the parameters describing the policy. If positive outcomes are associated with large values of the outcome variable $Y$, the goal is to solve
\begin{equation} \label{eq:prescriptive_ml}
    \max_{\bs{\theta}\in \Theta} \ \mathbb{E}_{\mathbb{P}}[Y(f_{\bs{\theta}}(X))],
\end{equation}
where $\mathbb{P}$ is the joint distribution describing the distribution of the samples and their outcomes under different treatments. 
Naturally, the distribution $\mathbb{P}$ needs to be estimated from data. 

Prescriptive methods to tackle \eqref{eq:prescriptive_ml} typically consist of two steps. First, a score function $s(\mathbf{x},a):\mathbb{R}^{d}\times \mathcal{A}\to \mathbb{R}$ which serves as a proxy for the outcome of samples with covariates $\mathbf{x}$ and treatment $a$ is learned. Then \eqref{eq:prescriptive_ml} can be approximated as
\begin{equation}\label{eq:policyLearning}\max_{\bs{\theta}\in \Theta}\sum_{i\in \mathcal{I}}s\left(\mathbf{x}^i, f_{\bs{\theta}}(\mathbf{x}^i)\right).\end{equation}
In high stakes domains, interpretable policies are preferred. 
Therefore, the MIO techniques discussed in section~\ref{sec:interpretability} are natural options to tackle \eqref{eq:policyLearning}. Note that modeling the score function may introduce additional non-convexities, resulting in more challenging MIOs with additional variables. Indeed, representing \eqref{eq:policyLearning} requires introduction of binary variables $z_a^i$ that indicate whether treatment $a$ is assigned to the policy specified by $\bs{\theta}$, so that \eqref{eq:policyLearning} can be reformulated as
$$\max_{(\bs{\theta},\mathbf{z})\in \mathcal{F}}\sum_{i\in \mathcal{I}}\sum_{a\in \mathcal{A}}s\left(\mathbf{x}^i,a\right)z_{a}^i,$$
where $\mathcal{F}$ is a feasible region including constraints $\bs{\theta}\in \Theta$, $\mathbf{z}\in \{0,1\}^{\mathcal{I}\times \mathcal{A}}$ as well as linking constraints used to encode the definition of $\mathbf{z}$,
see \cite{jo2021learning,zhang2020decision,amram2022optimal,subramanian2022constrained,aouad2023market} for examples in the context of decision trees. 
Techniques that account for robustness (section~\ref{sec:robustness}) or fairness (section~\ref{sec:privacy-fair}) can also be included. 

A possible approach for the score function $s(\mathbf{x},a)$ is to, for each possible treatment $a\in \mathcal{A}$,  learn a model $\nu_{a}(\mathbf{x}) : \mathbb{R}^d \mapsto \mathbb{R}$ that computes a predicted outcome for $\mathbf{x}$ under treatment $a \in \mathcal{A}$, where $\nu_a$ is obtained from any classical ML model using only the subpopulation of the data $\{\mathbf{x}^i, y^i\}_{i\in \mathcal{I}:a^i = a}$ that received treatment $a$. 
This direct approaches achieves excellent results in randomized control trials, but may lead to poor conclusions if the populations receiving each treatment differ significantly. 

An alternative approach is to use inverse propensity score weighting. The idea is to first learn a propensity model $\mu(\mathbf{x}, a) := \mathbb{P}(A =a \mid X = \mathbf{x})$ that predicts the probability of a treatment assignment given the covariates of an individual. Under the assumptions in section~\ref{sec:causal_assumptions}, accurate estimation using classical ML methods is possible using the entire training set $\mathcal{I}$. Then a good score function is obtained as 
\begin{equation}\label{eq:ipw}s(\mathbf{x}, a)=\frac{1}{|\mathcal{I}|\mu(\mathbf{x}^i,a^i)}\sum_{i\in \mathcal{I}:(\mathbf{x}^i,a^i)=(\mathbf{x}, a)}y^i.\end{equation} 
The estimates and policies obtained using \eqref{eq:ipw} can be shown to be asymptotically optimal provided that the propensity estimations $\mu(\mathbf{x}, a) := \mathbb{P}(A =a \mid X = \mathbf{x})$ are good \cite{jo2021learning}. Unfortunately, estimations can also have large variances, requiring large amounts of data to provide trustworthy solutions. A better approach is to use the doubly robust scores
\begin{equation}\label{eq:doublyRobust}s(\mathbf{x},a)=\nu_a(\mathbf{x})+\frac{1}{|\mathcal{I}|\mu(\mathbf{x}^i,a^i)}\sum_{i\in \mathcal{I}:(\mathbf{x}^i,a^i)=(\mathbf{x}, a)}\left(y^i-\nu_a(\mathbf{x})\right),\end{equation}
which intuitively use a direct method to estimate the score and inverse propensity weighting to estimate the errors of the direct method. Under reasonable conditions on the data and models $\nu$ and $\mu$, using \eqref{eq:doublyRobust} results in ideal asymptotic performance without incurring large variances, see \cite{dudik2014doubly} for an in-depth discussion.

\section{Fairness}\label{sec:privacy-fair}

When learning machine learning models, it is often desirable to make ML-based predictions and decisions that are fair across groups based on a protected characteristic (e.g., race, age, gender, economic status). For instance, consider again the \href{https://archive.ics.uci.edu/dataset/183/communities+and+crime}{``Communities and Crime"} dataset as shown in Figure~\ref{fig:outputsLinReg}, where the goal is to predict the number of violent crimes per capita. A major concern in developing a machine learning model using this dataset is that it may reinforce racial biases in predicting which communities are violent. In particular, models produced by elastic net, shown in Figures~\ref{fig:outputsLinReg}b and \ref{fig:outputsLinReg}c, include features related to the model with a positive coefficient, clearly indicating that (if all other features are identical) cities with large proportions of racial minorities will be associated with larger crime rates.

An initial attempt to learn a fair model may be to simply remove the protected features in the dataset and train the ML model without the protected characteristics. For example, the model obtained by MIO in Figure~\ref{fig:outputsLinReg}d does not use any racial features (even if this was not explicitly enforced when training), suggesting that it may be fair. This \textit{fairness through unawareness} or \textit{blindness} approach, however, does not guarantee that the learned model will be fair: features that are correlated with the protected characteristic(s) may still be used in the learned model to perpetuate biases or to introduce new biases and lead to unfairness. For example, location information can be used by an ML model as a proxy for demographics of a neighborhood, leading to bias in a learned model~\citep{Ingold_Soper_2016}. The interpretable model in Figure~\ref{fig:outputsLinReg}d may indeed be unfairly biased: a closer inspection reveals that, according to this model, only 23\% of cities with a large Hispanic community (which we define as at least 18\% of the population) have below average crime rate. Explicitly accounting for fairness can help alleviate these issues.
Fair ML models can be learned by adjusting the features of the training dataset to be uncorrelated with sensitive attributes (pre-processing), inserting constraints into the learning process of a fair classifier (in-training), or adjusting a trained model to be fair (post-processing)~\cite{barocas2023fairness}. We explore fairness in predictive models in section~\ref{sec:fair_predict}, and extend the fairness measures presented in section~\ref{sec:fair_predict} to other settings in section~\ref{sec:fair_extensions}.

\subsection{Fairness in Predictive Models} \label{sec:fair_predict}
MIOs are particularly useful for in-training methods for fair ML, as many fairness notions can be modeled as a set of additional constraints in MIO-based ML models. In general, binary variables $\zeta^i\in \{0,1\}$ for $i\in \mathcal{I}$ can be introduced such that $\zeta^i=1$ if and only if the outcome for $i$ is favorable. Let $\mathcal{F}\subseteq \mathcal{Y}$ denote the set of favorable outcomes. Then, a version of \eqref{eq:supervised} that can be modified with fairness constraints is
\begin{subequations}\label{eq:fair_setup}
    \begin{align}
    \min_{\bs{\theta}\in \Theta,\mathbf{z}\in \{0,1\}^n}\;& \sum_{i \in \mathcal{I}} \ell(f_{\bs{\theta}}(\mathbf{x}^i), y^i)\\
    \text{s.t.}\hspace{6.75mm} & \zeta^i=\mathbbm{1}[ f_{\bs{\theta}}(\mathbf{x}^i)\in \mathcal{F}]\hspace{10mm} \forall i\in \mathcal{I},\label{eq:fair_setup_Z}
    \end{align}
\end{subequations}
where indicator constraints \eqref{eq:fair_setup_Z} can often be linearized using standard MIO methods. Fairness constraints can be used to promote \textit{individual fairness}~\citep{dwork2012fairness}, which treats similar individuals similarly, or \textit{group fairness}, which treats different groups similarly. We focus on group fairness notions in this tutorial.

Group fairness constraints can be added to~\eqref{eq:fair_setup} by partitioning the training set, denoting $\mathcal{I}_0\subseteq \mathcal{I}$ as the observations belonging to a protected or underrepresented group and $\mathcal{I}_1 = \mathcal{I} \backslash \mathcal{I}_0$. In the following, we present some fairness notions in ML and their associated MIO constraints in~\eqref{eq:fair_setup}.

    \paragraph{Statistical Parity} Statistical parity guarantees that the demographics that receive positive classifications are identical to the demographics of the population as a whole~\cite{dwork2012fairness}. Statistical parity is a notion of \textit{group fairness} as it equalizes the predictions between groups. A major benefit of statistical parity is that it is the simplest form of group fairness. But because statistical parity does not consider any information on individuals in the data, statistical parity is particularly susceptible to not satisfying individual fairness. We can represent statistical parity with a tolerance of $\delta$ as
    \begin{equation}\label{eq:statistical_parity}
        \left|\frac{1}{|\mathcal{I}_0|}\sum_{i\in \mathcal{I}_0}\zeta^i-\frac{1}{|\mathcal{I}_1|}\sum_{i\in \mathcal{I}_1}\zeta^i\right|\leq \delta.
    \end{equation}

\begin{wrapfigure}{r!}{0.56\textwidth}
  \begin{center}

  \subfloat[Solution to \eqref{eq:l0l2_soc}, $R^2=0.63$]{\includegraphics[width=0.28\textwidth, trim={0cm 12.1cm 16cm 0cm},clip]{./images/best_subset.png}}
    \hfill\subfloat[Solution to \eqref{eq:l0l2_fair}, $R^2=0.60$]{\includegraphics[width=0.28\textwidth, trim={0cm 12.1cm 16cm 0cm},clip]{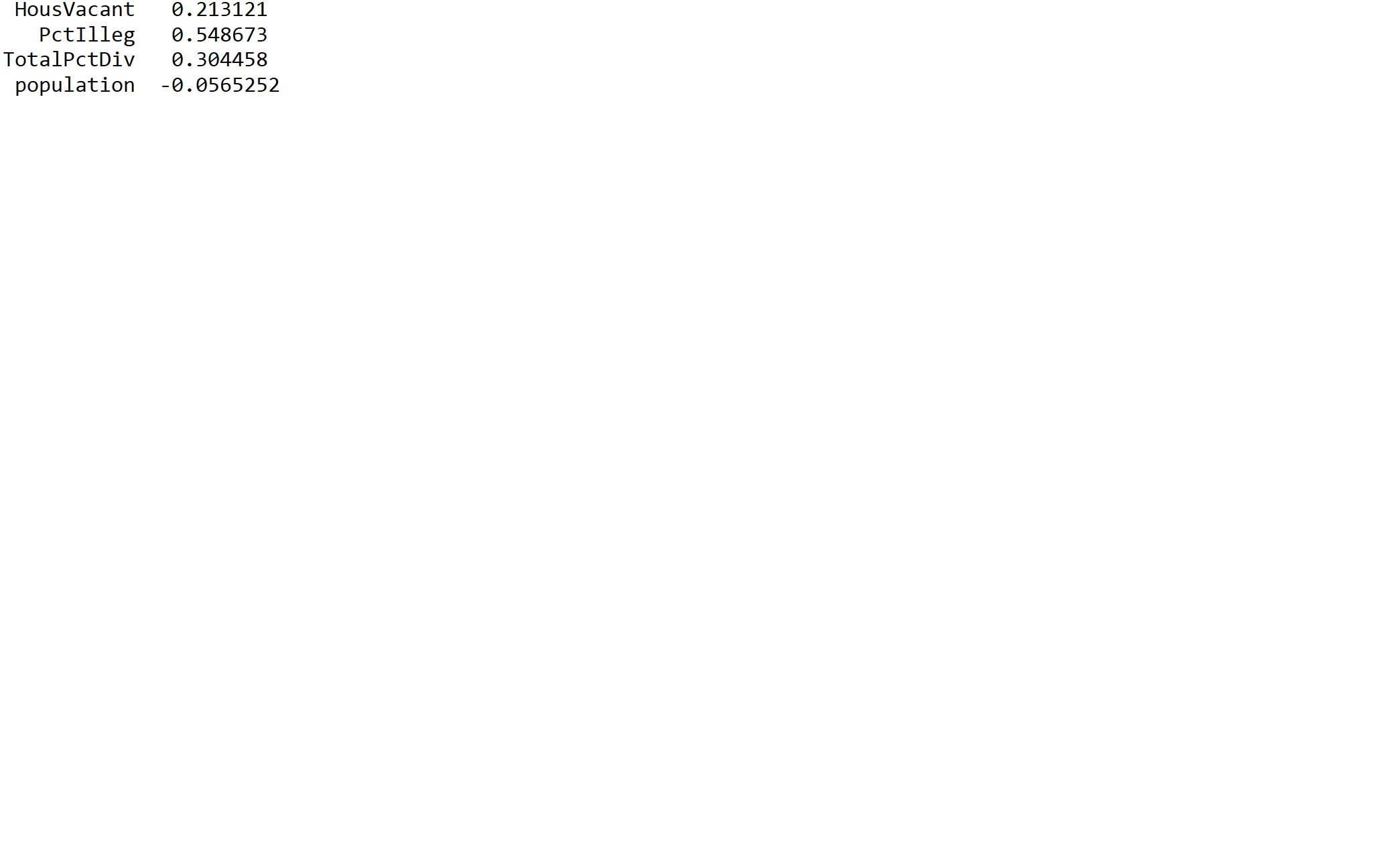}}
  \end{center}
  \caption{By imposing fairness, cities with large Hispanic populations and low estimated crime rates increase from 23\% to 37\%. 
  TotalPctDiv: percentage of population who are divorced, population: population for community.}
  \label{fig:fair_reg}
\end{wrapfigure}

For example, in (interpretable) least squares regression, define $\mathcal{I}_0$ as the set of cities with more than 18\% Hispanics, and define $\mathcal{F}=(-\infty,y_0)$ for all $i\in \mathcal{I}$, where $y_0$ is the average crime rate. Then we can build on model \eqref{eq:l0l2_soc} and impose fairness constraints, resulting in model
\begin{subequations}\label{eq:l0l2_fair}
    \begin{align}
        \min_{\bs{\theta},\mathbf{z},\mathbf{r},\bs{\zeta}} \ & \sum_{i\in \mathcal{I}} (y^i - \boldsymbol{\theta}^\top\mathbf{x}^i)^2  + \lambda_0 \sum_{j=1}^d z_j + \lambda_2 \sum_{j=1}^d r_j \\
        \text{s.t.} \hspace{1.5mm} & \theta_j^2 \leq  z_jr_j & \forall j \in [d] \\
        & -Mz_j \leq \theta_j \leq Mz_j & \forall j \in [d]\\
        &\left|\frac{1}{|\mathcal{I}_0|}\sum_{i\in \mathcal{I}_0}\zeta^i-\frac{1}{|\mathcal{I}_1|}\sum_{i\in \mathcal{I}_1}\zeta^i\right|\leq \delta\\       &-M(1-\zeta^i)\leq\boldsymbol{\theta}^\top\mathbf{x}^i-y_0\leq M\zeta^i&\forall i\in [n]\label{eq:l0l2_fair_bigM}\\
        &\boldsymbol{\theta} \in \mathbb{R}^d, \mathbf{z} \in \{0,1\}^d, \mathbf{r} \in \mathbb{R}^d,\bs{\zeta}\in \{0,1\}^n,
    \end{align}
\end{subequations}
where $y_0$ is a given threshold (e.g., the average crime rate) and constraints \eqref{eq:l0l2_fair_bigM} ensure that $\zeta^i=1$ if $\boldsymbol{\theta}^\top\mathbf{x}^i>y_0$ and $\zeta^i=0$ if $\boldsymbol{\theta}^\top\mathbf{x}^i<y_0$, encoding \eqref{eq:fair_setup_Z}. Figure~\ref{fig:fair_reg} shows the model obtained from solving \eqref{eq:l0l2_fair} with $\delta=0.2$: the estimated crime rate for cities with large Hispanic populations is substantially decreased with minimal accuracy loss.


\paragraph{Conditional Statistical Parity} A limitation of statistical parity is that it does not take into account for variables that are indicative of the outcome within each group. Conditional statistical parity equalizes predictions between groups, conditioned on legitimate variables that affect the outcome~\cite{corbett2017algorithmic}. The legitimate variables conditioned on are features that should impact the outcome regardless of group membership and thus are domain-specific. As opposed to statistical parity, conditional statistical parity is better at mitigating individual unfairness by incorporating relevant individual information when measuring fairness. A drawback of conditional statistical parity is that legitimate variables may be correlated with 
protected attributes, potentially reinforcing disparities in outcomes predicted between groups. 
    
    To formulate the MIO constraint, denote $\mathcal{W}$ as the set of all values of the legitimate features in the dataset, and let $\mathcal{I}_w \subseteq \mathcal{I}$ be the set of all datapoints with legitimate features $w \in \mathcal{W}$. Then, we can enforce conditional statistical parity with a bias tolerance of $\delta$ via the set of constraints
    \begin{equation}\label{eq:csp}
        \left|\frac{1}{|\mathcal{I}_0\cap \mathcal{I}_w|}\sum_{i\in \mathcal{I}_0 \cap \mathcal{I}_w }\zeta^i-\frac{1}{|\mathcal{I}_1\cap \mathcal{I}_w|}\sum_{i\in \mathcal{I}_1\cap \mathcal{I}_w}\zeta^i\right|\leq \delta \hspace{10mm} \forall w \in \mathcal{W}.
    \end{equation}
    In the Communities and Crime dataset, the legitimate features could, for example, be the 
police budget per population (PolicBudgPerPop), as the police budget is not based on individual identities in the community and may be linked to more resources for crime prevention. If we stratify the data into low and high police budget based on the mean (0.2), then we can let $\mathcal{W} = \{\text{low-budget, high-budget}\}$. We would then enforce statistical parity within the $w = \text{low-budget}$ and $w = \text{high-budget}$ group separately using constraint~\eqref{eq:csp}.
    
    \paragraph{Predictive Equality} Predictive equality balances the false positive rate between groups~\cite{chouldechova2017fair}. Predictive equality is useful for applications where a false positive is a serious concern and should be mitigated.
    One limitation of predictive equality is that it only considers the subset of data with an unfavorable outcome, and does not impose any fairness constraints on the datapoints with a favorable outcome.
    
    To formulate predictive equality as a constraint, we let $\mathcal{I}_-\subseteq \mathcal{I}$ be the subset of observations with an unfavorable outcome based on $\mathcal{F}$. Then, the following constraint can be added to enforce predictive equality:
    \begin{equation}\label{eq:predictive_equality}
        \left|\frac{1}{|\mathcal{I}_0\cap \mathcal{I}_-|}\sum_{i\in \mathcal{I}_0 \cap \mathcal{I}_- }\zeta^i-\frac{1}{|\mathcal{I}_1\cap \mathcal{I}_-|}\sum_{i\in \mathcal{I}_1\cap \mathcal{I}_-}\zeta^i\right|\leq \delta.
    \end{equation}
    If we enforce predictive equality the Communities and Crime dataset, we aim for cities with more than 18\% Hispanics and high-crime rates ($\mathcal{I}_0 \cap \mathcal{I}_-$) to be wrongly predicted as low-crime at the same rate as other high-crime cities ($\mathcal{I}_1 \cap \mathcal{I}_-$). An ML model that does not satisfy predictive equality may, for example, predict cities with less than 18\% Hispanics $\mathcal{I}_1$ with lower crime rates among all cities with high-crime. If the prediction model is used to inform crime-prevention resources, a lack of predictive equality may lead to high-crime cities with more than 18\% Hispanics to be given less resources than other high-crime cities.
    
    \paragraph{Equal Opportunity} Equal opportunity equalizes the true positive rates between different groups (assuming that a positive prediction is desirable)~\cite{hardt2016equality}. Equal opportunity is important in settings where non-discrimination among the positive outcome group is important. Similar to predictive equality, a downside of equal opportunity is that it only considers the subset of data with a favorable outcome, and does not enforce fairness on datapoints with an unfavorable outcome.
    
    Letting $\mathcal{I}_+\subseteq \mathcal{I}$ be the subset of observations with a favorable outcome based on $\mathcal{F}$, equal opportunity can be formulated as
    \begin{equation}\label{eq:equal_opp}
        \left|\frac{1}{|\mathcal{I}_0\cap \mathcal{I}_+|}\sum_{i\in \mathcal{I}_0 \cap \mathcal{I}_+}\zeta^i-\frac{1}{|\mathcal{I}_1\cap \mathcal{I}_+|}\sum_{i\in \mathcal{I}_1\cap \mathcal{I}_+}\zeta^i\right|\leq \delta.
    \end{equation}
    Enforcing equal opportunity in the Communities and Crime dataset means that among the cities with low crime rates, the proportion of low crime rate predictions is similar between the cities with more than 18\% Hispanics and other cities. If equal opportunity is not satisfied, then cities that have lower crime rates and more than 18\% Hispanics would be unfairly predicted with higher crime rates as opposed to other low-crime cities.
    \paragraph{Equalized Odds} Equalized odds enforces predictions to be independent of the protected group conditional on the true label~\cite{hardt2016equality}. Equivalently, equalized odds requires both predictive equality and equal opportunity, making each group have equivalent prediction errors. Naturally, enforcing equalized odds in~\eqref{eq:fair_setup} can be done by adding both constraints~\eqref{eq:predictive_equality} and~\eqref{eq:equal_opp}. Although this allows for two fairness notions to be enforced, this fairness notion may significantly reduce the accuracy of the model due to its strict requirements.


There are a couple of considerations for formulating and solving~\eqref{eq:fair_setup} with fairness constraints. First, the fairness constraints~\eqref{eq:statistical_parity},~\eqref{eq:csp},~\eqref{eq:predictive_equality}, and~\eqref{eq:equal_opp} are dense, coupling binary variables involving all datapoints, and typically increase the difficulty of solving the learning problems. Second, the specific MIO representation of the fairness constraint(s) used varies depending on the ML model and the structure of favorable outcomes, as exemplified by formulation~\eqref{eq:l0l2_fair}. For example in the context of decision rules or trees, if the decision-maker wants to ensure fairness in terms of accuracy of the ML model (e.g., predictive equality, equal opportunity, equalized odds), no additional variables are needed as variables $z^i$ already encode accuracy information in the formulation, as discussed for example in section~\ref{sec:decision_tree} -- this approach is used in \cite{lawless2021fair}. Alternatively if the label space is binary or categorical and $\mathcal{F}$ represents labels associated with a positive outcome in a downstream decision-making process, the formulations of decision rules or trees can be naturally expanded to include variables $z^i$ as discussed in \cite{aghaei2019learning,jo2023learning,lawless2023interpretable,rober2021rule}. In such cases, the computational effort to solve to optimality the fair ML problems is similar to their nominal counterparts.

In regression problems, where $\mathcal{Y}=\mathbb{R}$, the set of favorable outcomes is often represented as an interval, e.g., $\mathcal{F}^i=[t,\infty)$ for some prespecified threshold $y_0\in \mathbb{R}$, as shown in \eqref{eq:l0l2_fair_bigM}. In fact, several such constraints are added for different values of the threshold $y_0$ \cite{ye2020unbiased,ye2024distributionally,deza2024fair}. Due to the introduction of additional binary variables and weak big-M constraints, the fair ML model in this case is typically substantially harder to solve to optimality. Interestingly, as pointed out in \cite{deza2024fair,olfat2018spectral}, the natural continuous relaxation of the big-M fair constraints is related to convex proxies that have been recently proposed in the literature. 
We lastly point out that other notions of fairness have also been proposed in the MIO literature, see for example \cite{carrizosa2025enhancing} for alternative approaches to promote balance accuracy, and \cite{olfat2019convex,lawless2024fair} for fair methods in unsupervised learning.

A major bottleneck in the deployment of MIO methods for fair ML is that the approaches require the introduction of at least $|\mathcal{I}|$ binary variables, leading to methods that do not scale well to datasets with large numbers of observations. Nonetheless, in smaller datasets where these methods have been tested and compared with simpler heuristics proposed in the literature, MIO methods yield significantly better solutions in terms of fairness without compromising accuracy. 

\subsection{Extensions} \label{sec:fair_extensions}
We note that multi-group and intersectional fairness may be desired in applications, where the population can be divided up into more than two groups based on intersectional identities (e.g., Hispanic women, Hispanic men, White women, White men) and/or multiple groups within a protected characteristic (e.g., White, Black, Hispanic, Asian, etc.). A method of extending fairness to multiple groups in MIOs is to add a fairness constraint for each subgroup against the rest of the population~\citep{kearns2018preventing}. Furthermore, it may be desirable to impose one-sided fairness constraints if one group needs to be prioritized due to historical discrimination. This allows for a less-strict notion of fairness that may still meet desired fairness goals. This can be done by simply dropping the absolute value in the fairness constraint(s) and reordering the difference on the left-hand side.

Thus far, we have explored fairness in predictive ML models. Likewise, we can also define fairness measures for prescriptive ML models that assign treatments to outcomes (see section~\ref{sec:prescriptive_ml}). In this setting, there are two primary fairness notions to consider. We can impose \textit{fairness in treatment assignments}, where treatments are assigned fairly across groups. Fairness notions similar to those for predictive models can be used to define fair treatments. For instance, variable~$\zeta^i$ can be 1 if and only if $i$ receives a favorable treatment assignment in the learned model. Then, either constraints~\eqref{eq:statistical_parity},~\eqref{eq:csp},~\eqref{eq:predictive_equality}, and/or~\eqref{eq:equal_opp} can be added to the prescriptive MIO formulation. We may also be interested in \textit{fairness in treatment outcomes}, where the expected outcome should be similar between groups. This can be done by requiring the average expected outcome between groups to be similar and/or exceed a certain threshold. We refer to~\citet{jo2021learning} for an example of adding fairness constraints in MIOs to learn fair prescriptive trees.

\section{Conclusion and Future Directions}~\label{sec:conclusion}

In this tutorial, we explored several key challenges in responsible machine learning and presented MIO-based solutions to address them.
We illustrate how MIO is capable of addressing a wide range of problems in machine learning, including but not limited to interpretability, robustness, and fairness.
While this tutorial does not cover all aspects of responsible machine learning, the modeling flexibility of MIO and the continual advances in solver technology suggest that MIO has the potential to address a broader range of problems than those discussed here. We conclude by mentioning a couple of future research directions for using MIO to promote more responsible machine learning practices.

\paragraph{Privacy} With the widespread collection of data for machine learning, significant concerns have emerged around preserving individual privacy when querying or sharing that data in ML applications. One promising avenue to address such concerns is \textit{differential privacy}, which deliberately injects noise into the data to make it difficult to reverse engineer personal information. The goal of differential privacy is to add noise in a way that protects individual data contributions while preserving the overall utility of the dataset, which is often evaluated by comparing model performance on the privatized data to that on the original data. We note that MIOs have been used to learn differentially private datasets~\cite{fioretto2021differential}, but an area that can be explored is to incorporate differential privacy constraints or penalties into MIO-based ML model formulations such that private information is not leaked through the ML model. For instance, MIO-based interpretable models can be injected with differential privacy constraints that obscure private attributes during model training and in model output while still being transparent and accurate.

\paragraph{Scaling to Large Datasets} The abundance of data available today presents significant opportunities to improve the performance of machine learning models. However, training on large datasets often incurs substantial computational costs, and MIOs used for ML are no exception. This tutorial has highlighted several scalability challenges associated with MIO-based approaches and methods to improve computational efficiency through reformulations with tighter continuous relaxations and tailored solution methods. As we have shown, ongoing research continues to refine formulations and solution techniques for data-driven MIOs, and further advancements are necessary to keep pace with the ever-growing size of modern datasets.



\section{Acknowledgements}
Nathan Justin is funded in part by the National Science Foundation Graduate Research Fellowship Program (GRFP). Phebe Vayanos, Nathan Justin, and Qingshi Sun are funded in part by the National Science Foundation under CAREER grant 2046230. Andrés Gómez is funded in part by the Air Force Office of Scientific Research YIP award No. FA9550-24-1-0086. Phebe Vayanos and  Andrés Gómez were funded in part by the National Science Foundation under NRT grant 2346058. They are grateful for their support.

We use ChatGPT~\citep{chatgpt2024} to generate or help with Figures \ref{fig:mio}, \ref{fig:risk}, \ref{fig:SVM}, \ref{fig:ifelse}, \ref{fig:pca}, \ref{fig:bayesian}, and \ref{fig:matching} as well as some text editing.


\bibliographystyle{informs2014}
\bibliography{references.bib}

\begin{thebibliography}{207}
\providecommand{\natexlab}[1]{#1}
\providecommand{\url}[1]{\texttt{#1}}
\providecommand{\urlprefix}{URL }

\bibitem[{Aftabi et~al.(2025)Aftabi, Moradi, \protect\BIBand{} Mahroo}]{aftabi2025feed}
Aftabi N, Moradi N, Mahroo F (2025) Feed-forward neural networks as a mixed-integer program. \emph{Engineering with Computers} 1--19.

\bibitem[{Aghaei et~al.(2019)Aghaei, Azizi, \protect\BIBand{} Vayanos}]{aghaei2019learning}
Aghaei S, Azizi MJ, Vayanos P (2019) Learning optimal and fair decision trees for non-discriminative decision-making. \emph{Proceedings of the AAAI Conference on Artificial Intelligence}, volume~33, 1418--1426.

\bibitem[{Aghaei et~al.(2024)Aghaei, G{\'o}mez, \protect\BIBand{} Vayanos}]{aghaei2024strong}
Aghaei S, G{\'o}mez A, Vayanos P (2024) Strong optimal classification trees. \emph{Operations Research} .

\bibitem[{Akaike(1974)}]{akaike1974new}
Akaike H (1974) A new look at the statistical model identification problem. \emph{IEEE Trans. Autom. Control} 19:716.

\bibitem[{Akt{\"u}rk et~al.(2009)Akt{\"u}rk, Atamt{\"u}rk, \protect\BIBand{} G{\"u}rel}]{akturk2009strong}
Akt{\"u}rk MS, Atamt{\"u}rk A, G{\"u}rel S (2009) A strong conic quadratic reformulation for machine-job assignment with controllable processing times. \emph{Operations Research Letters} 37(3):187--191.

\bibitem[{Alcal{\'a} et~al.(2006)Alcal{\'a}, Alcal{\'a}-Fdez, Casillas, Cord{\'o}n, \protect\BIBand{} Herrera}]{alcala2006hybrid}
Alcal{\'a} R, Alcal{\'a}-Fdez J, Casillas J, Cord{\'o}n O, Herrera F (2006) Hybrid learning models to get the interpretability--accuracy trade-off in fuzzy modeling. \emph{Soft Computing} 10:717--734.

\bibitem[{Al{\`e}s et~al.(2024)Al{\`e}s, Hur{\'e}, \protect\BIBand{} Lambert}]{ales2024new}
Al{\`e}s Z, Hur{\'e} V, Lambert A (2024) New optimization models for optimal classification trees. \emph{Computers \& Operations Research} 164:106515.

\bibitem[{Amram et~al.(2022)Amram, Dunn, \protect\BIBand{} Zhuo}]{amram2022optimal}
Amram M, Dunn J, Zhuo YD (2022) Optimal policy trees. \emph{Machine Learning} 111(7):2741--2768.

\bibitem[{Anderson et~al.(2020{\natexlab{a}})Anderson, Ma, Li, \protect\BIBand{} Sojoudi}]{anderson2020tightened}
Anderson BG, Ma Z, Li J, Sojoudi S (2020{\natexlab{a}}) Tightened convex relaxations for neural network robustness certification. \emph{2020 59th IEEE Conference on Decision and Control (CDC)}, 2190--2197 (IEEE).

\bibitem[{Anderson et~al.(2020{\natexlab{b}})Anderson, Huchette, Ma, Tjandraatmadja, \protect\BIBand{} Vielma}]{anderson2020strong}
Anderson R, Huchette J, Ma W, Tjandraatmadja C, Vielma JP (2020{\natexlab{b}}) Strong mixed-integer programming formulations for trained neural networks. \emph{Mathematical Programming} 183(1):3--39.

\bibitem[{Angelino et~al.(2018)Angelino, Larus-Stone, Alabi, Seltzer, \protect\BIBand{} Rudin}]{angelino2018learning}
Angelino E, Larus-Stone N, Alabi D, Seltzer M, Rudin C (2018) Learning certifiably optimal rule lists for categorical data. \emph{Journal of Machine Learning Research} 18(234):1--78.

\bibitem[{Aouad et~al.(2023)Aouad, Elmachtoub, Ferreira, \protect\BIBand{} McNellis}]{aouad2023market}
Aouad A, Elmachtoub AN, Ferreira KJ, McNellis R (2023) Market segmentation trees. \emph{Manufacturing \& Service Operations Management} 25(2):648--667.

\bibitem[{Atamturk \protect\BIBand{} Gomez(2020)}]{pmlr-v119-atamturk20a}
Atamturk A, Gomez A (2020) Safe screening rules for {L}0-regression from perspective relaxations. III HD, Singh A, eds., \emph{Proceedings of the 37th International Conference on Machine Learning}, volume 119 of \emph{Proceedings of Machine Learning Research}, 421--430 (PMLR).

\bibitem[{Atamturk \protect\BIBand{} Gomez(2025)}]{atamturk2025rank}
Atamturk A, Gomez A (2025) Rank-one convexification for sparse regression. \emph{Forthcoming in Journal of Machine Learning Research} .

\bibitem[{Balvert(2024)}]{balvert2024iterative}
Balvert M (2024) Iterative rule extension for logic analysis of data: an milp-based heuristic to derive interpretable binary classifiers from large data sets. \emph{INFORMS Journal on Computing} 36(3):723--741.

\bibitem[{Barocas et~al.(2023)Barocas, Hardt, \protect\BIBand{} Narayanan}]{barocas2023fairness}
Barocas S, Hardt M, Narayanan A (2023) \emph{Fairness and machine learning: Limitations and opportunities} (MIT press).

\bibitem[{Bartlett \protect\BIBand{} Cussens(2017)}]{bartlett2017integer}
Bartlett M, Cussens J (2017) Integer linear programming for the bayesian network structure learning problem. \emph{Artificial Intelligence} 244:258--271.

\bibitem[{Baryannis et~al.(2019)Baryannis, Dani, \protect\BIBand{} Antoniou}]{baryannis2019predicting}
Baryannis G, Dani S, Antoniou G (2019) Predicting supply chain risks using machine learning: The trade-off between performance and interpretability. \emph{Future Generation Computer Systems} 101:993--1004.

\bibitem[{Belbasi et~al.(2023)Belbasi, Selvi, \protect\BIBand{} Wiesemann}]{belbasi2023s}
Belbasi R, Selvi A, Wiesemann W (2023) It's all in the mix: Wasserstein machine learning with mixed features. \emph{arXiv preprint arXiv:2312.12230} .

\bibitem[{Belotti et~al.(2013)Belotti, Kirches, Leyffer, Linderoth, Luedtke, \protect\BIBand{} Mahajan}]{belotti2013mixed}
Belotti P, Kirches C, Leyffer S, Linderoth J, Luedtke J, Mahajan A (2013) Mixed-integer nonlinear optimization. \emph{Acta Numerica} 22:1--131.

\bibitem[{Ben{\'\i}tez-Pe{\~n}a et~al.(2019)Ben{\'\i}tez-Pe{\~n}a, Blanquero, Carrizosa, \protect\BIBand{} Ram{\'\i}rez-Cobo}]{benitez2019cost}
Ben{\'\i}tez-Pe{\~n}a S, Blanquero R, Carrizosa E, Ram{\'\i}rez-Cobo P (2019) Cost-sensitive feature selection for support vector machines. \emph{Computers \& Operations Research} 106:169--178.

\bibitem[{Bennett et~al.(2020)Bennett, Vielma, \protect\BIBand{} Zubizarreta}]{bennett2020building}
Bennett M, Vielma JP, Zubizarreta JR (2020) Building representative matched samples with multi-valued treatments in large observational studies. \emph{Journal of Computational and Graphical Statistics} 29(4):744--757.

\bibitem[{Berk \protect\BIBand{} Bertsimas(2019)}]{berk2019certifiably}
Berk L, Bertsimas D (2019) Certifiably optimal sparse principal component analysis. \emph{Mathematical Programming Computation} 11:381--420.

\bibitem[{Bertsimas \protect\BIBand{} Dunn(2017)}]{bertsimas2017optimal}
Bertsimas D, Dunn J (2017) Optimal classification trees. \emph{Machine Learning} 106:1039--1082.

\bibitem[{Bertsimas et~al.(2016)Bertsimas, King, \protect\BIBand{} Mazumder}]{bertsimas2016best}
Bertsimas D, King A, Mazumder R (2016) Best subset selection via a modern optimization lens. \emph{The Annals of Statistics} 44(2):813--852.

\bibitem[{Bertsimas \protect\BIBand{} Mazumder(2014)}]{bertsimas2014least}
Bertsimas D, Mazumder R (2014) Least quantile regression via modern optimization. \emph{The Annals of Statistics} 2494--2525.

\bibitem[{Bertsimas et~al.(2021)Bertsimas, Orfanoudaki, \protect\BIBand{} Wiberg}]{bertsimas2021interpretable}
Bertsimas D, Orfanoudaki A, Wiberg H (2021) Interpretable clustering: an optimization approach. \emph{Machine Learning} 110(1):89--138.

\bibitem[{Bertsimas \protect\BIBand{} Van~Parys(2020)}]{bertsimas2020sparse}
Bertsimas D, Van~Parys B (2020) Sparse high-dimensional regression. \emph{The Annals of Statistics} 48(1):300--323.

\bibitem[{Bhathena et~al.(2025)Bhathena, Fattahi, G{\'o}mez, \protect\BIBand{} K{\"u}{\c{c}}{\"u}kyavuz}]{bhathena2025parametric}
Bhathena A, Fattahi S, G{\'o}mez A, K{\"u}{\c{c}}{\"u}kyavuz S (2025) A parametric approach for solving convex quadratic optimization with indicators over trees. \emph{Forthcoming in Mathematical Programming} .

\bibitem[{Boyd \protect\BIBand{} Vandenberghe(2004)}]{boyd2004convex}
Boyd S, Vandenberghe L (2004) \emph{Convex optimization} (Cambridge University Press).

\bibitem[{Breiman et~al.(2017)Breiman, Friedman, Olshen, \protect\BIBand{} Stone}]{Breiman2017ClassificationTrees}
Breiman L, Friedman JH, Olshen RA, Stone CJ (2017) \emph{{Classification and regression trees}} (Routledge), ISBN 9781315139470.

\bibitem[{Buolamwini \protect\BIBand{} Gebru(2018)}]{pmlr-v81-buolamwini18a}
Buolamwini J, Gebru T (2018) Gender shades: Intersectional accuracy disparities in commercial gender classification. Friedler SA, Wilson C, eds., \emph{Proceedings of the 1st Conference on Fairness, Accountability and Transparency}, volume~81 of \emph{Proceedings of Machine Learning Research}, 77--91 (PMLR).

\bibitem[{Carreira-Perpi{\~n}{\'a}n \protect\BIBand{} Hada(2021)}]{carreira2021counterfactual}
Carreira-Perpi{\~n}{\'a}n M{\'A}, Hada SS (2021) Counterfactual explanations for oblique decision trees: Exact, efficient algorithms. \emph{Proceedings of the AAAI Conference on Artificial Intelligence}, volume~35, 6903--6911.

\bibitem[{Carrizosa et~al.(2023)Carrizosa, Kurishchenko, Mar{\'\i}n, \protect\BIBand{} Morales}]{carrizosa2023clustering}
Carrizosa E, Kurishchenko K, Mar{\'\i}n A, Morales DR (2023) On clustering and interpreting with rules by means of mathematical optimization. \emph{Computers \& Operations Research} 154:106180.

\bibitem[{Carrizosa et~al.(2025)Carrizosa, Kurishchenko, \protect\BIBand{} Morales}]{carrizosa2025enhancing}
Carrizosa E, Kurishchenko K, Morales DR (2025) On enhancing the explainability and fairness of tree ensembles. \emph{European Journal of Operational Research} .

\bibitem[{Carrizosa et~al.(2021)Carrizosa, Molero-R{\'\i}o, \protect\BIBand{} Romero~Morales}]{carrizosa2021mathematical}
Carrizosa E, Molero-R{\'\i}o C, Romero~Morales D (2021) Mathematical optimization in classification and regression trees. \emph{Top} 29(1):5--33.

\bibitem[{Carrizosa et~al.(2022)Carrizosa, Mortensen, Morales, \protect\BIBand{} Sillero-Denamiel}]{carrizosa2022tree}
Carrizosa E, Mortensen LH, Morales DR, Sillero-Denamiel MR (2022) The tree based linear regression model for hierarchical categorical variables. \emph{Expert Systems with Applications} 203:117423.

\bibitem[{Carrizosa et~al.(2024{\natexlab{a}})Carrizosa, Ram{\'\i}rez-Ayerbe, \protect\BIBand{} Morales}]{carrizosa2024generating}
Carrizosa E, Ram{\'\i}rez-Ayerbe J, Morales DR (2024{\natexlab{a}}) Generating collective counterfactual explanations in score-based classification via mathematical optimization. \emph{Expert Systems with Applications} 238:121954.

\bibitem[{Carrizosa et~al.(2024{\natexlab{b}})Carrizosa, Ram{\'\i}rez-Ayerbe, \protect\BIBand{} Morales}]{carrizosa2024mathematical}
Carrizosa E, Ram{\'\i}rez-Ayerbe J, Morales DR (2024{\natexlab{b}}) Mathematical optimization modelling for group counterfactual explanations. \emph{European Journal of Operational Research} 319(2):399--412.

\bibitem[{Carrizosa et~al.(2024{\natexlab{c}})Carrizosa, Ram{\'\i}rez-Ayerbe, \protect\BIBand{} Romero~Morales}]{carrizosa2024new}
Carrizosa E, Ram{\'\i}rez-Ayerbe J, Romero~Morales D (2024{\natexlab{c}}) A new model for counterfactual analysis for functional data. \emph{Advances in Data Analysis and Classification} 18(4):981--1000.

\bibitem[{Cepeda et~al.(2024)Cepeda, G{\'o}mez, \protect\BIBand{} Han}]{cepeda2024robust}
Cepeda V, G{\'o}mez A, Han S (2024) Robust support vector machines via conic optimization. \emph{arXiv preprint arXiv:2402.01797} .

\bibitem[{Ceria \protect\BIBand{} Soares(1999)}]{ceria1999convex}
Ceria S, Soares J (1999) Convex programming for disjunctive convex optimization. \emph{Mathematical Programming} 86:595--614.

\bibitem[{Chen et~al.(2021)Chen, Dash, \protect\BIBand{} Gao}]{chen2021integer}
Chen R, Dash S, Gao T (2021) Integer programming for causal structure learning in the presence of latent variables. \emph{International Conference on Machine Learning}, 1550--1560 (PMLR).

\bibitem[{Chen(2023)}]{chen2023ethics}
Chen Z (2023) Ethics and discrimination in artificial intelligence-enabled recruitment practices. \emph{Humanities and Social Sciences Communications} 10(1):567.

\bibitem[{Chouldechova(2017)}]{chouldechova2017fair}
Chouldechova A (2017) Fair prediction with disparate impact: A study of bias in recidivism prediction instruments. \emph{Big Data} 5(2):153--163.

\bibitem[{Corbett-Davies et~al.(2017)Corbett-Davies, Pierson, Feller, Goel, \protect\BIBand{} Huq}]{corbett2017algorithmic}
Corbett-Davies S, Pierson E, Feller A, Goel S, Huq A (2017) Algorithmic decision making and the cost of fairness. \emph{Proceedings of the 23rd ACM SIGKDD International Conference on Knowledge Discovery and Data Mining}, 797--806.

\bibitem[{Cory-Wright \protect\BIBand{} G{\'o}mez(2023)}]{cory2023stability}
Cory-Wright R, G{\'o}mez A (2023) Stability-adjusted cross-validation for sparse linear regression. \emph{arXiv preprint arXiv:2306.14851} .

\bibitem[{Cousineau et~al.(2023)Cousineau, Verter, Murphy, \protect\BIBand{} Pineau}]{cousineau2023estimating}
Cousineau M, Verter V, Murphy SA, Pineau J (2023) Estimating causal effects with optimization-based methods: a review and empirical comparison. \emph{European Journal of Operational Research} 304(2):367--380.

\bibitem[{Cozad et~al.(2014)Cozad, Sahinidis, \protect\BIBand{} Miller}]{cozad2014learning}
Cozad A, Sahinidis NV, Miller DC (2014) Learning surrogate models for simulation-based optimization. \emph{AIChE Journal} 60(6):2211--2227.

\bibitem[{Cui et~al.(2015)Cui, Chen, He, \protect\BIBand{} Chen}]{cui2015optimal}
Cui Z, Chen W, He Y, Chen Y (2015) Optimal action extraction for random forests and boosted trees. \emph{Proceedings of the 21th ACM SIGKDD International Conference on Knowledge Discovery and Data Mining}, 179--188.

\bibitem[{Dash et~al.(2018)Dash, Gunluk, \protect\BIBand{} Wei}]{dash2018boolean}
Dash S, Gunluk O, Wei D (2018) Boolean decision rules via column generation. \emph{Advances in Neural Information Processing Systems} 31.

\bibitem[{d'Aspremont et~al.(2004)d'Aspremont, Ghaoui, Jordan, \protect\BIBand{} Lanckriet}]{d2004direct}
d'Aspremont A, Ghaoui L, Jordan M, Lanckriet G (2004) A direct formulation for sparse pca using semidefinite programming. \emph{Advances in Neural Information Processing Systems} 17.

\bibitem[{Dedieu et~al.(2021)Dedieu, Hazimeh, \protect\BIBand{} Mazumder}]{dedieu2021learning}
Dedieu A, Hazimeh H, Mazumder R (2021) Learning sparse classifiers: Continuous and mixed integer optimization perspectives. \emph{Journal of Machine Learning Research} 22(135):1--47.

\bibitem[{Dey et~al.(2022)Dey, Mazumder, \protect\BIBand{} Wang}]{dey2022using}
Dey SS, Mazumder R, Wang G (2022) Using $\ell$1-relaxation and integer programming to obtain dual bounds for sparse pca. \emph{Operations Research} 70(3):1914--1932.

\bibitem[{Deza \protect\BIBand{} Atamt{\"u}rk(2022)}]{deza2022safe}
Deza A, Atamt{\"u}rk A (2022) Safe screening for logistic regression with $\ell_0$-$\ell_2$ regularization. \emph{KDIR}, 119--126.

\bibitem[{Deza et~al.(2024)Deza, G{\'o}mez, \protect\BIBand{} Atamt{\"u}rk}]{deza2024fair}
Deza A, G{\'o}mez A, Atamt{\"u}rk A (2024) Fair and accurate regression: Strong formulations and algorithms. \emph{arXiv preprint arXiv:2412.17116} .

\bibitem[{Dietterich(1995)}]{dietterich1995overfitting}
Dietterich T (1995) Overfitting and undercomputing in machine learning. \emph{ACM Computing Surveys (CSUR)} 27(3):326--327.

\bibitem[{Dong et~al.(2015)Dong, Chen, \protect\BIBand{} Linderoth}]{dong2015regularization}
Dong H, Chen K, Linderoth J (2015) Regularization vs. relaxation: A conic optimization perspective of statistical variable selection. \emph{arXiv preprint arXiv:1510.06083} .

\bibitem[{Dud{\i}k et~al.(2014)Dud{\i}k, Erhan, Langford, \protect\BIBand{} Li}]{dudik2014doubly}
Dud{\i}k M, Erhan D, Langford J, Li L (2014) Doubly robust policy evaluation and optimization. \emph{Statistical Science} 29(4):485--511.

\bibitem[{Dwork et~al.(2012)Dwork, Hardt, Pitassi, Reingold, \protect\BIBand{} Zemel}]{dwork2012fairness}
Dwork C, Hardt M, Pitassi T, Reingold O, Zemel R (2012) Fairness through awareness. \emph{Proceedings of the 3rd Innovations in Theoretical Computer Science Conference}, 214--226.

\bibitem[{Dziugaite et~al.(2020)Dziugaite, Ben-David, \protect\BIBand{} Roy}]{dziugaite2020enforcing}
Dziugaite GK, Ben-David S, Roy DM (2020) Enforcing interpretability and its statistical impacts: Trade-offs between accuracy and interpretability. \emph{arXiv preprint arXiv:2010.13764} .

\bibitem[{D’Onofrio et~al.(2024)D’Onofrio, Grani, Monaci, \protect\BIBand{} Palagi}]{d2024margin}
D’Onofrio F, Grani G, Monaci M, Palagi L (2024) Margin optimal classification trees. \emph{Computers \& Operations Research} 161:106441.

\bibitem[{Eberhardt et~al.(2024)Eberhardt, Kaynar, \protect\BIBand{} Siddiq}]{eberhardt2024discovering}
Eberhardt F, Kaynar N, Siddiq A (2024) Discovering causal models with optimization: Confounders, cycles, and instrument validity. \emph{Management Science} .

\bibitem[{Elmachtoub et~al.(2020)Elmachtoub, Liang, \protect\BIBand{} McNellis}]{elmachtoub2020decision}
Elmachtoub AN, Liang JCN, McNellis R (2020) Decision trees for decision-making under the predict-then-optimize framework. \emph{International Conference on Machine Learning}, 2858--2867 (PMLR).

\bibitem[{{European Commission}(2021)}]{eu_ai_act_proposal_2021}
{European Commission} (2021) Proposal for a regulation of the european parliament and of the council laying down harmonised rules on artificial intelligence (artificial intelligence act) and amending certain union legislative acts. \textit{COM(2021) 206 final}.

\bibitem[{Fioretto et~al.(2021)Fioretto, Van~Hentenryck, \protect\BIBand{} Zhu}]{fioretto2021differential}
Fioretto F, Van~Hentenryck P, Zhu K (2021) Differential privacy of hierarchical census data: An optimization approach. \emph{Artificial Intelligence} 296:103475.

\bibitem[{Floudas(1995)}]{floudas1995nonlinear}
Floudas CA (1995) \emph{Nonlinear and mixed-integer optimization: fundamentals and applications} (Oxford University Press).

\bibitem[{Frangioni \protect\BIBand{} Gentile(2006)}]{frangioni2006perspective}
Frangioni A, Gentile C (2006) Perspective cuts for a class of convex 0--1 mixed integer programs. \emph{Mathematical Programming} 106:225--236.

\bibitem[{Freiesleben(2022)}]{freiesleben2022intriguing}
Freiesleben T (2022) The intriguing relation between counterfactual explanations and adversarial examples. \emph{Minds and Machines} 32(1):77--109.

\bibitem[{Giloni \protect\BIBand{} Padberg(2002)}]{giloni2002least}
Giloni A, Padberg M (2002) Least trimmed squares regression, least median squares regression, and mathematical programming. \emph{Mathematical and Computer Modelling} 35(9-10):1043--1060.

\bibitem[{Goldblum et~al.(2023)Goldblum, Tsipras, Xie, Chen, Schwarzschild, Song, Madry, Li, \protect\BIBand{} Goldstein}]{DataPoisoning}
Goldblum M, Tsipras D, Xie C, Chen X, Schwarzschild A, Song D, Madry A, Li B, Goldstein T (2023) { Dataset Security for Machine Learning: Data Poisoning, Backdoor Attacks, and Defenses }. \emph{IEEE Transactions on Pattern Analysis \& Machine Intelligence} 45(02):1563--1580, ISSN 1939-3539.

\bibitem[{G{\'o}mez(2021)}]{gomez2021outlier}
G{\'o}mez A (2021) Outlier detection in time series via mixed-integer conic quadratic optimization. \emph{SIAM Journal on Optimization} 31(3):1897--1925.

\bibitem[{G\'{o}mez et~al.(2024)G\'{o}mez, Han, \protect\BIBand{} Lozano}]{gomez2024real}
G\'{o}mez A, Han S, Lozano L (2024) Real-time solution of quadratic optimization problems with banded matrices and indicator variables. \emph{arXiv preprint arXiv:2405.03051} .

\bibitem[{G{\'o}mez \protect\BIBand{} Neto(2023)}]{gomez2023outlier}
G{\'o}mez A, Neto J (2023) Outlier detection in regression: conic quadratic formulations. \emph{arXiv preprint arXiv:2307.05975} .

\bibitem[{G{\'o}mez \protect\BIBand{} Prokopyev(2021)}]{gomez2021mixed}
G{\'o}mez A, Prokopyev OA (2021) A mixed-integer fractional optimization approach to best subset selection. \emph{INFORMS Journal on Computing} 33(2):551--565.

\bibitem[{Goodman \protect\BIBand{} Flaxman(2017)}]{goodman2017european}
Goodman B, Flaxman S (2017) European union regulations on algorithmic decision-making and a “right to explanation”. \emph{AI Magazine} 38(3):50--57.

\bibitem[{Guan et~al.(2009)Guan, Gray, \protect\BIBand{} Leyffer}]{guan2009mixed}
Guan W, Gray A, Leyffer S (2009) Mixed-integer support vector machine. \emph{NIPS Workshop on Optimization for Machine Learning}.

\bibitem[{Guidotti(2024)}]{guidotti2024counterfactual}
Guidotti R (2024) Counterfactual explanations and how to find them: Literature review and benchmarking. \emph{Data Mining and Knowledge Discovery} 38(5):2770--2824.

\bibitem[{G{\"u}nl{\"u}k et~al.(2021)G{\"u}nl{\"u}k, Kalagnanam, Li, Menickelly, \protect\BIBand{} Scheinberg}]{gunluk2021optimal}
G{\"u}nl{\"u}k O, Kalagnanam J, Li M, Menickelly M, Scheinberg K (2021) Optimal decision trees for categorical data via integer programming. \emph{Journal of Global Optimization} 81:233--260.

\bibitem[{G{\"u}nl{\"u}k \protect\BIBand{} Linderoth(2010)}]{gunluk2010perspective}
G{\"u}nl{\"u}k O, Linderoth J (2010) Perspective reformulations of mixed integer nonlinear programs with indicator variables. \emph{Mathematical Programming} 124:183--205.

\bibitem[{{Gurobi Optimization, LLC}(2024)}]{gurobi}
{Gurobi Optimization, LLC} (2024) {Gurobi Optimizer Reference Manual}.

\bibitem[{Han \protect\BIBand{} G{\'o}mez(2024)}]{han2024compact}
Han S, G{\'o}mez A (2024) Compact extended formulations for low-rank functions with indicator variables. \emph{Mathematics of Operations Research} .

\bibitem[{Hardt et~al.(2016)Hardt, Price, \protect\BIBand{} Srebro}]{hardt2016equality}
Hardt M, Price E, Srebro N (2016) Equality of opportunity in supervised learning. \emph{Advances in Neural Information Processing Systems} 29.

\bibitem[{Hastie et~al.(2009)Hastie, Tibshirani, \protect\BIBand{} Friedman}]{hastie2009elements}
Hastie T, Tibshirani R, Friedman J (2009) \emph{The Elements of Statistical Learning: Data Mining, Inference, and Prediction} (Springer), 2nd edition.

\bibitem[{Hastie et~al.(2017)Hastie, Tibshirani, \protect\BIBand{} Tibshirani}]{hastie2017extended}
Hastie T, Tibshirani R, Tibshirani RJ (2017) Extended comparisons of best subset selection, forward stepwise selection, and the lasso. \emph{arXiv preprint arXiv:1707.08692} .

\bibitem[{Hazimeh \protect\BIBand{} Mazumder(2020)}]{hazimeh2020fast}
Hazimeh H, Mazumder R (2020) Fast best subset selection: Coordinate descent and local combinatorial optimization algorithms. \emph{Operations Research} 68(5):1517--1537.

\bibitem[{Hazimeh et~al.(2022)Hazimeh, Mazumder, \protect\BIBand{} Saab}]{hazimeh2022sparse}
Hazimeh H, Mazumder R, Saab A (2022) Sparse regression at scale: Branch-and-bound rooted in first-order optimization. \emph{Mathematical Programming} 196(1):347--388.

\bibitem[{Hern{\'a}n \protect\BIBand{} Robins(2010)}]{hernan2010causal}
Hern{\'a}n MA, Robins JM (2010) Causal inference.

\bibitem[{Hojny et~al.(2024)Hojny, Zhang, Campos, \protect\BIBand{} Misener}]{hojny2024verifying}
Hojny C, Zhang S, Campos JS, Misener R (2024) Verifying message-passing neural networks via topology-based bounds tightening. \emph{arXiv preprint arXiv:2402.13937} .

\bibitem[{Hosmer \protect\BIBand{} Lemeshow(2000)}]{logisticregression}
Hosmer DW, Lemeshow S (2000) \emph{Applied logistic regression} (John Wiley and Sons), ISBN 0471356328, 9780471356325.

\bibitem[{Hua et~al.(2022)Hua, Ren, \protect\BIBand{} Cao}]{hua2022scalable}
Hua K, Ren J, Cao Y (2022) A scalable deterministic global optimization algorithm for training optimal decision tree. \emph{Advances in Neural Information Processing Systems} 35:8347--8359.

\bibitem[{Ibrahim et~al.(2024)Ibrahim, Afriat, Behdin, \protect\BIBand{} Mazumder}]{ibrahim2024grand}
Ibrahim S, Afriat G, Behdin K, Mazumder R (2024) Grand-slamin’ interpretable additive modeling with structural constraints. \emph{Advances in Neural Information Processing Systems} 36.

\bibitem[{Ingold \protect\BIBand{} Soper(2016)}]{Ingold_Soper_2016}
Ingold D, Soper S (2016) Amazon doesn’t consider the race of its customers. should it? \urlprefix\url{https://www.bloomberg.com/graphics/2016-amazon-same-day/}.

\bibitem[{{Innovation, Science and Economic Development Canada}(2023)}]{ised_ai_code_2023}
{Innovation, Science and Economic Development Canada} (2023) Voluntary code of conduct on the responsible development and management of advanced generative ai systems. \url{https://ised-isde.canada.ca/}.

\bibitem[{Insolia et~al.(2021)Insolia, Kenney, Calovi, \protect\BIBand{} Chiaromonte}]{stats4030040}
Insolia L, Kenney A, Calovi M, Chiaromonte F (2021) Robust variable selection with optimality guarantees for high-dimensional logistic regression 4(3):665--681, ISSN 2571-905X.

\bibitem[{Insolia et~al.(2022)Insolia, Kenney, Chiaromonte, \protect\BIBand{} Felici}]{insolia2022simultaneous}
Insolia L, Kenney A, Chiaromonte F, Felici G (2022) Simultaneous feature selection and outlier detection with optimality guarantees. \emph{Biometrics} 78(4):1592--1603.

\bibitem[{Jaakkola et~al.(2010)Jaakkola, Sontag, Globerson, \protect\BIBand{} Meila}]{jaakkola2010learning}
Jaakkola T, Sontag D, Globerson A, Meila M (2010) Learning bayesian network structure using lp relaxations. \emph{Proceedings of the Thirteenth International Conference on Artificial Intelligence and Statistics}, 358--365 (JMLR Workshop and Conference Proceedings).

\bibitem[{Jo et~al.(2023)Jo, Aghaei, Benson, Gomez, \protect\BIBand{} Vayanos}]{jo2023learning}
Jo N, Aghaei S, Benson J, Gomez A, Vayanos P (2023) Learning optimal fair decision trees: Trade-offs between interpretability, fairness, and accuracy. \emph{Proceedings of the 2023 AAAI/ACM Conference on AI, Ethics, and Society}, 181--192.

\bibitem[{Jo et~al.(2021)Jo, Aghaei, G{\'o}mez, \protect\BIBand{} Vayanos}]{jo2021learning}
Jo N, Aghaei S, G{\'o}mez A, Vayanos P (2021) Learning optimal prescriptive trees from observational data. \emph{arXiv preprint arXiv:2108.13628} .

\bibitem[{Johansson et~al.(2011)Johansson, S{\"o}nstr{\"o}d, Norinder, \protect\BIBand{} Bostr{\"o}m}]{johansson2011trade}
Johansson U, S{\"o}nstr{\"o}d C, Norinder U, Bostr{\"o}m H (2011) Trade-off between accuracy and interpretability for predictive in silico modeling. \emph{Future Medicinal Chemistry} 3(6):647--663.

\bibitem[{Justin et~al.(2021)Justin, Aghaei, Gomez, \protect\BIBand{} Vayanos}]{justin2021optimal}
Justin N, Aghaei S, Gomez A, Vayanos P (2021) Optimal robust classification trees. \emph{The AAAI-22 workshop on Adversarial Machine Learning and Beyond}.

\bibitem[{Justin et~al.(2023)Justin, Aghaei, G{\'o}mez, \protect\BIBand{} Vayanos}]{justin2023learning}
Justin N, Aghaei S, G{\'o}mez A, Vayanos P (2023) Learning optimal classification trees robust to distribution shifts. \emph{arXiv preprint arXiv:2310.17772} .

\bibitem[{Kallus(2017)}]{kallus2017framework}
Kallus N (2017) A framework for optimal matching for causal inference. \emph{Artificial Intelligence and Statistics}, 372--381 (PMLR).

\bibitem[{Kanamori et~al.(2020)Kanamori, Takagi, Kobayashi, \protect\BIBand{} Arimura}]{kanamori2020dace}
Kanamori K, Takagi T, Kobayashi K, Arimura H (2020) Dace: Distribution-aware counterfactual explanation by mixed-integer linear optimization. \emph{IJCAI}, 2855--2862.

\bibitem[{Kanamori et~al.(2021)Kanamori, Takagi, Kobayashi, Ike, Uemura, \protect\BIBand{} Arimura}]{kanamori2021ordered}
Kanamori K, Takagi T, Kobayashi K, Ike Y, Uemura K, Arimura H (2021) Ordered counterfactual explanation by mixed-integer linear optimization. \emph{Proceedings of the AAAI Conference on Artificial Intelligence}, volume~35, 11564--11574.

\bibitem[{Kantchelian et~al.(2016)Kantchelian, Tygar, \protect\BIBand{} Joseph}]{kantchelian2016evasion}
Kantchelian A, Tygar JD, Joseph A (2016) Evasion and hardening of tree ensemble classifiers. \emph{International Conference on Machine Learning}, 2387--2396 (PMLR).

\bibitem[{Kearns et~al.(2018)Kearns, Neel, Roth, \protect\BIBand{} Wu}]{kearns2018preventing}
Kearns M, Neel S, Roth A, Wu ZS (2018) Preventing fairness gerrymandering: Auditing and learning for subgroup fairness. \emph{International conference on machine learning}, 2564--2572 (PMLR).

\bibitem[{Khalid et~al.(2023)Khalid, Qayyum, Bilal, Al-Fuqaha, \protect\BIBand{} Qadir}]{KHALID2023106848}
Khalid N, Qayyum A, Bilal M, Al-Fuqaha A, Qadir J (2023) Privacy-preserving artificial intelligence in healthcare: Techniques and applications. \emph{Computers in Biology and Medicine} 158:106848, ISSN 0010-4825.

\bibitem[{Khalil et~al.(2018)Khalil, Gupta, \protect\BIBand{} Dilkina}]{khalil2018combinatorial}
Khalil EB, Gupta A, Dilkina B (2018) Combinatorial attacks on binarized neural networks. \emph{arXiv preprint arXiv:1810.03538} .

\bibitem[{Kim et~al.(2022)Kim, Tawarmalani, \protect\BIBand{} Richard}]{kim2022convexification}
Kim J, Tawarmalani M, Richard JPP (2022) Convexification of permutation-invariant sets and an application to sparse principal component analysis. \emph{Mathematics of Operations Research} 47(4):2547--2584.

\bibitem[{K{\"u}{\c{c}}{\"u}kyavuz et~al.(2023)K{\"u}{\c{c}}{\"u}kyavuz, Shojaie, Manzour, Wei, \protect\BIBand{} Wu}]{kucukyavuz2023consistent}
K{\"u}{\c{c}}{\"u}kyavuz S, Shojaie A, Manzour H, Wei L, Wu HH (2023) Consistent second-order conic integer programming for learning bayesian networks. \emph{Journal of Machine Learning Research} 24(322):1--38.

\bibitem[{Kuhn et~al.(2019)Kuhn, Esfahani, Nguyen, \protect\BIBand{} Shafieezadeh-Abadeh}]{kuhn2019wasserstein}
Kuhn D, Esfahani PM, Nguyen VA, Shafieezadeh-Abadeh S (2019) Wasserstein distributionally robust optimization: Theory and applications in machine learning. \emph{Operations Research \& Management Science in the Age of Analytics}, 130--166 (Informs).

\bibitem[{Kumar et~al.(2020)Kumar, Vishnu, Mitra, \protect\BIBand{} Mohan}]{9425267}
Kumar KN, Vishnu C, Mitra R, Mohan CK (2020) Black-box adversarial attacks in autonomous vehicle technology. \emph{2020 IEEE Applied Imagery Pattern Recognition Workshop (AIPR)}, 1--7.

\bibitem[{Kurtz \protect\BIBand{} Bah(2021)}]{kurtz2021efficient}
Kurtz J, Bah B (2021) Efficient and robust mixed-integer optimization methods for training binarized deep neural networks. \emph{arXiv preprint arXiv:2110.11382} .

\bibitem[{Laporte \protect\BIBand{} Nobert(1987)}]{laporte1987exact}
Laporte G, Nobert Y (1987) Exact algorithms for the vehicle routing problem. \emph{North-Holland Mathematics Studies}, volume 132, 147--184 (Elsevier).

\bibitem[{Lawless et~al.(2023)Lawless, Dash, Gunluk, \protect\BIBand{} Wei}]{lawless2023interpretable}
Lawless C, Dash S, Gunluk O, Wei D (2023) Interpretable and fair boolean rule sets via column generation. \emph{Journal of Machine Learning Research} 24(229):1--50.

\bibitem[{Lawless \protect\BIBand{} Gunluk(2021)}]{lawless2021fair}
Lawless C, Gunluk O (2021) Fair decision rules for binary classification. \emph{arXiv preprint arXiv:2107.01325} .

\bibitem[{Lawless \protect\BIBand{} Gunluk(2024)}]{lawless2024fair}
Lawless C, Gunluk O (2024) Fair minimum representation clustering via integer programming. \emph{arXiv preprint arXiv:2409.02963} .

\bibitem[{Lawless et~al.(2022)Lawless, Kalagnanam, Nguyen, Phan, \protect\BIBand{} Reddy}]{lawless2022interpretable}
Lawless C, Kalagnanam J, Nguyen LM, Phan D, Reddy C (2022) Interpretable clustering via multi-polytope machines. \emph{Proceedings of the AAAI Conference on Artificial Intelligence}, volume~36, 7309--7316.

\bibitem[{Lazimy(1982)}]{lazimy1982mixed}
Lazimy R (1982) Mixed-integer quadratic programming. \emph{Mathematical Programming} 22:332--349.

\bibitem[{Lee et~al.(2022)Lee, Yoon, \protect\BIBand{} Won}]{lee2022mixed}
Lee IG, Yoon SW, Won D (2022) A mixed integer linear programming support vector machine for cost-effective group feature selection: branch-cut-and-price approach. \emph{European Journal of Operational Research} 299(3):1055--1068.

\bibitem[{Lewis(2013)}]{lewis2013counterfactuals}
Lewis D (2013) \emph{Counterfactuals} (John Wiley \& Sons).

\bibitem[{Li et~al.(2023)Li, Qi, Liu, Di, Liu, Pei, Yi, \protect\BIBand{} Zhou}]{TrustworthyAI}
Li B, Qi P, Liu B, Di S, Liu J, Pei J, Yi J, Zhou B (2023) Trustworthy ai: From principles to practices. \emph{ACM Comput. Surv.} 55(9), ISSN 0360-0300.

\bibitem[{Li et~al.(2022)Li, Cummings, \protect\BIBand{} Mintz}]{li2022optimal}
Li Q, Cummings R, Mintz Y (2022) Optimal local explainer aggregation for interpretable prediction. \emph{Proceedings of the AAAI Conference on Artificial Intelligence}, volume~36, 12000--12007.

\bibitem[{Li \protect\BIBand{} Xie(2024)}]{li2024exact}
Li Y, Xie W (2024) Exact and approximation algorithms for sparse principal component analysis. \emph{INFORMS Journal on Computing} .

\bibitem[{Liu et~al.(2021)Liu, Ding, Shaham, Rahayu, Farokhi, \protect\BIBand{} Lin}]{PrivacyASurvey}
Liu B, Ding M, Shaham S, Rahayu W, Farokhi F, Lin Z (2021) When machine learning meets privacy: A survey and outlook. \emph{ACM Comput. Surv.} 54(2), ISSN 0360-0300.

\bibitem[{Liu et~al.(2024)Liu, Hu, Allen, \protect\BIBand{} Hermes}]{liu2024optimal}
Liu E, Hu T, Allen TT, Hermes C (2024) Optimal classification trees with leaf-branch and binary constraints. \emph{Computers \& Operations Research} 166:106629.

\bibitem[{Magnanti \protect\BIBand{} Wolsey(1995)}]{magnanti1995optimal}
Magnanti TL, Wolsey LA (1995) Optimal trees. \emph{Handbooks in Operations Research and Management Science} 7:503--615.

\bibitem[{Manzour et~al.(2021)Manzour, K{\"u}{\c{c}}{\"u}kyavuz, Wu, \protect\BIBand{} Shojaie}]{manzour2021integer}
Manzour H, K{\"u}{\c{c}}{\"u}kyavuz S, Wu HH, Shojaie A (2021) Integer programming for learning directed acyclic graphs from continuous data. \emph{INFORMS Journal on Optimization} 3(1):46--73.

\bibitem[{Maragno et~al.(2024)Maragno, Kurtz, R{\"o}ber, Goedhart, Birbil, \protect\BIBand{} den Hertog}]{maragno2024finding}
Maragno D, Kurtz J, R{\"o}ber TE, Goedhart R, Birbil {\c{S}}I, den Hertog D (2024) Finding regions of counterfactual explanations via robust optimization. \emph{INFORMS Journal on Computing} .

\bibitem[{Mazumder et~al.(2023)Mazumder, Radchenko, \protect\BIBand{} Dedieu}]{mazumder2023subset}
Mazumder R, Radchenko P, Dedieu A (2023) Subset selection with shrinkage: Sparse linear modeling when the snr is low. \emph{Operations Research} 71(1):129--147.

\bibitem[{McCormick(1976)}]{mccormick1976computability}
McCormick GP (1976) Computability of global solutions to factorable nonconvex programs: Part i—convex underestimating problems. \emph{Mathematical Programming} 10(1):147--175.

\bibitem[{Mehrabi et~al.(2021)Mehrabi, Morstatter, Saxena, Lerman, \protect\BIBand{} Galstyan}]{mehrabi2021survey}
Mehrabi N, Morstatter F, Saxena N, Lerman K, Galstyan A (2021) A survey on bias and fairness in machine learning. \emph{ACM Computing Surveys (CSUR)} 54(6):1--35.

\bibitem[{Miller(1984)}]{miller1984selection}
Miller AJ (1984) Selection of subsets of regression variables. \emph{Journal of the Royal Statistical Society Series A: Statistics in Society} 147(3):389--410.

\bibitem[{Miyashiro \protect\BIBand{} Takano(2015{\natexlab{a}})}]{miyashiro2015mixed}
Miyashiro R, Takano Y (2015{\natexlab{a}}) Mixed integer second-order cone programming formulations for variable selection in linear regression. \emph{European Journal of Operational Research} 247(3):721--731.

\bibitem[{Miyashiro \protect\BIBand{} Takano(2015{\natexlab{b}})}]{miyashiro2015subset}
Miyashiro R, Takano Y (2015{\natexlab{b}}) Subset selection by mallows’ cp: A mixed integer programming approach. \emph{Expert Systems with Applications} 42(1):325--331.

\bibitem[{Moghaddam et~al.(2005)Moghaddam, Weiss, \protect\BIBand{} Avidan}]{moghaddam2005spectral}
Moghaddam B, Weiss Y, Avidan S (2005) Spectral bounds for sparse pca: Exact and greedy algorithms. \emph{Advances in Neural Information Processing Systems} 18.

\bibitem[{Mohammadi et~al.(2021)Mohammadi, Karimi, Barthe, \protect\BIBand{} Valera}]{mohammadi2021scaling}
Mohammadi K, Karimi AH, Barthe G, Valera I (2021) Scaling guarantees for nearest counterfactual explanations. \emph{Proceedings of the 2021 AAAI/ACM Conference on AI, Ethics, and Society}, 177--187.

\bibitem[{Mohassel \protect\BIBand{} Zhang(2017)}]{SecureML}
Mohassel P, Zhang Y (2017) Secureml: A system for scalable privacy-preserving machine learning. \emph{2017 IEEE Symposium on Security and Privacy (SP)}, 19--38.

\bibitem[{Molero-R{\'\i}o \protect\BIBand{} D’Ambrosio(2024)}]{molero2024optimal}
Molero-R{\'\i}o C, D’Ambrosio C (2024) Optimal risk scores for continuous predictors. \emph{International Conference on Machine Learning, Optimization, and Data Science}, 148--162 (Springer).

\bibitem[{Moreno-Torres et~al.(2012)Moreno-Torres, Raeder, Alaiz-Rodríguez, Chawla, \protect\BIBand{} Herrera}]{MORENOTORRES2012521}
Moreno-Torres JG, Raeder T, Alaiz-Rodríguez R, Chawla NV, Herrera F (2012) A unifying view on dataset shift in classification. \emph{Pattern Recognition} 45(1):521--530, ISSN 0031-3203.

\bibitem[{Morucci et~al.(2022)Morucci, Noor-E-Alam, \protect\BIBand{} Rudin}]{morucci2022robust}
Morucci M, Noor-E-Alam M, Rudin C (2022) A robust approach to quantifying uncertainty in matching problems of causal inference. \emph{INFORMS Journal on Data Science} 1(2):156--171.

\bibitem[{Navarro-García et~al.(2025)Navarro-García, Guerrero, Durban, \protect\BIBand{} {del Cerro}}]{navarro2023feature}
Navarro-García M, Guerrero V, Durban M, {del Cerro} A (2025) Feature and functional form selection in additive models via mixed-integer optimization. \emph{Computers \& Operations Research} 176:106945, ISSN 0305-0548.

\bibitem[{Obermeyer et~al.(2019)Obermeyer, Powers, Vogeli, \protect\BIBand{} Mullainathan}]{GroupBaisinHealth}
Obermeyer Z, Powers B, Vogeli C, Mullainathan S (2019) Dissecting racial bias in an algorithm used to manage the health of populations. \emph{Science} 366(6464):447--453.

\bibitem[{Olfat \protect\BIBand{} Aswani(2018)}]{olfat2018spectral}
Olfat M, Aswani A (2018) Spectral algorithms for computing fair support vector machines. \emph{International Conference on Artificial Intelligence and Statistics}, 1933--1942 (PMLR).

\bibitem[{Olfat \protect\BIBand{} Aswani(2019)}]{olfat2019convex}
Olfat M, Aswani A (2019) Convex formulations for fair principal component analysis. \emph{Proceedings of the AAAI Conference on Artificial Intelligence}, volume~33, 663--670.

\bibitem[{OpenAI(2025)}]{chatgpt2024}
OpenAI (2025) Chatgpt (june 2025 version). \url{https://chat.openai.com/chat}.

\bibitem[{Park \protect\BIBand{} Klabjan(2017)}]{park2017bayesian}
Park YW, Klabjan D (2017) Bayesian network learning via topological order. \emph{Journal of Machine Learning Research} 18(99):1--32.

\bibitem[{Parmentier \protect\BIBand{} Vidal(2021)}]{parmentier2021optimal}
Parmentier A, Vidal T (2021) Optimal counterfactual explanations in tree ensembles. \emph{International Conference on Machine Learning}, 8422--8431 (PMLR).

\bibitem[{Patil \protect\BIBand{} Mintz(2022)}]{patil2022mixed}
Patil V, Mintz Y (2022) A mixed-integer programming approach to training dense neural networks. \emph{arXiv preprint arXiv:2201.00723} .

\bibitem[{Poljak \protect\BIBand{} Wolkowicz(1995)}]{poljak1995convex}
Poljak S, Wolkowicz H (1995) Convex relaxations of (0, 1)-quadratic programming. \emph{Mathematics of Operations Research} 20(3):550--561.

\bibitem[{Puerto \protect\BIBand{} Torrej{\'o}n(2024)}]{puerto2024fresh}
Puerto J, Torrej{\'o}n A (2024) A fresh view on least quantile of squares regression based on new optimization approaches. \emph{arXiv preprint arXiv:2410.17793} .

\bibitem[{Qayyum et~al.(2021)Qayyum, Qadir, Bilal, \protect\BIBand{} Al-Fuqaha}]{9153891}
Qayyum A, Qadir J, Bilal M, Al-Fuqaha A (2021) Secure and robust machine learning for healthcare: A survey. \emph{IEEE Reviews in Biomedical Engineering} 14:156--180.

\bibitem[{Qui{\~n}onero-Candela et~al.(2022)Qui{\~n}onero-Candela, Sugiyama, Schwaighofer, \protect\BIBand{} Lawrence}]{quinonero2022dataset}
Qui{\~n}onero-Candela J, Sugiyama M, Schwaighofer A, Lawrence ND (2022) \emph{Dataset shift in machine learning} (Mit Press).

\bibitem[{Ribeiro et~al.(2018)Ribeiro, Singh, \protect\BIBand{} Guestrin}]{ribeiro2018anchors}
Ribeiro MT, Singh S, Guestrin C (2018) Anchors: High-precision model-agnostic explanations. \emph{Proceedings of the AAAI Conference on Artificial Intelligence}, volume~32.

\bibitem[{R{\"o}ber et~al.(2021)R{\"o}ber, Lumadjeng, Aky{\"u}z, \protect\BIBand{} Birbil}]{rober2021rule}
R{\"o}ber TE, Lumadjeng AC, Aky{\"u}z MH, Birbil {\c{S}}{\.I} (2021) Rule generation for classification: Scalability, interpretability, and fairness. \emph{arXiv preprint arXiv:2104.10751} .

\bibitem[{Rousseeuw(1984)}]{rousseeuw1984least}
Rousseeuw PJ (1984) Least median of squares regression. \emph{Journal of the American Statistical Association} 79(388):871--880.

\bibitem[{Rousseeuw(1987)}]{rousseeuw1987silhouettes}
Rousseeuw PJ (1987) Silhouettes: a graphical aid to the interpretation and validation of cluster analysis. \emph{Journal of Computational and Applied Mathematics} 20:53--65.

\bibitem[{Rousseeuw \protect\BIBand{} Leroy(2003)}]{rousseeuw2003robust}
Rousseeuw PJ, Leroy AM (2003) \emph{Robust regression and outlier detection} (John Wiley \& Sons).

\bibitem[{Rudin(2019)}]{rudin2019stop}
Rudin C (2019) Stop explaining black box machine learning models for high stakes decisions and use interpretable models instead. \emph{Nature Machine Intelligence} 1(5):206--215.

\bibitem[{Rudin et~al.(2022)Rudin, Chen, Chen, Huang, Semenova, \protect\BIBand{} Zhong}]{rudin2022interpretable}
Rudin C, Chen C, Chen Z, Huang H, Semenova L, Zhong C (2022) Interpretable machine learning: Fundamental principles and 10 grand challenges. \emph{Statistic Surveys} 16:1--85.

\bibitem[{Rudin \protect\BIBand{} Ertekin(2018)}]{rudin2018learning}
Rudin C, Ertekin {\c{S}} (2018) Learning customized and optimized lists of rules with mathematical programming. \emph{Mathematical Programming Computation} 10:659--702.

\bibitem[{Russell(2019)}]{russell2019efficient}
Russell C (2019) Efficient search for diverse coherent explanations. \emph{Proceedings of the conference on Fairness, Accountability, and Transparency}, 20--28.

\bibitem[{Sato et~al.(2016)Sato, Takano, Miyashiro, \protect\BIBand{} Yoshise}]{sato2016feature}
Sato T, Takano Y, Miyashiro R, Yoshise A (2016) Feature subset selection for logistic regression via mixed integer optimization. \emph{Computational Optimization and Applications} 64:865--880.

\bibitem[{Sayad et~al.(2019)Sayad, Mousannif, \protect\BIBand{} {Al Moatassime}}]{SAYAD2019130}
Sayad YO, Mousannif H, {Al Moatassime} H (2019) Predictive modeling of wildfires: A new dataset and machine learning approach. \emph{Fire Safety Journal} 104:130--146, ISSN 0379-7112.

\bibitem[{Schwarz(1978)}]{schwarz1978estimating}
Schwarz G (1978) Estimating the dimension of a model. \emph{The Annals of Statistics} 461--464.

\bibitem[{Shafahi et~al.(2018)Shafahi, Huang, Najibi, Suciu, Studer, Dumitras, \protect\BIBand{} Goldstein}]{poisonFrogs}
Shafahi A, Huang WR, Najibi M, Suciu O, Studer C, Dumitras T, Goldstein T (2018) Poison frogs! targeted clean-label poisoning attacks on neural networks. \emph{Proceedings of the 32nd International Conference on Neural Information Processing Systems}, 6106–6116, NIPS'18 (Red Hook, NY, USA: Curran Associates Inc.).

\bibitem[{Shafiee \protect\BIBand{} K{\i}l{\i}n{\c{c}}-Karzan(2024)}]{shafiee2024constrained}
Shafiee S, K{\i}l{\i}n{\c{c}}-Karzan F (2024) Constrained optimization of rank-one functions with indicator variables. \emph{Mathematical Programming} 208(1):533--579.

\bibitem[{Subramanian et~al.(2022)Subramanian, Sun, Drissi, \protect\BIBand{} Ettl}]{subramanian2022constrained}
Subramanian S, Sun W, Drissi Y, Ettl M (2022) Constrained prescriptive trees via column generation. \emph{Proceedings of the AAAI Conference on Artificial Intelligence}, volume~36, 4602--4610.

\bibitem[{Sun \protect\BIBand{} Nikolaev(2016)}]{sun2016mutual}
Sun L, Nikolaev AG (2016) Mutual information based matching for causal inference with observational data. \emph{Journal of Machine Learning Research} 17(199):1--31.

\bibitem[{Sun et~al.(2025)Sun, Justin, Gomez, \protect\BIBand{} Vayanos}]{sun2025mixed}
Sun Q, Justin N, Gomez A, Vayanos P (2025) Mixed-feature logistic regression robust to distribution shifts. \emph{The 28th International Conference on Artificial Intelligence and Statistics}.

\bibitem[{Thorbjarnarson \protect\BIBand{} Yorke-Smith(2021)}]{thorbjarnarson2021training}
Thorbjarnarson T, Yorke-Smith N (2021) On training neural networks with mixed integer programming. \emph{IJCAI-PRICAI’20 Workshop on Data Science Meets Optimisation}.

\bibitem[{Thorbjarnarson \protect\BIBand{} Yorke-Smith(2023)}]{thorbjarnarson2023optimal}
Thorbjarnarson T, Yorke-Smith N (2023) Optimal training of integer-valued neural networks with mixed integer programming. \emph{Plos One} 18(2):e0261029.

\bibitem[{Tibshirani(1996)}]{tibshirani1996regression}
Tibshirani R (1996) Regression shrinkage and selection via the lasso. \emph{Journal of the Royal Statistical Society Series B: Statistical Methodology} 58(1):267--288.

\bibitem[{Tillmann et~al.(2024)Tillmann, Bienstock, Lodi, \protect\BIBand{} Schwartz}]{tillmann2024cardinality}
Tillmann AM, Bienstock D, Lodi A, Schwartz A (2024) Cardinality minimization, constraints, and regularization: a survey. \emph{SIAM Review} 66(3):403--477.

\bibitem[{Tjandraatmadja et~al.(2020)Tjandraatmadja, Anderson, Huchette, Ma, Patel, \protect\BIBand{} Vielma}]{tjandraatmadja2020convex}
Tjandraatmadja C, Anderson R, Huchette J, Ma W, Patel KK, Vielma JP (2020) The convex relaxation barrier, revisited: Tightened single-neuron relaxations for neural network verification. \emph{Advances in Neural Information Processing Systems} 33:21675--21686.

\bibitem[{Tjeng et~al.(2017)Tjeng, Xiao, \protect\BIBand{} Tedrake}]{tjeng2017evaluating}
Tjeng V, Xiao K, Tedrake R (2017) Evaluating robustness of neural networks with mixed integer programming. \emph{arXiv preprint arXiv:1711.07356} .

\bibitem[{{U.S. Congress}(1996)}]{hipaa1996}
{US Congress} (1996) Health insurance portability and accountability act of 1996. Public Law 104-191.

\bibitem[{Ustun \protect\BIBand{} Rudin(2019)}]{ustun2019learning}
Ustun B, Rudin C (2019) Learning optimized risk scores. \emph{Journal of Machine Learning Research} 20(150):1--75.

\bibitem[{Ustun et~al.(2019)Ustun, Spangher, \protect\BIBand{} Liu}]{ustun2019actionable}
Ustun B, Spangher A, Liu Y (2019) Actionable recourse in linear classification. \emph{Proceedings of the conference on Fairness, Accountability, and Transparency}, 10--19.

\bibitem[{van~der Linden et~al.(2024)van~der Linden, Vos, de~Weerdt, Verwer, \protect\BIBand{} Demirovi{\'c}}]{van2024optimal}
van~der Linden JG, Vos D, de~Weerdt MM, Verwer S, Demirovi{\'c} E (2024) Optimal or greedy decision trees? revisiting their objectives, tuning, and performance. \emph{arXiv preprint arXiv:2409.12788} .

\bibitem[{Venzke \protect\BIBand{} Chatzivasileiadis(2020)}]{venzke2020verification}
Venzke A, Chatzivasileiadis S (2020) Verification of neural network behaviour: Formal guarantees for power system applications. \emph{IEEE Transactions on Smart Grid} 12(1):383--397.

\bibitem[{Verwer \protect\BIBand{} Zhang(2017)}]{verwer2017learning}
Verwer S, Zhang Y (2017) Learning decision trees with flexible constraints and objectives using integer optimization. \emph{Integration of AI and OR Techniques in Constraint Programming: 14th International Conference, CPAIOR 2017, Padua, Italy, June 5-8, 2017, Proceedings 14}, 94--103 (Springer).

\bibitem[{Verwer \protect\BIBand{} Zhang(2019)}]{verwer2019learning}
Verwer S, Zhang Y (2019) Learning optimal classification trees using a binary linear program formulation. \emph{Proceedings of the AAAI Conference on Artificial Intelligence}, volume~33, 1625--1632.

\bibitem[{Vielma(2015)}]{vielma2015mixed}
Vielma JP (2015) Mixed integer linear programming formulation techniques. \emph{Siam Review} 57(1):3--57.

\bibitem[{Vos \protect\BIBand{} Verwer(2022)}]{vos2022robust}
Vos D, Verwer S (2022) Robust optimal classification trees against adversarial examples. \emph{Proceedings of the AAAI Conference on Artificial Intelligence}, volume~36, 8520--8528.

\bibitem[{Walsh et~al.(2017)Walsh, Ribeiro, \protect\BIBand{} Franklin}]{RiskofSuicide}
Walsh CG, Ribeiro JD, Franklin JC (2017) Predicting risk of suicide attempts over time through machine learning. \emph{Clinical Psychological Science} 5(3):457--469.

\bibitem[{Wang \protect\BIBand{} Dey(2020)}]{wang2020upper}
Wang G, Dey S (2020) Upper bounds for model-free row-sparse principal component analysis. \emph{International Conference on Machine Learning}, 9868--9875 (PMLR).

\bibitem[{Wang \protect\BIBand{} Rudin(2015)}]{wang2015learning}
Wang T, Rudin C (2015) Learning optimized or's of and's. \emph{arXiv preprint arXiv:1511.02210} .

\bibitem[{Wei et~al.(2022)Wei, G{\'o}mez, \protect\BIBand{} K{\"u}{\c{c}}{\"u}kyavuz}]{wei2022ideal}
Wei L, G{\'o}mez A, K{\"u}{\c{c}}{\"u}kyavuz S (2022) Ideal formulations for constrained convex optimization problems with indicator variables. \emph{Mathematical Programming} 192(1):57--88.

\bibitem[{Wilson \protect\BIBand{} Sahinidis(2017)}]{wilson2017alamo}
Wilson ZT, Sahinidis NV (2017) The alamo approach to machine learning. \emph{Computers \& Chemical Engineering} 106:785--795.

\bibitem[{Wolsey \protect\BIBand{} Nemhauser(1999)}]{wolsey1999integer}
Wolsey LA, Nemhauser GL (1999) \emph{Integer and combinatorial optimization} (John Wiley \& Sons).

\bibitem[{Xiao et~al.(2015)Xiao, Biggio, Brown, Fumera, Eckert, \protect\BIBand{} Roli}]{pmlr-v37-xiao15}
Xiao H, Biggio B, Brown G, Fumera G, Eckert C, Roli F (2015) Is feature selection secure against training data poisoning? Bach F, Blei D, eds., \emph{Proceedings of the 32nd International Conference on Machine Learning}, volume~37 of \emph{Proceedings of Machine Learning Research}, 1689--1698 (Lille, France: PMLR).

\bibitem[{Xie \protect\BIBand{} Deng(2020)}]{xie2020scalable}
Xie W, Deng X (2020) Scalable algorithms for the sparse ridge regression. \emph{SIAM Journal on Optimization} 30(4):3359--3386.

\bibitem[{Yala et~al.(2019)Yala, Lehman, Schuster, Portnoi, \protect\BIBand{} Barzilay}]{yala2019deep}
Yala A, Lehman C, Schuster T, Portnoi T, Barzilay R (2019) A deep learning mammography-based model for improved breast cancer risk prediction. \emph{Radiology} 292(1):60--66.

\bibitem[{Ye et~al.(2024)Ye, Hanasusanto, \protect\BIBand{} Xie}]{ye2024distributionally}
Ye Q, Hanasusanto GA, Xie W (2024) Distributionally fair stochastic optimization using wasserstein distance. \emph{arXiv preprint arXiv:2402.01872} .

\bibitem[{Ye \protect\BIBand{} Xie(2020)}]{ye2020unbiased}
Ye Q, Xie W (2020) Unbiased subdata selection for fair classification: A unified framework and scalable algorithms. \emph{arXiv preprint arXiv:2012.12356} .

\bibitem[{You et~al.(2022)You, Sun, Guo, Tan, \protect\BIBand{} Jiang}]{you2022interpretability}
You Y, Sun J, Guo Y, Tan Y, Jiang J (2022) Interpretability and accuracy trade-off in the modeling of belief rule-based systems. \emph{Knowledge-Based Systems} 236:107491.

\bibitem[{Zhang et~al.(2020)Zhang, Wang, Liu, \protect\BIBand{} Wang}]{zhang2020decision}
Zhang F, Wang Y, Liu S, Wang H (2020) Decision-based evasion attacks on tree ensemble classifiers. \emph{World Wide Web} 23:2957--2977.

\bibitem[{Zhang et~al.(2022)Zhang, Lou, Wang, Wu, Lu, \protect\BIBand{} Jia}]{IEEEautonomousVehicle}
Zhang J, Lou Y, Wang J, Wu K, Lu K, Jia X (2022) Evaluating adversarial attacks on driving safety in vision-based autonomous vehicles. \emph{IEEE Internet of Things Journal} 9(5):3443--3456.

\bibitem[{Zharmagambetov et~al.(2021)Zharmagambetov, Hada, Gabidolla, \protect\BIBand{} Carreira-Perpin{\'a}n}]{zharmagambetov2021non}
Zharmagambetov A, Hada SS, Gabidolla M, Carreira-Perpin{\'a}n MA (2021) Non-greedy algorithms for decision tree optimization: An experimental comparison. \emph{2021 International Joint Conference on Neural Networks (IJCNN)}, 1--8 (IEEE).

\bibitem[{Zhou et~al.(2021)Zhou, Verma, Mittal, \protect\BIBand{} Chen}]{9635182}
Zhou J, Verma S, Mittal M, Chen F (2021) Understanding relations between perception of fairness and trust in algorithmic decision making. \emph{2021 8th International Conference on Behavioral and Social Computing (BESC)}, 1--5.

\bibitem[{Zhu et~al.(2020)Zhu, Murali, Phan, Nguyen, \protect\BIBand{} Kalagnanam}]{zhu2020scalable}
Zhu H, Murali P, Phan D, Nguyen L, Kalagnanam J (2020) A scalable mip-based method for learning optimal multivariate decision trees. \emph{Advances in Neural Information Processing Systems} 33:1771--1781.

\bibitem[{Zioutas \protect\BIBand{} Avramidis(2005)}]{zioutas2005deleting}
Zioutas G, Avramidis A (2005) Deleting outliers in robust regression with mixed integer programming. \emph{Acta Mathematicae Applicatae Sinica} 21:323--334.

\bibitem[{Zioutas et~al.(2009)Zioutas, Pitsoulis, \protect\BIBand{} Avramidis}]{zioutas2009quadratic}
Zioutas G, Pitsoulis L, Avramidis A (2009) Quadratic mixed integer programming and support vectors for deleting outliers in robust regression. \emph{Annals of Operations Research} 166:339--353.

\bibitem[{Zou \protect\BIBand{} Hastie(2005)}]{zou2005regularization}
Zou H, Hastie T (2005) Regularization and variable selection via the elastic net. \emph{Journal of the Royal Statistical Society Series B: Statistical Methodology} 67(2):301--320.

\bibitem[{Zubizarreta(2012)}]{zubizarreta2012using}
Zubizarreta JR (2012) Using mixed integer programming for matching in an observational study of kidney failure after surgery. \emph{Journal of the American Statistical Association} 107(500):1360--1371.

\end{thebibliography}

\end{document}